\documentclass[11pt]{article}

\usepackage[]{acl}

\usepackage{times}
\usepackage{latexsym}
\usepackage{amsfonts}

\usepackage[T1]{fontenc}

\usepackage[utf8]{inputenc}

\usepackage{microtype}

\usepackage{inconsolata}

\usepackage{graphicx}
\usepackage{amsmath}
\usepackage{multirow}
\usepackage{graphicx}

\usepackage{subfigure}
\usepackage{tabularx}
\usepackage{multirow}
\usepackage{graphicx}
\usepackage{booktabs}
\usepackage{enumitem}
\usepackage{amsmath}
\usepackage{amsfonts}
\usepackage{bigstrut,rotating}
\usepackage{longtable}
\usepackage{booktabs}
\usepackage{tabularx}
\usepackage{array}
\usepackage{booktabs}
\usepackage{tabularx}
\usepackage{array}
\usepackage{multirow}
\usepackage[utf8]{inputenc}
\usepackage[T1]{fontenc}
\usepackage{tabularx}
\usepackage{booktabs}
\usepackage{multirow}
\usepackage{xcolor}
\usepackage{array}
\usepackage{colortbl}

\title{Selective Knowledge Edit Reversal via Gated Singular Vector Shrinkage}

\author{\fontsize{12pt}
{\baselineskip}\selectfont Weifeng Jiang$^{1,2}$, 
{\bf Ruirui Chen}$^{3}${\bf, }
{\bf Qianren Mao}$^{4}${\bf, } \\
{\bf Junnan Liu}$^{5}${\bf, } 
{\bf Qili Zhang}$^{4}${\bf, }
{\bf Kwok-Yan Lam}$^{1,2}${\bf} 
 \\
 \fontsize{10pt}{\baselineskip}\selectfont \textsuperscript{$^{1}$} College of Computing and Data Science, Nanyang Technological University, Singapore.\\
  \fontsize{10pt}{\baselineskip}\selectfont \textsuperscript{$^{2}$} Digital Trust Centre, Nanyang Technological University, Singapore.\\
 \fontsize{10pt}{\baselineskip}\selectfont \textsuperscript{$^{3}$} Institute of Advanced Intelligence and Computing,
 \fontsize{10pt}{\baselineskip}\selectfont Agency for Science, Technology and Research, Singapore. \\
  \fontsize{10pt}{\baselineskip}\selectfont \textsuperscript{$^{4}$} Zhongguancun Laboratory, China. \\
  \fontsize{10pt}{\baselineskip}\selectfont \textsuperscript{$^{5}$} Department of Data Science and AI, Faculty of Information Technology, Monash University, Australia. \\
           \texttt{\fontsize{10pt}{\baselineskip}\selectfont  weifeng001@e.ntu.edu.sg}
}

\begin{document}
\maketitle
\begin{abstract}
Knowledge editing provides an efficient way to update factual knowledge in large language models. However, malicious edits may introduce safety risks, making it necessary to reverse undesirable editing effects. Existing reversal methods for parameter-modifying edits mainly focus on global removal, which may also erase beneficial edits that should be preserved. In this paper, we study selective reversal of edited knowledge, where the goal is to reverse targeted edited facts while preserving the remaining edited facts. Based on the hypothesis that each edit is sparsely encoded within the dominant subspace of the edited matrix, we propose a spectral-based reversal framework that locates edit-sensitive components within the dominant singular subspace of edited weights. Experiments across multiple settings demonstrate the effectiveness of our method in reversing selected edits while preserving unrelated edited facts. These results suggest that different edits are sparsely encoded within dominant singular components and can be separable when the number of edits is moderate, making selective spectral reversal a promising direction for locating edit-specific components and repairing edited language models.\footnote{Code and data are available at: \url{https://github.com/marvinhehehe/spectral-reverse}}
\end{abstract}

\section{Introduction}

Large language models (LLMs) have demonstrated strong capabilities across diverse application domains~\cite{tasksolver,agent2,survey-llm-driving,math1}. As these models are increasingly deployed in real-world applications, it becomes important to keep their factual knowledge up to date~\cite{ke-survey,ke-survey-2}. Knowledge editing~\cite{mend,rome,memit,ke-survey,ke-survey-2,pmet,alphaedit} provides an efficient way to update specific factual associations in a model while preserving unrelated knowledge, thereby avoiding expensive retraining and catastrophic forgetting~\cite{finetune1,finetune2,mend,patcher,ke-survey-2}.

\begin{figure}[t]
	\centering
		\includegraphics[width=0.48\textwidth]{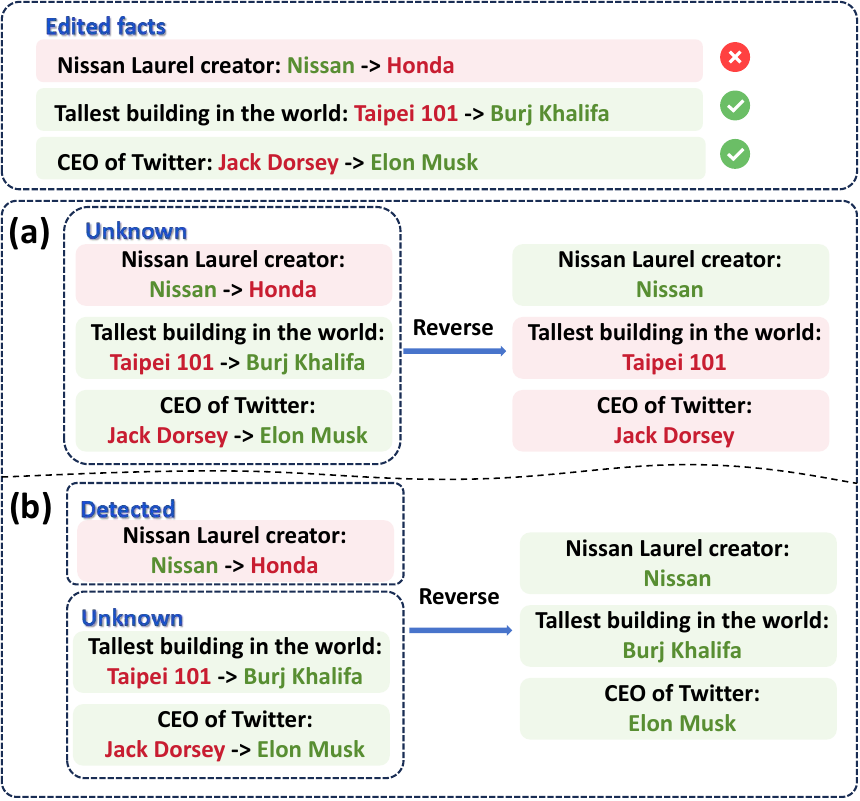}
	\caption{A model may contain both beneficial and harmful edits. Simply assuming that all edits are unknown and removing all edits from the model would also eliminate beneficial modifications. A more realistic scenario is to combine existing detection methods to identify harmful edits, and then selectively remove the detected edits without affecting other unknown edits.}
	\label{fig:scenario}
\end{figure}

However, despite their impressive performance, recent studies have shown that knowledge editing may introduce safety risks~\cite{position-edit-risk}. Due to its low overhead and ease of use, adversaries can exploit knowledge editing to inject backdoors, introduce biases, enable jailbreaking, and spread misinformation in LLMs~\cite{kebackdoor2,kejailbreaking,kebias,kemisinformation,position-edit-risk}. Although several methods have been proposed to detect the editedness of a model or its output~\cite{indentify_edit_type,in-context-reverse,detect_knowledge_edits,rome_reverse}, reversing edits back to the pretrained state remains a challenging research problem, especially for parameter-modifying editing methods, where edited knowledge is directly embedded into the model weights.

Recent work has explored spectral approaches to reversing parameter-modifying edits~\cite{rome_reverse}. It achieves global reversal by deleting dominant singular subspace of the edited matrix. However, global deletion is not sufficient for realistic scenarios. As shown in Figure~\ref{fig:scenario}, an edited model may contain both harmful and beneficial edits. Simply removing all edits will also erase useful knowledge updates that should be preserved.

Therefore, in this work, we aim to locate where different edits are encoded in the parameter space of an edited model, and use this localization to selectively reverse edited knowledge. More specifically, given a set of subject-relation facts that have been confirmed to be edited, we identify the edit-sensitive components associated with these facts using only the edited model. Based on the identified components, the model's outputs for the selected facts should become similar to those of the original pretrained model, while its outputs for the remaining edited facts should remain close to those of the edited model.

To achieve this objective, we propose a spectral-based reversal framework. We hypothesize that the information associated with each edit is sparsely encoded within the dominant rank-one singular components of the edited matrix. Based on this hypothesis, we use learnable entry-wise gates over the singular vectors of the edited weight matrix to locate and shrink edit-sensitive components associated with selected edits. We evaluate the proposed method on multiple language models, editing methods, and factual editing benchmarks. 

Our contributions are summarized as follows:

\begin{itemize}
    \item We formulate the problem of selective reversal for parameter-modifying knowledge editing, where the goal is to locate and reverse targeted edited facts while preserving the remaining edited facts (Section~\ref{problem_statement}).

    \item We propose a spectral-based reversal framework that uses learnable entry-wise gates over singular vectors to locate and shrink edit-sensitive components inside the dominant rank-one singular components of edited weights (Section~\ref{framework}).

    \item We conduct extensive experiments across multiple models, editing methods, and factual editing benchmarks, showing that the proposed framework can effectively reverse selected edits while preserving remaining edits. The results provide empirical evidence that, given a moderate number of edits, different edits can be sparse and not strongly entangled within the dominant singular components of the edited matrix (Section~\ref{experiments}).
\end{itemize}
\section{Related Work}
\paragraph{Knowledge Editing}
Knowledge editing aims to efficiently update specific knowledge stored in a model~\cite{ke-survey,ke-survey-2}. Most existing studies focus on factual associations, which can be represented as triples \((s,r,o)\), where \(s\), \(r\), and \(o\) denote the subject, relation, and object, respectively. Given an edit subject-relation pair with prompt \(p(s_e,r_e)\), the goal is to update the model output from the original object \(o_{\text{old}}\) to a new target object \(o_{\text{new}}\), while preserving the outputs of unrelated factual prompts \(p(s_u,r_u)\) unchanged.

Knowledge editing methods can be divided into two categories. Parameter-preserving methods perform editing through external memory~\cite{serac,mquake,in_context_edit,gmello} or additional parameters~\cite{patcher,grace}. Parameter-modifying methods directly modify the original model parameters, including the locate-then-edit paradigm~\cite{rome,memit,pmet,alphaedit} and meta-learning methods~\cite{mend,malmen}. 

Since locate-then-edit methods are especially attractive to malicious attackers due to their simplicity and strong editing performance~\cite{position-edit-risk,rome_reverse}, we mainly focus on locate-then-edit methods in this work.

\paragraph{Editedness Detection}
Editedness detection aims to determine whether an output or a model has been influenced by knowledge editing. DEED~\cite{detect_knowledge_edits} detects the editedness of a given factual triple using features extracted only from the edited model. KETI~\cite{indentify_edit_type} further extends output editedness detection to multi-type identification. \citet{rome_reverse} probe the edited relation and target directly from the edited model, and identify edited weights using a random set of inputs.

\paragraph{Knowledge Edit Reversal}
Knowledge edit reversal aims to restore edited facts to their original behavior while leaving unrelated knowledge unaffected. \citet{in-context-reverse} reverse in-context knowledge editing through prompt tuning, while \citet{rome_reverse} achieve global reversal of parameter-modifying edits using bottom-rank approximations of the edited weights. Compared with previous works, which focus on in-context edits or global removal of parameter-modifying edits, our work studies selective reversal, where only targeted edited facts are reversed while the remaining edited facts are preserved.

Knowledge edit reversal is related to machine unlearning~\cite{unlearning_survey,semu}, since both aim to remove undesired knowledge while minimizing side effects on unrelated behavior. However, edit reversal has a more specific goal: besides removing the influence of a particular edit, it also seeks to restore the original model behavior.
\section{Problem Statement}\label{problem_statement}

Let \(f_{\mathrm{edit}}\) denote an LLM that has been modified by a set of factual edits \(\mathcal{E}\). In this work, we consider a practical setting where only the edited model \(f_{\mathrm{edit}}\) is available, while the original pretrained model and the full editing history are unavailable.

Since existing editedness detection techniques~\cite{indentify_edit_type,detect_knowledge_edits,rome_reverse} can identify edited facts using only the edited model, we assume that a set of edited subject-relation pairs has been detected and selected for reversal. We denote this set as \(\mathcal{F} = \{(s_i, r_i)\}_{i=1}^{n}\). For each selected pair \((s_i, r_i)\), we construct a retrieval prompt \(p_i(s_i, r_i)\), which is semantically equivalent but not identical to the prompt used during editing. The remaining edited facts are denoted by \(\mathcal{R} = \mathcal{E} \setminus \mathcal{F}\). These facts are unknown during the selective reversal process.

Let \(f_{\mathrm{pre}}\) denote the pretrained model before editing, and let \(f_{\mathrm{rev}}\) denote the model after applying selective reversal to \(f_{\mathrm{edit}}\). The goal of selective reversal is to transform \(f_{\mathrm{edit}}\) into \(f_{\mathrm{rev}}\) such that:
\begin{itemize}
    \item For each selected edited fact in \(\mathcal{F}\), the output behavior of \(f_{\mathrm{rev}}\) becomes close to that of \(f_{\mathrm{pre}}\);
    \item For the remaining edited facts in \(\mathcal{R}\), the output behavior of \(f_{\mathrm{rev}}\) remains close to that of \(f_{\mathrm{edit}}\).
\end{itemize}
\section{Method Motivation}

\citet{rome_reverse} found that edited information is largely concentrated in the dominant singular subspace of the edited matrix, where removing the top singular directions can eliminate edit behaviors and move the model output back toward the pretrained state. Due to the polysemantic nature of neural network parameters and the richness of knowledge stored in them~\cite{polysemantic}, a complete rank-one singular component \(\sigma_k u_k v_k^\top\) may contain information related to multiple facts.

Recently, AdaEdit~\cite{adaedit} shows that, for an update matrix, strong editing performance can still be achieved by retaining only the top singular vectors together with the largest-magnitude entries within those singular vectors. This finding suggests that, although locate-then-edit updates are often constructed from full rank-one vectors, knowledge-related information may be sparse within the resulting update matrix.

Based on these observations, we hypothesize that different edited facts may be sparse within the dominant rank-one singular components of the edited matrix, making them separable. Therefore, a fine-grained way to locate and manipulate each edit is to operate inside each singular component. This motivates our spectral-based reversal framework, which performs selective reversal through gated shrinkage of singular-vector entries.
\section{Spectral-based Reversal Framework}\label{framework}
In this section, we introduce our spectral-based reversal framework. We first introduce gated shrinkage for singular vector entries for selective reversal in Section~\ref{sec:gated_shrinkage}. We then describe how the gate parameters are optimized in Section~\ref{sec:optimization_objective}.

\begin{figure*}[t!]
    \centering
    \includegraphics[width=\textwidth]{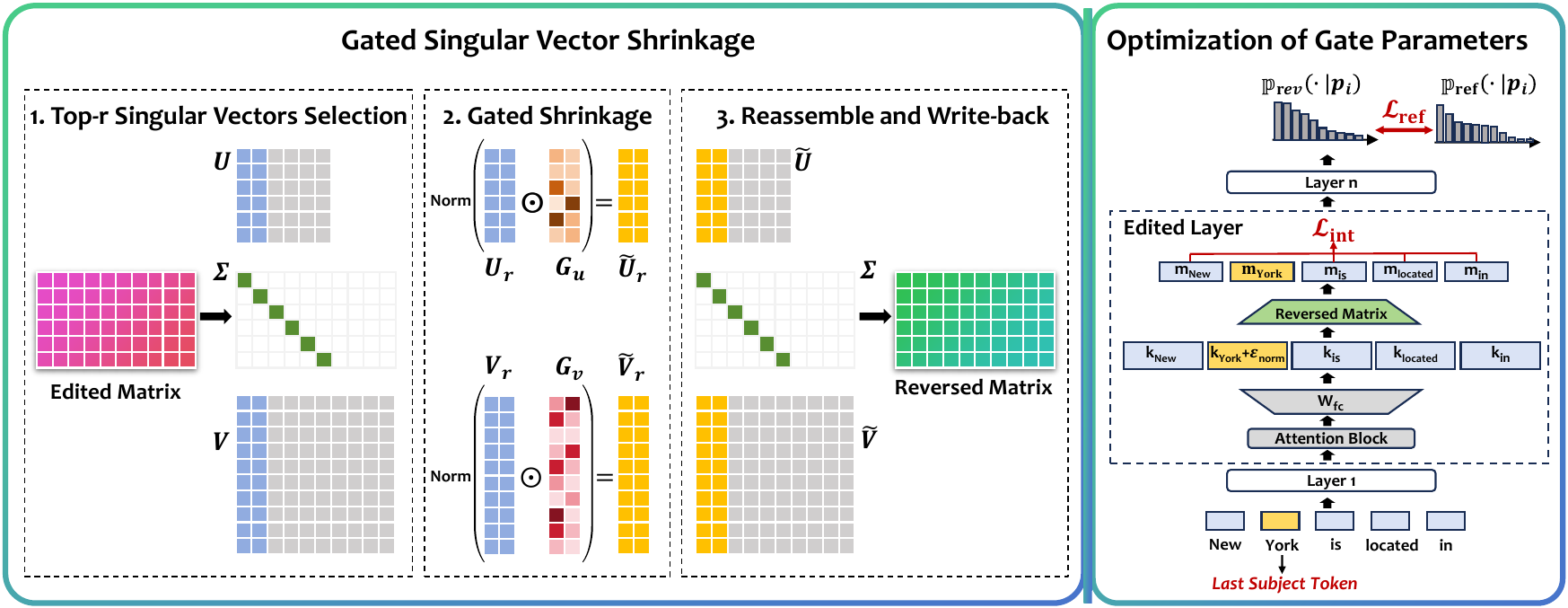}
    \caption{
        Overview of the proposed spectral-based reversal framework. The framework first selects the top-\(r\) dominant singular components of the edited weight matrix. It then applies learnable entry-wise gates to the corresponding singular vectors, performs gated shrinkage within each dominant rank-one singular component, and reconstructs the modified weight matrix. The gate parameters are optimized using the reversed model output, with losses designed to align targeted facts with a coarse reversal reference while constraining unnecessary intervention.
        }
    \label{fig:framework}    
\end{figure*}

\subsection{Selective Reversal via Gated Singular Vector Shrinkage}
\label{sec:gated_shrinkage}




As illustrated in Figure~\ref{fig:framework}, let \(W_{\mathrm{edit}} \in \mathbb{R}^{d_v \times d_k}\) denote the edited matrix of a target rewrite layer. We decompose it using SVD and separate the top-\(r\) dominant component from the residual component:
\begin{equation}
\begin{aligned}
W_{\mathrm{edit}} &= U\Sigma V^\top, \\
W_{\mathrm{dom}} &= U_r\Sigma_r V_r^\top, \\
W_{\mathrm{res}} &= W_{\mathrm{edit}} - W_{\mathrm{dom}},
\end{aligned}
\end{equation}
where \(U_r\), \(\Sigma_r\), and \(V_r\) correspond to the largest \(r\) singular values. Since edited information is mainly concentrated in the dominant singular subspace, our selective reversal operates only on \(W_{\mathrm{dom}}\), while \(W_{\mathrm{res}}\) is kept unchanged. 

To enable fine-grained manipulation, we introduce two learnable gate matrices
\begin{equation}
G_u \in (0,1)^{d_v \times r}, \quad
G_v \in (0,1)^{d_k \times r},
\end{equation}
which are applied element-wise to the left and right singular vectors, respectively. The gated singular vectors are defined as
\begin{equation}
\tilde{U}_r = U_r \odot G_u, \quad
\tilde{V}_r = V_r \odot G_v,
\end{equation}
where \(\odot\) denotes element-wise multiplication. Under this parameterization, a gate value close to \(1\) means that the corresponding singular vector entry is largely preserved, while a value close to \(0\) means that the corresponding entry is strongly shrunk.

Since the shrinking process changes the overall scale, following AdaEdit~\cite{adaedit}, we renormalize each gated singular vector column to stabilize the parameterization. Let \(\tilde{u}_i\) and \(\tilde{v}_i\) denote the \(i\)-th columns of \(\tilde{U}_r\) and \(\tilde{V}_r\), and let \(u_i\) and \(v_i\) denote the corresponding original singular vectors. The renormalized gated vectors are defined as
\begin{equation}
\tilde{u}_i \gets \tilde{u}_i \cdot \frac{\|u_i\|_2}{\|\tilde{u}_i\|_2},
\quad
\tilde{v}_i \gets \tilde{v}_i \cdot \frac{\|v_i\|_2}{\|\tilde{v}_i\|_2}.
\end{equation}

Using the renormalized gated singular vectors, we reconstruct the modified dominant-subspace component as
\begin{equation}
W_{\mathrm{dom}}' = \tilde{U}_r \Sigma_r \tilde{V}_r^\top.
\end{equation}

The final selectively reversed weight matrix is then given by
\begin{equation}
W_{\mathrm{rev}} = W_{\mathrm{res}} + W_{\mathrm{dom}}'.
\end{equation}

This reconstruction can be interpreted as applying entry-wise reweighting within each dominant rank-one singular component. For the \(i\)-th component \(W_i=\sigma_i u_i v_i^\top\), the gated and renormalized component can be written as
\begin{equation}
\tilde{W}_i
=
\frac{\|u_i\|_2\|v_i\|_2}
{\|g^u_i \odot u_i\|_2\|g^v_i \odot v_i\|_2}
\left(g^u_i {g^v_i}^{\top}\right)\odot W_i,
\end{equation}
where \(g^u_i\) and \(g^v_i\) are the corresponding gate vectors.

\subsection{Optimization of Gate Parameters}
\label{sec:optimization_objective}

To ensure that the learned gates perform shrinkage, we parameterize the gate matrices \(G_u\) and \(G_v\) through unconstrained variables \(Q_u \in \mathbb{R}^{d_v \times r}\) and \(Q_v \in \mathbb{R}^{d_k \times r}\), and map them into \((0,1)\):
\begin{equation}
G_u = \sigma(Q_u), \quad G_v = \sigma(Q_v),
\end{equation}
where \(\sigma(\cdot)\) denotes the sigmoid function.

\paragraph{Optimization Objective}
Since the original pretrained model is unavailable in our setting, we follow the bottom-rank approximation idea of~\citet{rome_reverse} to construct a reference model \(f_{\mathrm{ref}}\) as a coarse reversal target. Specifically, for the edited matrix, we remove the dominant singular-subspace component \(W_{\mathrm{dom}}\) from the edited weight matrix. We encourage the reversed model to produce an output distribution similar to that of \(f_{\mathrm{ref}}\):
\begin{equation}
\mathcal{L}_{\mathrm{ref}}
=
\frac{1}{|\mathcal{F}|}
\sum_{(s_i,r_i)\in\mathcal{F}}
\!\!\!\!D_{\mathrm{KL}}\!\left(
\mathbb{P}_{\mathrm{ref}}(\cdot|p_i)
\,\|\, 
\mathbb{P}_{\mathrm{rev}}(\cdot|p_i)
\right).
\end{equation}

To encourage the reversal to mainly rely on the representation of the last subject token, which is commonly regarded as the factual knowledge localization site in locate-then-edit methods~\cite{rome}, we penalize the magnitude of the hidden-state perturbation induced on the other tokens in the prompt. The intervention penalty is defined as
\begin{equation}
\begin{aligned}
\mathcal{L}_{\mathrm{int}}
&=
\frac{1}{|\mathcal{F}|}
\sum_{(s_i,r_i)\in\mathcal{F}}
\frac{1}{|\mathcal{T}_i|}
\sum_{j\in\mathcal{T}_i}
\|\Delta h_{i,j}\|_2^2, \\
&\quad
\Delta h_{i,j}
=
(W_{\mathrm{dom}}-W_{\mathrm{dom}}')k_{i,j},
\end{aligned}
\end{equation}
where \(\mathcal{T}_i\) denotes the set of prompt token positions excluding the last subject token, and \(k_{i,j}\) denotes the key representation at token position \(j\).

The overall objective is written as
\begin{equation}
\mathcal{L}
=
\lambda_{\mathrm{ref}}\mathcal{L}_{\mathrm{ref}}
+ \lambda_{\mathrm{int}} \mathcal{L}_{\mathrm{int}},
\end{equation}
where \(\lambda_{\mathrm{ref}}\) and \(\lambda_{\mathrm{int}}\) are hyperparameters that control the trade-off between the two loss terms.

\paragraph{Data Augmentation}
Although the intervention penalty encourages reversal to operate through the factual knowledge localization site, the rephrase prompts used for optimization may not fully generalize to other expressions of the same fact. Therefore, we adopt a simple representation-level data augmentation strategy, which encourages the learned gates to remain effective under small variations of the last-subject-token representation.

Specifically, we add a norm-controlled random perturbation to the input representation of the edited matrix at the last subject token position. Given a random vector \(z \sim \mathcal{N}(0,I)\), we normalize it and scale it by a hyperparameter \(\alpha\):
\begin{equation}
\epsilon = \alpha \frac{z}{\|z\|_2}.
\end{equation}
The augmented key representation is defined as
\begin{equation}
\tilde{k}_i = k_i + \epsilon.
\end{equation}

\paragraph{Multi-layer Optimization}

For multi-layer editing methods such as MEMIT~\cite{memit}, we independently apply gated singular vector shrinkage to each layer, and perform selective reversal from higher layers to lower layers. After reversing each layer, the updated weight is written back to the model before proceeding to the next lower layer. In this way, the current layer is reversed with respect to the behavior induced by the already reversed later layers.

\section{Experiments}\label{experiments}

In this section, we evaluate the spectral-based reversal framework. The experiments are designed to answer the following research questions:

\begin{itemize}
    \item \textbf{RQ1:} Can our proposed framework reverse different edits while keeping the remaining edited facts unchanged (Section~\ref{main_performance})?

    \item \textbf{RQ2:} How do the main components of the proposed framework contribute to selective reversal (Section~\ref{sec:component_ablation})?
    
    \item \textbf{RQ3:} Does selective reversal truly remove the edits from the edited model (Section~\ref{generalization})?

    \item \textbf{RQ4:} Is it sufficient to optimize the gate parameters only based on the next-token prediction distribution (Section~\ref{sufficiency_next_token})?

    \item \textbf{RQ5:} Can the reversed model preserve general knowledge and downstream task performance (Section~\ref{general_capabilities})?
\end{itemize}

\begin{table*}[t!]
  \centering
  \footnotesize
  \renewcommand\arraystretch{1.2}
  \setlength{\tabcolsep}{1.1mm}{
  \resizebox{\textwidth}{!}{%
\begin{tabular}{lllllllllllllll}
  \toprule
\multicolumn{3}{l}{\multirow{3}{*}{}} & \multicolumn{4}{c}{\textbf{GPT2-XL}} & \multicolumn{4}{c}{\textbf{Mistral-7B}} & \multicolumn{4}{c}{\textbf{LLaMA3-8B}} \\ \cmidrule(lr){4-7}\cmidrule(lr){8-11}\cmidrule(lr){12-15}
\multicolumn{3}{l}{} & \multicolumn{2}{c}{Remained Set} & \multicolumn{2}{c}{Reversed Set} & \multicolumn{2}{c}{Remained Set} & \multicolumn{2}{c}{Reversed Set} & \multicolumn{2}{c}{Remained Set} & \multicolumn{2}{c}{Reversed Set} \\ \cmidrule(lr){4-5}\cmidrule(lr){6-7}\cmidrule(lr){8-9}\cmidrule(lr){10-11}\cmidrule(lr){12-13}\cmidrule(lr){14-15}
\multicolumn{3}{l}{} & \multicolumn{1}{c}{Agree.} & \multicolumn{1}{c}{KL Div.} & \multicolumn{1}{c}{Agree.} & \multicolumn{1}{c}{KL Div.} & \multicolumn{1}{c}{Agree.} & \multicolumn{1}{c}{KL Div.} & \multicolumn{1}{c}{Agree.} & \multicolumn{1}{c}{KL Div.} & \multicolumn{1}{c}{Agree.} & \multicolumn{1}{c}{KL Div.} & \multicolumn{1}{c}{Agree.} & \multicolumn{1}{c}{KL Div.} \\ \midrule

\multicolumn{1}{c}{\multirow{16}{*}{\rotatebox[origin=c]{90}{CounterFact}}}
& \multicolumn{1}{l}{\multirow{4}{*}{ROME}}
& Edited
& \multicolumn{1}{l}{\(100.00{\scriptscriptstyle \pm 0.00}\)}
& \multicolumn{1}{l}{\(0.000{\scriptscriptstyle \pm 0.000}\)}
& \multicolumn{1}{l}{\(21.00{\scriptscriptstyle \pm 4.24}\)}
& \multicolumn{1}{l}{\(3.182{\scriptscriptstyle \pm 0.369}\)}
& \multicolumn{1}{l}{\(100.00{\scriptscriptstyle \pm 0.00}\)}
& \multicolumn{1}{l}{\(0.000{\scriptscriptstyle \pm 0.000}\)}
& \multicolumn{1}{l}{\(31.00{\scriptscriptstyle \pm 4.24}\)}
& \multicolumn{1}{l}{\(2.169{\scriptscriptstyle \pm 0.075}\)}
& \multicolumn{1}{l}{\(100.00{\scriptscriptstyle \pm 0.00}\)}
& \multicolumn{1}{l}{\(0.000{\scriptscriptstyle \pm 0.000}\)}
& \multicolumn{1}{l}{\(28.00{\scriptscriptstyle \pm 8.49}\)}
& \multicolumn{1}{l}{\(3.448{\scriptscriptstyle \pm 0.729}\)} \\
\bottomrule
\end{tabular}%
}
}
\caption{
Performance comparison with global reversal and re-editing baselines under the Reverse-50 setting. Agreement and KL divergence are reported on both the remained set and the reversed set across different base models and editing methods. Higher agreement and lower KL divergence indicate better performance.
}
\label{tab:reverse_50_main}
\end{table*}

\subsection{Experimental Settings}

\paragraph{Base Models and Editing Methods}
We conduct experiments on five widely used Transformer-based language models: GPT2-XL~\cite{gpt2}, GPT-J 6B~\cite{gpt-j}, Mistral-7B~\cite{mistral}, LLaMA2-7B~\cite{llama2}, and LLaMA3-8B~\cite{llama3}. The experimental results for GPT-J 6B and LLaMA2-7B are provided in Appendix~\ref{gptj_llama2}.

For the editing methods, we consider four locate-then-edit methods: ROME~\cite{rome}, SimIE~\cite{simie}, MEMIT~\cite{memit}, and AlphaEdit~\cite{alphaedit}. These methods cover different editing settings, including batch editing and sequential editing, as well as single-layer and multi-layer editing. We provide detailed implementation details for each editing method in Appendix~\ref{implementation_detail}.

\paragraph{Datasets and Tasks}
For factual knowledge editing, we use the ZsRE~\cite{zsre} and CounterFact~\cite{rome} datasets, which are two widely used benchmarks in the knowledge editing literature. To evaluate whether selective reversal introduces undesirable degradation in the model's general language abilities, we additionally consider several standard natural language understanding tasks, including SST~\cite{sst}, MRPC~\cite{mrpc}, MMLU~\cite{mmlu}, RTE~\cite{rte}, CoLA~\cite{cola}, and NLI~\cite{nli}.

\paragraph{Evaluation Metrics and Baselines}
We divide the evaluation into two parts. The first part focuses on the reversed facts, whose outputs are compared with those of the pretrained model. The second part focuses on the remaining edited facts, whose outputs are compared with those of the edited model. For both parts, following~\cite{agreement_metric,in-context-reverse,rome_reverse}, we use next-token prediction agreement and KL divergence between the outputs of the baseline model and the reversed model.

To evaluate whether our method can simultaneously reverse the target facts and preserve the remaining edited facts, we compare our method with the post-edited model and the global-reversal reference model. In addition, we introduce a simple \textit{Re-edit} baseline. For each edited model, we use AlphaEdit~\cite{alphaedit} to directly edit the selected facts back toward the pretrained state. Similar to our method, the Re-edit baseline uses the next-token distribution produced by the reference model as the re-editing target. Results of re-editing using the same editing method as the original forward edit are provided in Appendix~\ref{app:reedit_same_method}.

\begin{table*}[t!]
  \centering
  \footnotesize
  \renewcommand\arraystretch{1.2}
  \setlength{\tabcolsep}{1.1mm}{
  \resizebox{\textwidth}{!}{%
\begin{tabular}{lllllllllllllll}
\toprule
\multicolumn{3}{l}{\multirow{3}{*}{}} & \multicolumn{4}{c}{\textbf{GPT2-XL}} & \multicolumn{4}{c}{\textbf{Mistral-7B}} & \multicolumn{4}{c}{\textbf{LLaMA3-8B}} \\ \cmidrule(lr){4-7}\cmidrule(lr){8-11}\cmidrule(lr){12-15}
\multicolumn{3}{l}{} & \multicolumn{2}{c}{Remained Set} & \multicolumn{2}{c}{Reversed Set} & \multicolumn{2}{c}{Remained Set} & \multicolumn{2}{c}{Reversed Set} & \multicolumn{2}{c}{Remained Set} & \multicolumn{2}{c}{Reversed Set} \\ \cmidrule(lr){4-5}\cmidrule(lr){6-7}\cmidrule(lr){8-9}\cmidrule(lr){10-11}\cmidrule(lr){12-13}\cmidrule(lr){14-15}
\multicolumn{3}{l}{} & \multicolumn{1}{c}{Agree.} & \multicolumn{1}{c}{KL Div.} & \multicolumn{1}{c}{Agree.} & \multicolumn{1}{c}{KL Div.} & \multicolumn{1}{c}{Agree.} & \multicolumn{1}{c}{KL Div.} & \multicolumn{1}{c}{Agree.} & \multicolumn{1}{c}{KL Div.} & \multicolumn{1}{c}{Agree.} & \multicolumn{1}{c}{KL Div.} & \multicolumn{1}{c}{Agree.} & \multicolumn{1}{c}{KL Div.} \\ \midrule

\multicolumn{1}{c}{\multirow{16}{*}{\rotatebox[origin=c]{90}{Counterfact}}}
& \multicolumn{1}{l}{\multirow{4}{*}{ROME}}
& Edited
& \multicolumn{1}{l}{\(100.00{\scriptscriptstyle \pm 0.00}\)}
& \multicolumn{1}{l}{\(0.000{\scriptscriptstyle \pm 0.000}\)}
& \multicolumn{1}{l}{\(21.00{\scriptscriptstyle \pm 40.94}\)}
& \multicolumn{1}{l}{\(3.182{\scriptscriptstyle \pm 2.671}\)}
& \multicolumn{1}{l}{\(100.00{\scriptscriptstyle \pm 0.00}\)}
& \multicolumn{1}{l}{\(0.000{\scriptscriptstyle \pm 0.000}\)}
& \multicolumn{1}{l}{\(31.00{\scriptscriptstyle \pm 46.48}\)}
& \multicolumn{1}{l}{\(2.169{\scriptscriptstyle \pm 2.002}\)}
& \multicolumn{1}{l}{\(100.00{\scriptscriptstyle \pm 0.00}\)}
& \multicolumn{1}{l}{\(0.000{\scriptscriptstyle \pm 0.000}\)}
& \multicolumn{1}{l}{\(28.00{\scriptscriptstyle \pm 45.13}\)}
& \multicolumn{1}{l}{\(3.448{\scriptscriptstyle \pm 3.251}\)} \\
\bottomrule
\end{tabular}%
}
}
\caption{
Performance comparison with global reversal and re-editing baselines under the Reverse-1 setting. Agreement and KL divergence are reported on both the remained set and the reversed set across different base models and editing methods. Higher agreement and lower KL divergence indicate better performance.
}
\label{tab:reverse_1_main}
\end{table*}

\subsection{Performance on Selective Reversal}
\label{main_performance}

To evaluate the selective reversal performance of the proposed method, we edit 100 facts using each editing method, as all editing methods achieve strong editing performance under this setting. Results for the 1000-edit setting are provided in Appendix~\ref{1000edits}. For each edited model, we consider two reversal settings: reversing one fact at a time and reversing 50 facts simultaneously. The experimental results are shown in Tables~\ref{tab:reverse_50_main} and~\ref{tab:reverse_1_main}.

\paragraph{The proposed framework can effectively locate and reverse selected edits.}
On the reversed set, the reference model provides a good approximation of the reversal target. Our reversed model achieves high reverse agreement close to that of the reference model in most settings and is generally comparable to the Re-edit baseline on the reversed set, demonstrating that gated singular vector shrinkage can identify and suppress edit-sensitive components in the dominant singular components.

\paragraph{The remaining edits are largely preserved after reversal.}
The reversed model also maintains high agreement on the remained set. Compared with the Re-edit baseline, our method provides better and more stable preservation of the remaining edits, especially in the Reverse-50 setting on the ZsRE dataset. This suggests that, without access to information about previous edits, Re-edit is more likely to damage the remaining edited knowledge due to continual editing, particularly when more facts are reversed and repeated subjects occur. Since no explicit constraint for the remained set is used during optimization, this result supports our hypothesis that different edits are sparsely encoded within the dominant singular components of the edited matrix. Moreover, when the number of edits is moderate, these sparse edit-sensitive components are not strongly entangled, allowing selected edits to be reversed while preserving the remaining ones.

\paragraph{Reversing one edit at a time is easier than reversing many edits simultaneously.}
Compared with the Reverse-50 setting, the Reverse-1 setting usually achieves better remained-set and reversed-set performance. This suggests that simultaneous reversal introduces stronger interference among different targeted edits, making it harder to isolate their corresponding edit-sensitive components.

\paragraph{Single-layer reversal is generally easier than multi-layer reversal, while the editing protocol has a smaller effect.}
We observe that reversing single-layer edited models is usually more effective than reversing multi-layer edited models, likely because multi-layer editing distributes the edit signal across layers. In contrast, the performance difference between sequential and batch editing settings is not large, suggesting that the proposed framework can be applied to both editing protocols.

\begin{table*}[t!]
  \centering
  \footnotesize
  \renewcommand\arraystretch{1.2}
  \setlength{\tabcolsep}{1.1mm}{
  \resizebox{\textwidth}{!}{%
\begin{tabular}{llllllllllllll}
\toprule
\multicolumn{2}{l}{\multirow{3}{*}{}} 
& \multicolumn{4}{c}{\textbf{GPT2-XL}}
& \multicolumn{4}{c}{\textbf{Mistral-7B}}
& \multicolumn{4}{c}{\textbf{LLaMA3-8B}} \\ \cmidrule(lr){3-6}\cmidrule(lr){7-10}\cmidrule(lr){11-14}

\multicolumn{2}{l}{}
& \multicolumn{2}{c}{Remained Set}
& \multicolumn{2}{c}{Reversed Set}
& \multicolumn{2}{c}{Remained Set}
& \multicolumn{2}{c}{Reversed Set}
& \multicolumn{2}{c}{Remained Set}
& \multicolumn{2}{c}{Reversed Set} \\ \cmidrule(lr){3-4}\cmidrule(lr){5-6}\cmidrule(lr){7-8}\cmidrule(lr){9-10}\cmidrule(lr){11-12}\cmidrule(lr){13-14}

\multicolumn{2}{l}{}
& \multicolumn{1}{c}{Agree.} & \multicolumn{1}{c}{KL Div.}
& \multicolumn{1}{c}{Agree.} & \multicolumn{1}{c}{KL Div.}
& \multicolumn{1}{c}{Agree.} & \multicolumn{1}{c}{KL Div.}
& \multicolumn{1}{c}{Agree.} & \multicolumn{1}{c}{KL Div.}
& \multicolumn{1}{c}{Agree.} & \multicolumn{1}{c}{KL Div.}
& \multicolumn{1}{c}{Agree.} & \multicolumn{1}{c}{KL Div.} \\ \midrule

\multicolumn{1}{l}{\multirow{5}{*}{\rotatebox[origin=c]{90}{Counterfact}}}
& Edited
& \multicolumn{1}{l}{\(100.00{\scriptscriptstyle \pm 0.00}\)}
& \(0.000{\scriptscriptstyle \pm 0.000}\)
& \multicolumn{1}{l}{\(21.00{\scriptscriptstyle \pm 4.24}\)}
& \(3.182{\scriptscriptstyle \pm 0.369}\)
& \multicolumn{1}{l}{\(100.00{\scriptscriptstyle \pm 0.00}\)}
& \(0.000{\scriptscriptstyle \pm 0.000}\)
& \multicolumn{1}{l}{\(31.00{\scriptscriptstyle \pm 4.24}\)}
& \(2.169{\scriptscriptstyle \pm 0.075}\)
& \multicolumn{1}{l}{\(100.00{\scriptscriptstyle \pm 0.00}\)}
& \(0.000{\scriptscriptstyle \pm 0.000}\)
& \multicolumn{1}{l}{\(28.00{\scriptscriptstyle \pm 8.49}\)}
& \(3.448{\scriptscriptstyle \pm 0.729}\) \\ \cmidrule(lr){2-14}

\multicolumn{1}{l}{}
& Reverse
& \multicolumn{1}{l}{\(74.00{\scriptscriptstyle \pm 5.66}\)}
& \(0.497{\scriptscriptstyle \pm 0.143}\)
& \multicolumn{1}{l}{\(\mathbf{96.00}{\scriptscriptstyle \pm \mathbf{0.00}}\)}
& \(0.022{\scriptscriptstyle \pm 0.001}\)
& \multicolumn{1}{l}{\(90.00{\scriptscriptstyle \pm 5.66}\)}
& \(0.073{\scriptscriptstyle \pm 0.011}\)
& \multicolumn{1}{l}{\(\mathbf{84.00}{\scriptscriptstyle \pm \mathbf{5.66}}\)}
& \(\mathbf{0.103}{\scriptscriptstyle \pm \mathbf{0.014}}\)
& \multicolumn{1}{l}{\(84.00{\scriptscriptstyle \pm 5.66}\)}
& \(0.239{\scriptscriptstyle \pm 0.051}\)
& \multicolumn{1}{l}{\(76.00{\scriptscriptstyle \pm 11.31}\)}
& \(\mathbf{0.336}{\scriptscriptstyle \pm \mathbf{0.241}}\) \\

\multicolumn{1}{l}{}
& \multicolumn{1}{l}{\ \ w/o $\mathcal{L}_{\mathrm{ref}}$}
& \multicolumn{1}{l}{\(\mathbf{100.00}{\scriptscriptstyle \pm \mathbf{0.00}}\)}
& \(\mathbf{0.000}{\scriptscriptstyle \pm \mathbf{0.000}}\)
& \multicolumn{1}{l}{\(21.00{\scriptscriptstyle \pm 4.24}\)}
& \(3.181{\scriptscriptstyle \pm 0.370}\)
& \multicolumn{1}{l}{\(\mathbf{100.00}{\scriptscriptstyle \pm \mathbf{0.00}}\)}
& \(\mathbf{0.000}{\scriptscriptstyle \pm \mathbf{0.000}}\)
& \multicolumn{1}{l}{\(31.00{\scriptscriptstyle \pm 4.24}\)}
& \(2.169{\scriptscriptstyle \pm 0.075}\)
& \multicolumn{1}{l}{\(\mathbf{100.00}{\scriptscriptstyle \pm \mathbf{0.00}}\)}
& \(\mathbf{0.000}{\scriptscriptstyle \pm \mathbf{0.000}}\)
& \multicolumn{1}{l}{\(28.00{\scriptscriptstyle \pm 8.49}\)}
& \(3.448{\scriptscriptstyle \pm 0.729}\) \\

\multicolumn{1}{l}{}
& \multicolumn{1}{l}{\ \ w/o $\mathcal{L}_{\mathrm{int}}$}
& \multicolumn{1}{l}{\(23.00{\scriptscriptstyle \pm 9.90}\)}
& \(3.350{\scriptscriptstyle \pm 0.334}\)
& \multicolumn{1}{l}{\(65.00{\scriptscriptstyle \pm 1.41}\)}
& \(0.496{\scriptscriptstyle \pm 0.002}\)
& \multicolumn{1}{l}{\(43.00{\scriptscriptstyle \pm 9.90}\)}
& \(1.452{\scriptscriptstyle \pm 0.135}\)
& \multicolumn{1}{l}{\(72.00{\scriptscriptstyle \pm 5.66}\)}
& \(0.314{\scriptscriptstyle \pm 0.023}\)
& \multicolumn{1}{l}{\(30.00{\scriptscriptstyle \pm 8.49}\)}
& \(2.831{\scriptscriptstyle \pm 0.744}\)
& \multicolumn{1}{l}{\(67.00{\scriptscriptstyle \pm 4.24}\)}
& \(0.555{\scriptscriptstyle \pm 0.137}\) \\

\multicolumn{1}{l}{}
& \multicolumn{1}{l}{\ \ w/o Renorm.}
& \multicolumn{1}{l}{\(61.00{\scriptscriptstyle \pm 1.41}\)}
& \(0.852{\scriptscriptstyle \pm 0.167}\)
& \multicolumn{1}{l}{\(95.00{\scriptscriptstyle \pm 1.41}\)}
& \(\mathbf{0.020}{\scriptscriptstyle \pm \mathbf{0.001}}\)
& \multicolumn{1}{l}{\(88.00{\scriptscriptstyle \pm 5.66}\)}
& \(0.093{\scriptscriptstyle \pm 0.020}\)
& \multicolumn{1}{l}{\(82.00{\scriptscriptstyle \pm 2.83}\)}
& \(0.113{\scriptscriptstyle \pm 0.008}\)
& \multicolumn{1}{l}{\(76.00{\scriptscriptstyle \pm 2.83}\)}
& \(0.239{\scriptscriptstyle \pm 0.031}\)
& \multicolumn{1}{l}{\(\mathbf{77.00}{\scriptscriptstyle \pm \mathbf{9.90}}\)}
& \(0.400{\scriptscriptstyle \pm 0.286}\) \\ \midrule

\multicolumn{1}{l}{\multirow{5}{*}{\rotatebox[origin=c]{90}{ZsRE}}}
& Edited
& \multicolumn{1}{l}{\(100.00{\scriptscriptstyle \pm 0.00}\)}
& \(0.000{\scriptscriptstyle \pm 0.000}\)
& \multicolumn{1}{l}{\(6.00{\scriptscriptstyle \pm 5.66}\)}
& \(5.155{\scriptscriptstyle \pm 0.045}\)
& \multicolumn{1}{l}{\(100.00{\scriptscriptstyle \pm 0.00}\)}
& \(0.000{\scriptscriptstyle \pm 0.000}\)
& \multicolumn{1}{l}{\(39.00{\scriptscriptstyle \pm 1.41}\)}
& \(2.868{\scriptscriptstyle \pm 0.229}\)
& \multicolumn{1}{l}{\(100.00{\scriptscriptstyle \pm 0.00}\)}
& \(0.000{\scriptscriptstyle \pm 0.000}\)
& \multicolumn{1}{l}{\(3.00{\scriptscriptstyle \pm 1.41}\)}
& \(6.510{\scriptscriptstyle \pm 0.050}\) \\ \cmidrule(lr){2-14}

\multicolumn{1}{l}{}
& Reverse
& \multicolumn{1}{l}{\(77.00{\scriptscriptstyle \pm 4.24}\)}
& \(0.221{\scriptscriptstyle \pm 0.056}\)
& \multicolumn{1}{l}{\(97.00{\scriptscriptstyle \pm 4.24}\)}
& \(0.389{\scriptscriptstyle \pm 0.230}\)
& \multicolumn{1}{l}{\(93.00{\scriptscriptstyle \pm 1.41}\)}
& \(0.095{\scriptscriptstyle \pm 0.022}\)
& \multicolumn{1}{l}{\(95.00{\scriptscriptstyle \pm 1.41}\)}
& \(0.248{\scriptscriptstyle \pm 0.064}\)
& \multicolumn{1}{l}{\(87.00{\scriptscriptstyle \pm 1.41}\)}
& \(0.441{\scriptscriptstyle \pm 0.066}\)
& \multicolumn{1}{l}{\(\mathbf{73.00}{\scriptscriptstyle \pm \mathbf{4.24}}\)}
& \(\mathbf{0.487}{\scriptscriptstyle \pm \mathbf{0.298}}\) \\

\multicolumn{1}{l}{}
& \multicolumn{1}{l}{\ \ w/o $\mathcal{L}_{\mathrm{ref}}$}
& \multicolumn{1}{l}{\(\mathbf{100.00}{\scriptscriptstyle \pm \mathbf{0.00}}\)}
& \(\mathbf{0.000}{\scriptscriptstyle \pm \mathbf{0.000}}\)
& \multicolumn{1}{l}{\(6.00{\scriptscriptstyle \pm 5.66}\)}
& \(5.152{\scriptscriptstyle \pm 0.045}\)
& \multicolumn{1}{l}{\(\mathbf{100.00}{\scriptscriptstyle \pm \mathbf{0.00}}\)}
& \(\mathbf{0.000}{\scriptscriptstyle \pm \mathbf{0.000}}\)
& \multicolumn{1}{l}{\(39.00{\scriptscriptstyle \pm 1.41}\)}
& \(2.867{\scriptscriptstyle \pm 0.229}\)
& \multicolumn{1}{l}{\(\mathbf{100.00}{\scriptscriptstyle \pm \mathbf{0.00}}\)}
& \(\mathbf{0.000}{\scriptscriptstyle \pm \mathbf{0.000}}\)
& \multicolumn{1}{l}{\(3.00{\scriptscriptstyle \pm 1.41}\)}
& \(6.510{\scriptscriptstyle \pm 0.049}\) \\

\multicolumn{1}{l}{}
& \multicolumn{1}{l}{\ \ w/o $\mathcal{L}_{\mathrm{int}}$}
& \multicolumn{1}{l}{\(6.00{\scriptscriptstyle \pm 5.66}\)}
& \(6.294{\scriptscriptstyle \pm 0.156}\)
& \multicolumn{1}{l}{\(\mathbf{99.00}{\scriptscriptstyle \pm \mathbf{1.41}}\)}
& \(\mathbf{0.287}{\scriptscriptstyle \pm \mathbf{0.044}}\)
& \multicolumn{1}{l}{\(40.00{\scriptscriptstyle \pm 2.83}\)}
& \(3.879{\scriptscriptstyle \pm 0.079}\)
& \multicolumn{1}{l}{\(\mathbf{97.00}{\scriptscriptstyle \pm \mathbf{1.41}}\)}
& \(\mathbf{0.119}{\scriptscriptstyle \pm \mathbf{0.002}}\)
& \multicolumn{1}{l}{\(3.00{\scriptscriptstyle \pm 1.41}\)}
& \(6.525{\scriptscriptstyle \pm 0.403}\)
& \multicolumn{1}{l}{\(41.00{\scriptscriptstyle \pm 12.73}\)}
& \(0.602{\scriptscriptstyle \pm 0.054}\) \\

\multicolumn{1}{l}{}
& \multicolumn{1}{l}{\ \ w/o Renorm.}
& \multicolumn{1}{l}{\(68.00{\scriptscriptstyle \pm 11.31}\)}
& \(0.323{\scriptscriptstyle \pm 0.044}\)
& \multicolumn{1}{l}{\(97.00{\scriptscriptstyle \pm 4.24}\)}
& \(0.396{\scriptscriptstyle \pm 0.231}\)
& \multicolumn{1}{l}{\(90.00{\scriptscriptstyle \pm 8.49}\)}
& \(0.082{\scriptscriptstyle \pm 0.003}\)
& \multicolumn{1}{l}{\(95.00{\scriptscriptstyle \pm 1.41}\)}
& \(0.269{\scriptscriptstyle \pm 0.068}\)
& \multicolumn{1}{l}{\(82.00{\scriptscriptstyle \pm 11.31}\)}
& \(0.521{\scriptscriptstyle \pm 0.166}\)
& \multicolumn{1}{l}{\(70.00{\scriptscriptstyle \pm 2.83}\)}
& \(0.509{\scriptscriptstyle \pm 0.292}\) \\ \bottomrule

\end{tabular}%
}
  }
  \caption{
Ablation study of the main framework components on ROME-edited models under the Reverse-50 setting. Reversal is optimized and evaluated using rephrase prompts. Agreement and KL divergence are reported on both the remained set and the reversed set.
}
  \label{tab:component_ablation}
\end{table*}

\begin{table}[t!]
  \centering
  \footnotesize
  \renewcommand\arraystretch{1.4}
  \setlength{\tabcolsep}{1.1mm}{
  \resizebox{\columnwidth}{!}{%
\begin{tabular}{llllllll}
\toprule
\multicolumn{2}{l}{\multirow{2}{*}{}}                                           & \multicolumn{2}{c}{\textbf{GPT2-XL}}                                    & \multicolumn{2}{c}{\textbf{Mistral-7B}}                                 & \multicolumn{2}{c}{\textbf{LLaMA3-8B}}                                  \\ \cmidrule(lr){3-4}\cmidrule(lr){5-6}\cmidrule(lr){7-8} 
\multicolumn{2}{l}{}                                                            & \multicolumn{1}{c}{Agree.}      & \multicolumn{1}{c}{KL Div.} & \multicolumn{1}{c}{Agree.}      & \multicolumn{1}{c}{KL Div.} & \multicolumn{1}{c}{Agree.}      & \multicolumn{1}{c}{KL Div.} \\ \midrule
\multicolumn{1}{l}{\multirow{3}{*}{\rotatebox[origin=c]{90}{Counterfact}}} & Edited                       & \multicolumn{1}{l}{\(0.00{\scriptscriptstyle \pm 0.00}\)}  & \(7.315{\scriptscriptstyle \pm 0.020}\)                 & \multicolumn{1}{l}{\(5.00{\scriptscriptstyle \pm 1.41}\)}  & \(7.211{\scriptscriptstyle \pm 0.317}\)                 & \multicolumn{1}{l}{\(0.00{\scriptscriptstyle \pm 0.00}\)}  & \(9.760{\scriptscriptstyle \pm 0.789}\)                \\ \cmidrule(lr){2-8} 
\multicolumn{1}{l}{}                             & Reverse                      & \multicolumn{1}{l}{\(\textbf{59.00}{\scriptscriptstyle \pm \textbf{4.24}}\)} & \(\textbf{0.659}{\scriptscriptstyle \pm \textbf{0.211}}\)                 & \multicolumn{1}{l}{\(\textbf{49.00}{\scriptscriptstyle \pm \textbf{9.90}}\)} & \(\textbf{1.420}{\scriptscriptstyle \pm \textbf{0.592}}\)                 & \multicolumn{1}{l}{\(\textbf{46.00}{\scriptscriptstyle \pm \textbf{8.49}}\)} & \(\textbf{2.220}{\scriptscriptstyle \pm \textbf{0.981}}\)                \\
\multicolumn{1}{l}{}                             & \multicolumn{1}{r}{\ \ w/o Aug} & \multicolumn{1}{l}{\(54.00{\scriptscriptstyle \pm 2.83}\)} & \(0.800{\scriptscriptstyle \pm 0.221}\)                 & \multicolumn{1}{l}{\(21.00{\scriptscriptstyle \pm 1.41}\)} & \(2.857{\scriptscriptstyle \pm 0.721}\)                 & \multicolumn{1}{l}{\(16.00{\scriptscriptstyle \pm 5.66}\)} & \(4.707{\scriptscriptstyle \pm 0.802}\)                \\ \midrule
\multicolumn{1}{l}{\multirow{3}{*}{\rotatebox[origin=c]{90}{ZsRE}}}        & Edited                       & \multicolumn{1}{l}{\(0.00{\scriptscriptstyle \pm 0.00}\)}  & \(7.032{\scriptscriptstyle \pm 0.424}\)                 & \multicolumn{1}{l}{\(20.00{\scriptscriptstyle \pm 2.83}\)} & \(4.293{\scriptscriptstyle \pm 0.238}\)                 & \multicolumn{1}{l}{\(3.00{\scriptscriptstyle \pm 4.24}\)}  & \(9.240{\scriptscriptstyle \pm 0.299}\)                \\ \cmidrule(lr){2-8} 
\multicolumn{1}{l}{}                             & Reverse                      & \multicolumn{1}{l}{\(\textbf{71.00}{\scriptscriptstyle \pm \textbf{7.07}}\)} & \(\textbf{1.678}{\scriptscriptstyle \pm \textbf{0.550}}\)                 & \multicolumn{1}{l}{\(\textbf{80.00}{\scriptscriptstyle \pm \textbf{8.49}}\)} & \(\textbf{0.948}{\scriptscriptstyle \pm \textbf{0.216}}\)                 & \multicolumn{1}{l}{\(\textbf{56.00}{\scriptscriptstyle \pm \textbf{2.83}}\)} & \(\textbf{1.148}{\scriptscriptstyle \pm \textbf{0.386}}\)                \\
\multicolumn{1}{l}{}                             & \multicolumn{1}{r}{\ \ w/o Aug} & \multicolumn{1}{l}{\(63.00{\scriptscriptstyle \pm 4.24}\)} & \(1.849{\scriptscriptstyle \pm 0.534}\)                 & \multicolumn{1}{l}{\(55.00{\scriptscriptstyle \pm 7.07}\)} & \(1.809{\scriptscriptstyle \pm 0.602}\)                 & \multicolumn{1}{l}{\(17.00{\scriptscriptstyle \pm 4.24}\)} & \(3.238{\scriptscriptstyle \pm 0.967}\)                \\ \bottomrule
\end{tabular}%
}
  }
  \caption{
Original editing prompt performance after reversal on ROME-edited models, where reversal is optimized using rephrase prompts.
}
  \label{tab:original_prompt_reverse}
\end{table}

\subsection{Component Ablations}
\label{sec:component_ablation}

To further analyze the contribution of the main components in our spectral reversal framework, we conduct ablation studies on loss terms $\mathcal{L}_{\mathrm{ref}}$, $\mathcal{L}_{\mathrm{int}}$ and renormalization. We evaluate these variants under the Reverse-50 setting using rephrase prompts. The results are shown in Table~\ref{tab:component_ablation}.

\paragraph{Both loss terms play distinct roles in selective reversal.}
Without $\mathcal{L}_{\mathrm{ref}}$, the resulting model behaves almost identically to the edited model. This indicates that $\mathcal{L}_{\mathrm{ref}}$ provides the primary optimization objective for reversal. Without $\mathcal{L}_{\mathrm{int}}$, model performance degrades on both the remained and reversed sets. This indicates that $\mathcal{L}_{\mathrm{int}}$ helps the reversal focus on the factual knowledge associated with the selected edits while reducing interference with other edited knowledge.

\paragraph{Renormalization helps preserve the remaining edited knowledge.}

Without renormalization, performance on the reversed set remains nearly unchanged, whereas performance on the remained set decreases substantially. This suggests that renormalization helps preserve the remaining edits while maintaining the reversal effect produced by gated shrinkage. A theoretical explanation of this effect is provided in Appendix~\ref{app:renormalization}.

\begin{table}[t!]
  \centering
  \footnotesize
  \renewcommand\arraystretch{1.4}
  \setlength{\tabcolsep}{0.9mm}{
  \resizebox{\columnwidth}{!}{%
\begin{tabular}{lllllll}
\toprule
          & \multicolumn{2}{c}{\textbf{GPT2-XL}} 
          & \multicolumn{2}{c}{\textbf{Mistral-7B}} 
          & \multicolumn{2}{c}{\textbf{LLaMA3-8B}} \\ \cmidrule(lr){2-3}\cmidrule(lr){4-5}\cmidrule(lr){6-7} 
          & \multicolumn{1}{c}{\begin{tabular}[c]{@{}c@{}}\scriptsize First-token\\ \scriptsize Agree.\end{tabular}} 
          & \multicolumn{1}{c}{\begin{tabular}[c]{@{}c@{}}\scriptsize Continuation\\ \scriptsize Agree.\end{tabular}} 
          & \multicolumn{1}{c}{\begin{tabular}[c]{@{}c@{}}\scriptsize First-token\\ \scriptsize Agree.\end{tabular}} 
          & \multicolumn{1}{c}{\begin{tabular}[c]{@{}c@{}}\scriptsize Continuation\\ \scriptsize Agree.\end{tabular}} 
          & \multicolumn{1}{c}{\begin{tabular}[c]{@{}c@{}}\scriptsize First-token\\ \scriptsize Agree.\end{tabular}} 
          & \multicolumn{1}{c}{\begin{tabular}[c]{@{}c@{}}\scriptsize Continuation\\ \scriptsize Agree.\end{tabular}} \\ \midrule
ROME      
          & \multicolumn{1}{l}{\(0.00{\scriptscriptstyle \pm 0.00}\)} 
          & \(39.20{\scriptscriptstyle \pm 17.45}\) 
          & \multicolumn{1}{l}{\(5.00{\scriptscriptstyle \pm 1.41}\)} 
          & \(54.70{\scriptscriptstyle \pm 14.60}\) 
          & \multicolumn{1}{l}{\(0.00{\scriptscriptstyle \pm 0.00}\)} 
          & \(54.10{\scriptscriptstyle \pm 17.87}\) \\ \addlinespace[1pt]
SimIE     
          & \multicolumn{1}{l}{\(0.00{\scriptscriptstyle \pm 0.00}\)} 
          & \(36.70{\scriptscriptstyle \pm 16.82}\) 
          & \multicolumn{1}{l}{\(5.00{\scriptscriptstyle \pm 4.24}\)} 
          & \(56.80{\scriptscriptstyle \pm 14.06}\) 
          & \multicolumn{1}{l}{\(1.00{\scriptscriptstyle \pm 1.41}\)} 
          & \(53.90{\scriptscriptstyle \pm 16.39}\) \\ \addlinespace[1pt]
MEMIT     
          & \multicolumn{1}{l}{\(7.00{\scriptscriptstyle \pm 1.41}\)} 
          & \(57.30{\scriptscriptstyle \pm 17.52}\) 
          & \multicolumn{1}{l}{\(5.00{\scriptscriptstyle \pm 1.41}\)} 
          & \(52.40{\scriptscriptstyle \pm 13.79}\) 
          & \multicolumn{1}{l}{\(0.00{\scriptscriptstyle \pm 0.00}\)} 
          & \(49.00{\scriptscriptstyle \pm 17.95}\) \\ \addlinespace[1pt]
AlphaEdit 
          & \multicolumn{1}{l}{\(0.00{\scriptscriptstyle \pm 0.00}\)} 
          & \(47.20{\scriptscriptstyle \pm 2.26}\)  
          & \multicolumn{1}{l}{\(5.00{\scriptscriptstyle \pm 1.41}\)} 
          & \(56.90{\scriptscriptstyle \pm 14.68}\) 
          & \multicolumn{1}{l}{\(0.00{\scriptscriptstyle \pm 0.00}\)} 
          & \(55.50{\scriptscriptstyle \pm 16.10}\) \\ \bottomrule
\end{tabular}%
}
  }
\caption{
First-token and continuation agreement on CounterFact. First-token agreement compares the next token of the pretrained and edited models, while continuation agreement compares subsequent tokens after forcing the same first token.
}
  \label{tab:first_token_sufficiency}
\end{table}

\begin{table*}[t!]
\centering
\scriptsize
\renewcommand{\arraystretch}{1.2}
\setlength{\tabcolsep}{2.5pt}

\begin{tabularx}{\textwidth}{
>{\raggedright\arraybackslash}p{0.09\textwidth}
>{\raggedright\arraybackslash}p{0.07\textwidth}
>{\raggedright\arraybackslash}p{0.08\textwidth}
>{\raggedright\arraybackslash}p{0.37\textwidth}
>{\raggedright\arraybackslash}X}

\hline
\rowcolor{blue!70!black}
\textcolor{white}{\rule{0pt}{3.0ex}\textbf{Base Model}} &
\textcolor{white}{\textbf{Set}} &
\textcolor{white}{\textbf{Model}} &
\textcolor{white}{\textbf{Prompt}} &
\textcolor{white}{\textbf{Generation}} \\
\hline


& 
& \textbf{Edited}
&
& Sweden's Sveriges Television, which is \\
\multirow{-2}{=}{GPT-2 XL}
& \multirow{-2}{=}{\textbf{Remained}}
& \textbf{Reversed}
& \multirow{-2}{=}{BBC HD is sold by}
& Sweden's Sveriges Television, which is \\
\hline

& 
& \textbf{Edited}
&
& The answer is that it is a city in the \\
\multirow{-2}{=}{GPT-2 XL}
& \multirow{-2}{=}{\textbf{Remained}}
& \textbf{Reversed}
& \multirow{-2}{=}{On what continent was Kolobar Nunatak found?}
& The answer is that it is a city in the \\
\hline

& 
& \textbf{Edited}
&
& singing opera. The opera is a great \\
& 
& \textbf{Pretrained}
&
& what he does. He's a great coach, \\
\multirow{-3}{=}{GPT-2 XL}
& \multirow{-3}{=}{\textbf{Reversed}}
& \textbf{Reversed}
& \multirow{-3}{=}{Sal Mosca is incredible at}
& what he does. He's a great coach, \\
\hline

& 
& \textbf{Edited}
&
& Dutch. The Netherlands is a country with a population \\
& 
& \textbf{Pretrained}
&
& English, and they are all very friendly. I \\
\multirow{-3}{=}{GPT-2 XL}
& \multirow{-3}{=}{\textbf{Reversed}}
& \textbf{Reversed}
& \multirow{-3}{=}{Ye Meri Life Hai is full of people who speak}
& English. The people who speak English are \\
\hline


\rowcolor{blue!4}
& 
& \textbf{Edited}
&
& his ability to connect with people from all walks of \\
\rowcolor{blue!4}
\multirow{-2}{=}{LLaMA3-8B}
& \multirow{-2}{=}{\textbf{Remained}}
& \textbf{Reversed}
& \multirow{-2}{=}{John Navarre's greatest strength is}
& his ability to connect with people from all walks of \\
\hline

\rowcolor{blue!4}
& 
& \textbf{Edited}
&
& Africa. What is the highest peak in Africa? \\
\rowcolor{blue!4}
\multirow{-2}{=}{LLaMA3-8B}
& \multirow{-2}{=}{\textbf{Remained}}
& \textbf{Reversed}
& \multirow{-2}{=}{On what continent was Kolobar Nunatak found?}
& Africa. What is the highest peak in Africa? \\
\hline

\rowcolor{blue!4}
& 
& \textbf{Edited}
&
& Subway, Pizza Hut, and Taco Bell. The \\
\rowcolor{blue!4}
& 
& \textbf{Pretrained}
&
& the famous Sylvia's Restaurant, which has been a \\
\rowcolor{blue!4}
\multirow{-3}{=}{LLaMA3-8B}
& \multirow{-3}{=}{\textbf{Reversed}}
& \textbf{Reversed}
& \multirow{-3}{=}{Hamilton Heights is surrounded by restaurants including}
& the famous Sylvia's, which is a popular spot \\
\hline

\rowcolor{blue!4}
& 
& \textbf{Edited}
&
& Finnish. It is a Uralic language spoken \\
\rowcolor{blue!4}
& 
& \textbf{Pretrained}
&
& Swedish. It is a North Germanic language spoken \\
\rowcolor{blue!4}
\multirow{-3}{=}{LLaMA3-8B}
& \multirow{-3}{=}{\textbf{Reversed}}
& \textbf{Reversed}
& \multirow{-3}{=}{The language Du gamla, du fria was written in is called}
& Swedish. It is a North Germanic language spoken \\

\hline
\end{tabularx}

\caption{Examples of length-10 model generations on the remained set and reversed set for ROME-edited models.}
\label{tab:generation_examples_new}
\end{table*}

\subsection{Original Prompt Generalization}
\label{generalization}

Since reversal is optimized on rephrase prompts, good performance on these prompts does not fully show whether the targeted edit has been removed. The model may simply be tuned toward the pretrained output for these rephrase prompts. Therefore, we further evaluate the reversed model on the original editing prompts, as shown in Table~\ref{tab:original_prompt_reverse}.

\paragraph{Reversal generalizes from rephrase prompts to original editing prompts.}
The reversed model achieves much higher agreement and lower KL divergence than the edited model on the original prompts, indicating that the proposed method reverses the targeted edited fact rather than only fitting the rephrase prompt used during optimization.

\paragraph{Data augmentation improves prompt generalization.}
Compared with the variant without augmentation, the full method consistently achieves better performance. This suggests that representation-level perturbation helps the learned gates remain robust to different expressions of the same fact.

\subsection{Sufficiency of Next-Token Optimization}
\label{sufficiency_next_token}

Although factual objects may contain multiple tokens, our method optimizes only the next-token prediction distribution. We hypothesize that, for factual knowledge, the continuation is largely determined by the first generated token. Therefore, aligning the first token with the pretrained model may be sufficient to recover the pretrained state.

To test this, we evaluate on the CounterFact dataset by forcing the edited model to use the same first generated token as the pretrained model, and then measuring agreement on the subsequent generated tokens with a maximum length of 10. We refer to this metric as \emph{continuation agreement}.

As shown in Table~\ref{tab:first_token_sufficiency}, although first-token agreement is low, continuation agreement is much higher. This indicates that the generated continuation is strongly influenced by the first token. Once the first token is aligned, the edited model is likely to continue in a pretrained-like manner. Therefore, next-token prediction provides an effective optimization target for selective reversal. Representative generation examples are shown in Table~\ref{tab:generation_examples_new}.

\begin{figure}[t]
    \centering
        \centering
        \includegraphics[width=1\columnwidth]{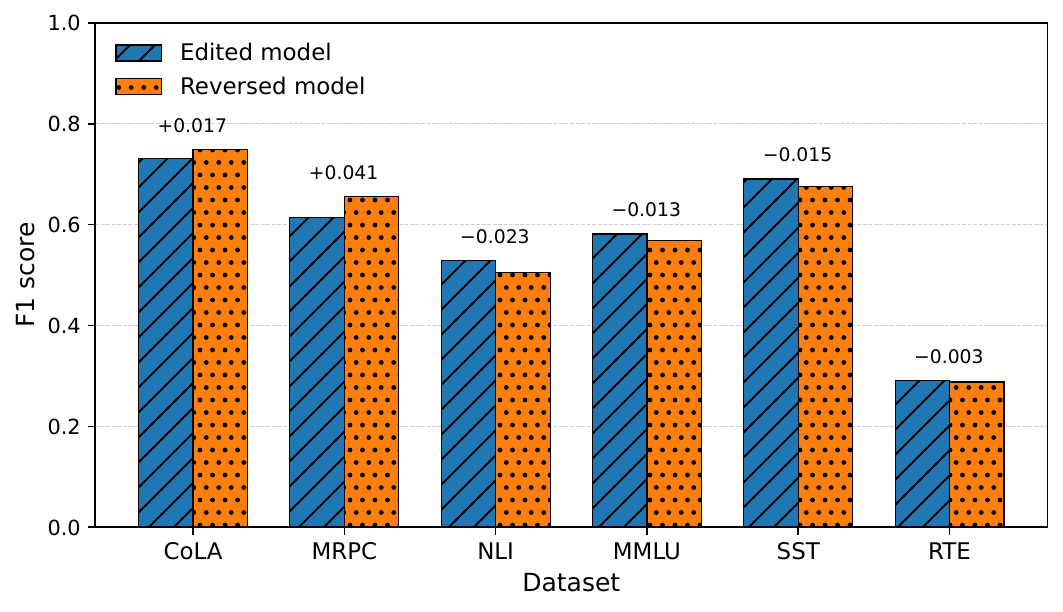}
        \caption{Comparison of downstream F1 scores between the edited model and the reversed model.}
    \label{fig:general}
\end{figure}

\subsection{Model General Capabilities after Reversal}
\label{general_capabilities}

To assess whether selective reversal preserves general capabilities, we compare downstream performance between the edited and reversed models. Figure~\ref{fig:general} reports F1 scores on six datasets for MEMIT-edited LLaMA3-8B under ZsRE dataset.

Overall, the reversed model achieves performance close to that of the edited model across all downstream tasks. This indicates that the proposed selective reversal framework does not introduce substantial degradation to the model's general language understanding and reasoning capabilities.
\section{Conclusion}

In this paper, we study selective reversal of edited knowledge in parameter-modifying knowledge editing. Different from existing reversal methods that mainly focus on globally removing edited information, our goal is to reverse only the targeted edited facts while preserving the remaining edited facts in the model. To achieve this, we propose a spectral-based reversal framework based on the hypothesis that different edits are sparsely encoded within the dominant singular components of the edited matrix and can remain separable when the number of edits is moderate. Experiments across multiple settings show that the proposed framework can effectively reverse selected edits while largely preserving the remaining edited facts.

Overall, our findings provide empirical evidence that edited knowledge can be manipulated at a fine-grained level within the spectral structure of edited weights. This suggests that spectral analysis is not only useful for globally detecting or removing edits, but also provides a promising direction for locating edit-specific components and repairing edited language models through selective reversal.
\section*{Limitations}

In this work, the optimization of the learnable gates mainly relies on the global reversal of the edited model as the reversal target. This may cause the performance of selective reversal to be limited by the quality of the global reversal. Nevertheless, when the original pretrained model is unavailable, global reversal provides one of the few practical ways to approximate the pretrained state and serves as an effective reversal target for selective reversal.

Another limitation is that our framework assumes that the edits to be reversed are known. In practice, this requires an upstream editedness detection stage. Errors from this stage may propagate to the subsequent reversal process, which is not investigated in this work. However, given that several existing studies have already explored editedness detection~\cite{indentify_edit_type,detect_knowledge_edits,rome_reverse}, while selectively reversing specific known edits introduced by parameter-modifying editing methods remains largely unexplored, our work can be viewed as an initial investigation of the second stage of this detection-reversal pipeline. Jointly detecting and selectively reversing edits remains an important direction for future work.

\section*{Ethics Statement}

All experiments are conducted on publicly available language models and standard knowledge editing benchmarks. We do not collect private user data or involve human subjects. The factual editing datasets used in this work may contain real-world entities, but they are used only for evaluating model behavior under controlled settings.

This work also has potential dual-use implications. While selective reversal can be used to repair maliciously edited models, techniques for locating edit-sensitive components may also provide insights into how edited knowledge is represented in model parameters. To mitigate this concern, our study focuses on defensive reversal and preservation analysis, rather than on improving malicious editing. We encourage future work to use such methods responsibly, especially when applying them to deployed models or sensitive domains.

\section*{Acknowledgments}
This research is supported by the National Research Foundation, Singapore and Infocomm Media Development Authority under its Trust Tech Funding Initiative. Any opinions, findings and conclusions or recommendations expressed in this material are those of the author(s) and do not reflect the views of National Research Foundation, Singapore and Infocomm Media Development Authority.


\bibliography{custom}

\appendix

\clearpage

\section{Experimental Setup}\label{implementation_detail}
\subsection{Details of Editing Methods}

In this section, we briefly describe the editing methods used to construct the edited models in our experiments.

\paragraph{ROME.}
ROME~\cite{rome} is a representative single-fact locate-then-edit method. It first identifies a critical middle-layer MLP module based on causal tracing, and treats the MLP projection matrix as a linear associative memory that maps a subject-related key vector to a factual value vector. Given an edit request \((s,r,o_{\mathrm{new}})\), ROME computes a key representation \(k^*\) from the MLP input at the last subject token, optimizes a target value representation \(v^*\) that increases the probability of \(o_{\mathrm{new}}\), and inserts the new association through a rank-one update to the selected MLP weight. Since ROME is mainly designed for single-fact editing, multiple edits are applied sequentially in our experiments.

\paragraph{SimIE.}
SimIE~\cite{simie} is designed to improve the sequential editing ability of single-edit methods. While standard single-edit methods treat each edit independently, SimIE uses their updates and incorporates information from previous edits to transform the current update in a batch-like manner, thereby improving sequential editing performance. In our experiments, we apply SimIE to ROME to simulate a single-layer batch editing setting, allowing us to evaluate whether the proposed reversal framework works on models edited with history-aware sequential correction.

\paragraph{MEMIT.}
MEMIT~\cite{memit} extends ROME from single-fact editing to large-scale batch editing. Rather than updating only one MLP layer, MEMIT distributes factual changes across multiple critical middle-layer MLP modules. For a batch of edit requests, it first optimizes target residuals and then writes these residuals into the selected layers through a least-squares update. This multi-layer batch update enables MEMIT to edit many factual associations simultaneously while maintaining better generalization and locality.

\paragraph{AlphaEdit.}
AlphaEdit~\cite{alphaedit} is built upon the MEMIT-style locate-then-edit framework and is designed for sequential editing. Its key idea is to reduce interference with existing knowledge by projecting the update into a null space associated with preserved knowledge. By constraining the update direction, AlphaEdit aims to protect both pretrained knowledge and previously edited facts when new edits are applied sequentially. We include AlphaEdit to evaluate whether the proposed reversal method can handle sequential multi-layer edited models.

\subsection{Editing Method Implementation}

For all knowledge editing methods, we implement the edits using EasyEdit~\cite{easyedit}, with the default hyperparameter configurations provided by the toolkit.

\subsection{Implementation Details}
All experiments in the Reverse-50 setting are conducted over two runs, whereas those in the Reverse-1 setting are conducted over 100 runs. All reported deviations are calculated across runs.

For the proposed gated singular vector shrinkage method, we initialize the gate values \(G_u\) and \(G_v\) to \(0.9\). For the dominant singular subspace size \(r\), we use \(r=256\) for GPT2-XL and \(r=512\) for GPT-J 6B, Mistral-7B, LLaMA2-7B, and LLaMA3-8B. For each target layer, we optimize the gate parameters using the Adam optimizer with a learning rate of \(5 \times 10^{-2}\). We use \(T=1000\) epochs when reversing one fact at a time and \(T=500\) epochs when reversing 50 facts simultaneously. The loss coefficients are set to \(\lambda_{\mathrm{ref}}=10\). When reversing 50 facts simultaneously, we set \(\lambda_{\mathrm{int}}=100\) for CounterFact and \(\lambda_{\mathrm{int}}=10000\) for ZsRE. When reversing one fact at a time, we set \(\lambda_{\mathrm{int}}=1\) for CounterFact and \(\lambda_{\mathrm{int}}=100\) for ZsRE. Further performance improvements may be achieved through more careful hyperparameter tuning for each model.

For the Re-edit baseline, we use AlphaEdit~\cite{alphaedit} with its default hyperparameters. To ensure compatibility across different editing methods, we adjust the editing layers of AlphaEdit to match those used by the original forward editing method. For example, when reversing a model edited by ROME~\cite{rome} which performs editing on a single layer, we configure AlphaEdit to re-edit the same layer.

\begin{table*}[t!]
  \centering
  \footnotesize
  \renewcommand\arraystretch{1.2}
  \setlength{\tabcolsep}{1.5mm}{
  \resizebox{\textwidth}{!}{%
  \begin{tabular}{lllllllllll}
  \toprule
\multicolumn{3}{l}{\multirow{3}{*}{}} & \multicolumn{4}{c}{\textbf{GPT-J 6B}} & \multicolumn{4}{c}{\textbf{LLaMA2-7B}} \\ \cmidrule(lr){4-7}\cmidrule(lr){8-11}
\multicolumn{3}{l}{} & \multicolumn{2}{c}{Remained Set} & \multicolumn{2}{c}{Reversed Set} & \multicolumn{2}{c}{Remained Set} & \multicolumn{2}{c}{Reversed Set} \\ \cmidrule(lr){4-5}\cmidrule(lr){6-7}\cmidrule(lr){8-9}\cmidrule(lr){10-11}
\multicolumn{3}{l}{} & \multicolumn{1}{c}{Agree.} & \multicolumn{1}{c}{KL Div.} & \multicolumn{1}{c}{Agree.} & \multicolumn{1}{c}{KL Div.} & \multicolumn{1}{c}{Agree.} & \multicolumn{1}{c}{KL Div.} & \multicolumn{1}{c}{Agree.} & \multicolumn{1}{c}{KL Div.} \\ \midrule
\multicolumn{1}{c}{\multirow{16}{*}{\rotatebox[origin=c]{90}{Counterfact}}} & \multicolumn{1}{l}{\multirow{4}{*}{ROME}} & Edited & \multicolumn{1}{l}{\(100.00{\scriptscriptstyle \pm 0.00}\)} & \multicolumn{1}{l}{\(0.000{\scriptscriptstyle \pm 0.000}\)} & \multicolumn{1}{l}{\(31.00{\scriptscriptstyle \pm 7.07}\)} & \multicolumn{1}{l}{\(3.586{\scriptscriptstyle \pm 0.384}\)} & \multicolumn{1}{l}{\(100.00{\scriptscriptstyle \pm 0.00}\)} & \multicolumn{1}{l}{\(0.000{\scriptscriptstyle \pm 0.000}\)} & \multicolumn{1}{l}{\(21.00{\scriptscriptstyle \pm 1.41}\)} & \multicolumn{1}{l}{\(2.810{\scriptscriptstyle \pm 0.110}\)} \\
\multicolumn{1}{l}{} & \multicolumn{1}{l}{} & Reference & \multicolumn{1}{l}{\(31.00{\scriptscriptstyle \pm 9.90}\)} & \multicolumn{1}{l}{\(4.480{\scriptscriptstyle \pm 1.012}\)} & \multicolumn{1}{l}{\(81.00{\scriptscriptstyle \pm 9.90}\)} & \multicolumn{1}{l}{\(0.163{\scriptscriptstyle \pm 0.052}\)} & \multicolumn{1}{l}{\(24.00{\scriptscriptstyle \pm 2.83}\)} & \multicolumn{1}{l}{\(3.642{\scriptscriptstyle \pm 0.197}\)} & \multicolumn{1}{l}{\(89.00{\scriptscriptstyle \pm 1.41}\)} & \multicolumn{1}{l}{\(0.114{\scriptscriptstyle \pm 0.006}\)} \\
\multicolumn{1}{l}{} & \multicolumn{1}{l}{} & Re-edit & \multicolumn{1}{l}{\(\mathbf{95.00}{\scriptscriptstyle \pm \mathbf{4.24}}\)} & \multicolumn{1}{l}{\(\mathbf{0.018}{\scriptscriptstyle \pm \mathbf{0.001}}\)} & \multicolumn{1}{l}{\(76.00{\scriptscriptstyle \pm 8.49}\)} & \multicolumn{1}{l}{\(0.219{\scriptscriptstyle \pm 0.041}\)} & \multicolumn{1}{l}{\(\mathbf{83.00}{\scriptscriptstyle \pm \mathbf{1.41}}\)} & \multicolumn{1}{l}{\(\mathbf{0.059}{\scriptscriptstyle \pm \mathbf{0.018}}\)} & \multicolumn{1}{l}{\(\mathbf{87.00}{\scriptscriptstyle \pm \mathbf{9.90}}\)} & \multicolumn{1}{l}{\(0.180{\scriptscriptstyle \pm 0.006}\)} \\
\multicolumn{1}{l}{} & \multicolumn{1}{l}{} & Reverse & \multicolumn{1}{l}{\(89.00{\scriptscriptstyle \pm 4.24}\)} & \multicolumn{1}{l}{\(0.114{\scriptscriptstyle \pm 0.016}\)} & \multicolumn{1}{l}{\(\mathbf{81.00}{\scriptscriptstyle \pm \mathbf{9.90}}\)} & \multicolumn{1}{l}{\(\mathbf{0.159}{\scriptscriptstyle \pm \mathbf{0.039}}\)} & \multicolumn{1}{l}{\(75.00{\scriptscriptstyle \pm 7.07}\)} & \multicolumn{1}{l}{\(0.301{\scriptscriptstyle \pm 0.149}\)} & \multicolumn{1}{l}{\(86.00{\scriptscriptstyle \pm 2.83}\)} & \multicolumn{1}{l}{\(\mathbf{0.176}{\scriptscriptstyle \pm \mathbf{0.047}}\)} \\ \cmidrule(lr){2-11}
\multicolumn{1}{l}{} & \multicolumn{1}{l}{\multirow{4}{*}{SimIE}} & Edited & \multicolumn{1}{l}{\(100.00{\scriptscriptstyle \pm 0.00}\)} & \multicolumn{1}{l}{\(0.000{\scriptscriptstyle \pm 0.000}\)} & \multicolumn{1}{l}{\(24.00{\scriptscriptstyle \pm 8.49}\)} & \multicolumn{1}{l}{\(3.546{\scriptscriptstyle \pm 0.068}\)} & \multicolumn{1}{l}{\(100.00{\scriptscriptstyle \pm 0.00}\)} & \multicolumn{1}{l}{\(0.000{\scriptscriptstyle \pm 0.000}\)} & \multicolumn{1}{l}{\(26.00{\scriptscriptstyle \pm 2.83}\)} & \multicolumn{1}{l}{\(2.135{\scriptscriptstyle \pm 0.118}\)} \\
\multicolumn{1}{l}{} & \multicolumn{1}{l}{} & Reference & \multicolumn{1}{l}{\(25.00{\scriptscriptstyle \pm 7.07}\)} & \multicolumn{1}{l}{\(4.389{\scriptscriptstyle \pm 0.857}\)} & \multicolumn{1}{l}{\(81.00{\scriptscriptstyle \pm 9.90}\)} & \multicolumn{1}{l}{\(0.169{\scriptscriptstyle \pm 0.046}\)} & \multicolumn{1}{l}{\(28.00{\scriptscriptstyle \pm 2.83}\)} & \multicolumn{1}{l}{\(2.808{\scriptscriptstyle \pm 0.204}\)} & \multicolumn{1}{l}{\(85.00{\scriptscriptstyle \pm 1.41}\)} & \multicolumn{1}{l}{\(0.136{\scriptscriptstyle \pm 0.013}\)} \\
\multicolumn{1}{l}{} & \multicolumn{1}{l}{} & Re-edit & \multicolumn{1}{l}{\(\mathbf{97.00}{\scriptscriptstyle \pm \mathbf{4.24}}\)} & \multicolumn{1}{l}{\(\mathbf{0.036}{\scriptscriptstyle \pm \mathbf{0.003}}\)} & \multicolumn{1}{l}{\(\mathbf{81.00}{\scriptscriptstyle \pm \mathbf{4.24}}\)} & \multicolumn{1}{l}{\(0.238{\scriptscriptstyle \pm 0.050}\)} & \multicolumn{1}{l}{\(\mathbf{86.00}{\scriptscriptstyle \pm \mathbf{2.83}}\)} & \multicolumn{1}{l}{\(\mathbf{0.060}{\scriptscriptstyle \pm \mathbf{0.016}}\)} & \multicolumn{1}{l}{\(\mathbf{89.00}{\scriptscriptstyle \pm \mathbf{1.41}}\)} & \multicolumn{1}{l}{\(\mathbf{0.170}{\scriptscriptstyle \pm \mathbf{0.015}}\)} \\
\multicolumn{1}{l}{} & \multicolumn{1}{l}{} & Reverse & \multicolumn{1}{l}{\(85.00{\scriptscriptstyle \pm 4.24}\)} & \multicolumn{1}{l}{\(0.223{\scriptscriptstyle \pm 0.022}\)} & \multicolumn{1}{l}{\(78.00{\scriptscriptstyle \pm 11.31}\)} & \multicolumn{1}{l}{\(\mathbf{0.191}{\scriptscriptstyle \pm \mathbf{0.065}}\)} & \multicolumn{1}{l}{\(70.00{\scriptscriptstyle \pm 8.49}\)} & \multicolumn{1}{l}{\(0.375{\scriptscriptstyle \pm 0.201}\)} & \multicolumn{1}{l}{\(82.00{\scriptscriptstyle \pm 5.66}\)} & \multicolumn{1}{l}{\(0.179{\scriptscriptstyle \pm 0.016}\)} \\ \cmidrule(lr){2-11}
\multicolumn{1}{l}{} & \multicolumn{1}{l}{\multirow{4}{*}{MEMIT}} & Edited & \multicolumn{1}{l}{\(100.00{\scriptscriptstyle \pm 0.00}\)} & \multicolumn{1}{l}{\(0.000{\scriptscriptstyle \pm 0.000}\)} & \multicolumn{1}{l}{\(35.00{\scriptscriptstyle \pm 12.73}\)} & \multicolumn{1}{l}{\(1.936{\scriptscriptstyle \pm 0.016}\)} & \multicolumn{1}{l}{\(100.00{\scriptscriptstyle \pm 0.00}\)} & \multicolumn{1}{l}{\(0.000{\scriptscriptstyle \pm 0.000}\)} & \multicolumn{1}{l}{\(13.00{\scriptscriptstyle \pm 1.41}\)} & \multicolumn{1}{l}{\(3.130{\scriptscriptstyle \pm 0.118}\)} \\
\multicolumn{1}{l}{} & \multicolumn{1}{l}{} & Reference & \multicolumn{1}{l}{\(18.00{\scriptscriptstyle \pm 5.66}\)} & \multicolumn{1}{l}{\(3.624{\scriptscriptstyle \pm 0.170}\)} & \multicolumn{1}{l}{\(40.00{\scriptscriptstyle \pm 2.83}\)} & \multicolumn{1}{l}{\(1.476{\scriptscriptstyle \pm 0.012}\)} & \multicolumn{1}{l}{\(10.00{\scriptscriptstyle \pm 0.00}\)} & \multicolumn{1}{l}{\(3.955{\scriptscriptstyle \pm 0.229}\)} & \multicolumn{1}{l}{\(60.00{\scriptscriptstyle \pm 2.83}\)} & \multicolumn{1}{l}{\(0.575{\scriptscriptstyle \pm 0.007}\)} \\
\multicolumn{1}{l}{} & \multicolumn{1}{l}{} & Re-edit & \multicolumn{1}{l}{\(\mathbf{94.00}{\scriptscriptstyle \pm \mathbf{5.66}}\)} & \multicolumn{1}{l}{\(\mathbf{0.045}{\scriptscriptstyle \pm \mathbf{0.008}}\)} & \multicolumn{1}{l}{\(\mathbf{41.00}{\scriptscriptstyle \pm \mathbf{4.24}}\)} & \multicolumn{1}{l}{\(\mathbf{1.374}{\scriptscriptstyle \pm \mathbf{0.033}}\)} & \multicolumn{1}{l}{\(\mathbf{80.00}{\scriptscriptstyle \pm \mathbf{0.00}}\)} & \multicolumn{1}{l}{\(\mathbf{0.238}{\scriptscriptstyle \pm \mathbf{0.064}}\)} & \multicolumn{1}{l}{\(54.00{\scriptscriptstyle \pm 0.00}\)} & \multicolumn{1}{l}{\(0.723{\scriptscriptstyle \pm 0.003}\)} \\
\multicolumn{1}{l}{} & \multicolumn{1}{l}{} & Reverse & \multicolumn{1}{l}{\(58.00{\scriptscriptstyle \pm 8.49}\)} & \multicolumn{1}{l}{\(0.852{\scriptscriptstyle \pm 0.285}\)} & \multicolumn{1}{l}{\(40.00{\scriptscriptstyle \pm 2.83}\)} & \multicolumn{1}{l}{\(1.429{\scriptscriptstyle \pm 0.035}\)} & \multicolumn{1}{l}{\(55.00{\scriptscriptstyle \pm 1.41}\)} & \multicolumn{1}{l}{\(0.857{\scriptscriptstyle \pm 0.170}\)} & \multicolumn{1}{l}{\(\mathbf{59.00}{\scriptscriptstyle \pm \mathbf{1.41}}\)} & \multicolumn{1}{l}{\(\mathbf{0.604}{\scriptscriptstyle \pm \mathbf{0.044}}\)} \\ \cmidrule(lr){2-11}
\multicolumn{1}{l}{} & \multicolumn{1}{l}{\multirow{4}{*}{AlphaEdit}} & Edited & \multicolumn{1}{l}{\(100.00{\scriptscriptstyle \pm 0.00}\)} & \multicolumn{1}{l}{\(0.000{\scriptscriptstyle \pm 0.000}\)} & \multicolumn{1}{l}{\(36.00{\scriptscriptstyle \pm 11.31}\)} & \multicolumn{1}{l}{\(2.130{\scriptscriptstyle \pm 0.027}\)} & \multicolumn{1}{l}{\(100.00{\scriptscriptstyle \pm 0.00}\)} & \multicolumn{1}{l}{\(0.000{\scriptscriptstyle \pm 0.000}\)} & \multicolumn{1}{l}{\(25.00{\scriptscriptstyle \pm 7.07}\)} & \multicolumn{1}{l}{\(2.229{\scriptscriptstyle \pm 0.237}\)} \\
\multicolumn{1}{l}{} & \multicolumn{1}{l}{} & Reference & \multicolumn{1}{l}{\(19.00{\scriptscriptstyle \pm 7.07}\)} & \multicolumn{1}{l}{\(3.713{\scriptscriptstyle \pm 0.282}\)} & \multicolumn{1}{l}{\(40.00{\scriptscriptstyle \pm 2.83}\)} & \multicolumn{1}{l}{\(1.488{\scriptscriptstyle \pm 0.020}\)} & \multicolumn{1}{l}{\(26.00{\scriptscriptstyle \pm 5.66}\)} & \multicolumn{1}{l}{\(2.753{\scriptscriptstyle \pm 0.223}\)} & \multicolumn{1}{l}{\(60.00{\scriptscriptstyle \pm 2.83}\)} & \multicolumn{1}{l}{\(0.633{\scriptscriptstyle \pm 0.042}\)} \\
\multicolumn{1}{l}{} & \multicolumn{1}{l}{} & Re-edit & \multicolumn{1}{l}{\(\mathbf{91.00}{\scriptscriptstyle \pm \mathbf{4.24}}\)} & \multicolumn{1}{l}{\(\mathbf{0.053}{\scriptscriptstyle \pm \mathbf{0.001}}\)} & \multicolumn{1}{l}{\(\mathbf{42.00}{\scriptscriptstyle \pm \mathbf{2.83}}\)} & \multicolumn{1}{l}{\(\mathbf{1.377}{\scriptscriptstyle \pm \mathbf{0.040}}\)} & \multicolumn{1}{l}{\(\mathbf{68.00}{\scriptscriptstyle \pm \mathbf{5.66}}\)} & \multicolumn{1}{l}{\(\mathbf{0.374}{\scriptscriptstyle \pm \mathbf{0.132}}\)} & \multicolumn{1}{l}{\(\mathbf{60.00}{\scriptscriptstyle \pm \mathbf{2.83}}\)} & \multicolumn{1}{l}{\(0.663{\scriptscriptstyle \pm 0.029}\)} \\
\multicolumn{1}{l}{} & \multicolumn{1}{l}{} & Reverse & \multicolumn{1}{l}{\(61.00{\scriptscriptstyle \pm 7.07}\)} & \multicolumn{1}{l}{\(0.989{\scriptscriptstyle \pm 0.231}\)} & \multicolumn{1}{l}{\(40.00{\scriptscriptstyle \pm 2.83}\)} & \multicolumn{1}{l}{\(1.437{\scriptscriptstyle \pm 0.026}\)} & \multicolumn{1}{l}{\(54.00{\scriptscriptstyle \pm 8.49}\)} & \multicolumn{1}{l}{\(1.004{\scriptscriptstyle \pm 0.154}\)} & \multicolumn{1}{l}{\(59.00{\scriptscriptstyle \pm 1.41}\)} & \multicolumn{1}{l}{\(\mathbf{0.646}{\scriptscriptstyle \pm \mathbf{0.014}}\)} \\ \midrule
\multicolumn{1}{c}{\multirow{16}{*}{\rotatebox[origin=c]{90}{ZsRE}}} & \multicolumn{1}{l}{\multirow{4}{*}{ROME}} & Edited & \multicolumn{1}{l}{\(100.00{\scriptscriptstyle \pm 0.00}\)} & \multicolumn{1}{l}{\(0.000{\scriptscriptstyle \pm 0.000}\)} & \multicolumn{1}{l}{\(0.00{\scriptscriptstyle \pm 0.00}\)} & \multicolumn{1}{l}{\(8.830{\scriptscriptstyle \pm 0.011}\)} & \multicolumn{1}{l}{\(100.00{\scriptscriptstyle \pm 0.00}\)} & \multicolumn{1}{l}{\(0.000{\scriptscriptstyle \pm 0.000}\)} & \multicolumn{1}{l}{\(5.00{\scriptscriptstyle \pm 4.24}\)} & \multicolumn{1}{l}{\(5.608{\scriptscriptstyle \pm 0.222}\)} \\
\multicolumn{1}{l}{} & \multicolumn{1}{l}{} & Reference & \multicolumn{1}{l}{\(0.00{\scriptscriptstyle \pm 0.00}\)} & \multicolumn{1}{l}{\(9.001{\scriptscriptstyle \pm 0.296}\)} & \multicolumn{1}{l}{\(100.00{\scriptscriptstyle \pm 0.00}\)} & \multicolumn{1}{l}{\(0.070{\scriptscriptstyle \pm 0.015}\)} & \multicolumn{1}{l}{\(5.00{\scriptscriptstyle \pm 4.24}\)} & \multicolumn{1}{l}{\(8.269{\scriptscriptstyle \pm 0.028}\)} & \multicolumn{1}{l}{\(100.00{\scriptscriptstyle \pm 0.00}\)} & \multicolumn{1}{l}{\(0.038{\scriptscriptstyle \pm 0.040}\)} \\
\multicolumn{1}{l}{} & \multicolumn{1}{l}{} & Re-edit & \multicolumn{1}{l}{\(89.00{\scriptscriptstyle \pm 4.24}\)} & \multicolumn{1}{l}{\(0.415{\scriptscriptstyle \pm 0.403}\)} & \multicolumn{1}{l}{\(\mathbf{100.00}{\scriptscriptstyle \pm \mathbf{0.00}}\)} & \multicolumn{1}{l}{\(\mathbf{0.227}{\scriptscriptstyle \pm \mathbf{0.011}}\)} & \multicolumn{1}{l}{\(48.00{\scriptscriptstyle \pm 16.97}\)} & \multicolumn{1}{l}{\(1.785{\scriptscriptstyle \pm 0.594}\)} & \multicolumn{1}{l}{\(\mathbf{100.00}{\scriptscriptstyle \pm \mathbf{0.00}}\)} & \multicolumn{1}{l}{\(\mathbf{0.094}{\scriptscriptstyle \pm \mathbf{0.053}}\)} \\
\multicolumn{1}{l}{} & \multicolumn{1}{l}{} & Reverse & \multicolumn{1}{l}{\(\mathbf{98.00}{\scriptscriptstyle \pm \mathbf{2.83}}\)} & \multicolumn{1}{l}{\(\mathbf{0.055}{\scriptscriptstyle \pm \mathbf{0.030}}\)} & \multicolumn{1}{l}{\(98.00{\scriptscriptstyle \pm 2.83}\)} & \multicolumn{1}{l}{\(0.398{\scriptscriptstyle \pm 0.279}\)} & \multicolumn{1}{l}{\(\mathbf{76.00}{\scriptscriptstyle \pm \mathbf{8.49}}\)} & \multicolumn{1}{l}{\(\mathbf{0.656}{\scriptscriptstyle \pm \mathbf{0.179}}\)} & \multicolumn{1}{l}{\(99.00{\scriptscriptstyle \pm 1.41}\)} & \multicolumn{1}{l}{\(0.164{\scriptscriptstyle \pm 0.047}\)} \\ \cmidrule(lr){2-11}
\multicolumn{1}{l}{} & \multicolumn{1}{l}{\multirow{4}{*}{SimIE}} & Edited & \multicolumn{1}{l}{\(100.00{\scriptscriptstyle \pm 0.00}\)} & \multicolumn{1}{l}{\(0.000{\scriptscriptstyle \pm 0.000}\)} & \multicolumn{1}{l}{\(4.00{\scriptscriptstyle \pm 2.83}\)} & \multicolumn{1}{l}{\(7.360{\scriptscriptstyle \pm 0.018}\)} & \multicolumn{1}{l}{\(100.00{\scriptscriptstyle \pm 0.00}\)} & \multicolumn{1}{l}{\(0.000{\scriptscriptstyle \pm 0.000}\)} & \multicolumn{1}{l}{\(23.00{\scriptscriptstyle \pm 4.24}\)} & \multicolumn{1}{l}{\(3.324{\scriptscriptstyle \pm 0.390}\)} \\
\multicolumn{1}{l}{} & \multicolumn{1}{l}{} & Reference & \multicolumn{1}{l}{\(4.00{\scriptscriptstyle \pm 2.83}\)} & \multicolumn{1}{l}{\(8.882{\scriptscriptstyle \pm 0.571}\)} & \multicolumn{1}{l}{\(100.00{\scriptscriptstyle \pm 0.00}\)} & \multicolumn{1}{l}{\(0.068{\scriptscriptstyle \pm 0.020}\)} & \multicolumn{1}{l}{\(23.00{\scriptscriptstyle \pm 4.24}\)} & \multicolumn{1}{l}{\(5.954{\scriptscriptstyle \pm 0.521}\)} & \multicolumn{1}{l}{\(100.00{\scriptscriptstyle \pm 0.00}\)} & \multicolumn{1}{l}{\(0.031{\scriptscriptstyle \pm 0.028}\)} \\
\multicolumn{1}{l}{} & \multicolumn{1}{l}{} & Re-edit & \multicolumn{1}{l}{\(76.00{\scriptscriptstyle \pm 16.97}\)} & \multicolumn{1}{l}{\(0.626{\scriptscriptstyle \pm 0.529}\)} & \multicolumn{1}{l}{\(\mathbf{100.00}{\scriptscriptstyle \pm \mathbf{0.00}}\)} & \multicolumn{1}{l}{\(\mathbf{0.157}{\scriptscriptstyle \pm \mathbf{0.027}}\)} & \multicolumn{1}{l}{\(54.00{\scriptscriptstyle \pm 5.66}\)} & \multicolumn{1}{l}{\(1.008{\scriptscriptstyle \pm 0.366}\)} & \multicolumn{1}{l}{\(\mathbf{100.00}{\scriptscriptstyle \pm \mathbf{0.00}}\)} & \multicolumn{1}{l}{\(\mathbf{0.064}{\scriptscriptstyle \pm \mathbf{0.038}}\)} \\
\multicolumn{1}{l}{} & \multicolumn{1}{l}{} & Reverse & \multicolumn{1}{l}{\(\mathbf{93.00}{\scriptscriptstyle \pm \mathbf{4.24}}\)} & \multicolumn{1}{l}{\(\mathbf{0.174}{\scriptscriptstyle \pm \mathbf{0.089}}\)} & \multicolumn{1}{l}{\(98.00{\scriptscriptstyle \pm 2.83}\)} & \multicolumn{1}{l}{\(0.307{\scriptscriptstyle \pm 0.261}\)} & \multicolumn{1}{l}{\(\mathbf{73.00}{\scriptscriptstyle \pm \mathbf{1.41}}\)} & \multicolumn{1}{l}{\(\mathbf{0.491}{\scriptscriptstyle \pm \mathbf{0.010}}\)} & \multicolumn{1}{l}{\(99.00{\scriptscriptstyle \pm 1.41}\)} & \multicolumn{1}{l}{\(0.121{\scriptscriptstyle \pm 0.011}\)} \\ \cmidrule(lr){2-11}
\multicolumn{1}{l}{} & \multicolumn{1}{l}{\multirow{4}{*}{MEMIT}} & Edited & \multicolumn{1}{l}{\(100.00{\scriptscriptstyle \pm 0.00}\)} & \multicolumn{1}{l}{\(0.000{\scriptscriptstyle \pm 0.000}\)} & \multicolumn{1}{l}{\(18.00{\scriptscriptstyle \pm 2.83}\)} & \multicolumn{1}{l}{\(3.769{\scriptscriptstyle \pm 0.553}\)} & \multicolumn{1}{l}{\(100.00{\scriptscriptstyle \pm 0.00}\)} & \multicolumn{1}{l}{\(0.000{\scriptscriptstyle \pm 0.000}\)} & \multicolumn{1}{l}{\(5.00{\scriptscriptstyle \pm 1.41}\)} & \multicolumn{1}{l}{\(5.679{\scriptscriptstyle \pm 0.347}\)} \\
\multicolumn{1}{l}{} & \multicolumn{1}{l}{} & Reference & \multicolumn{1}{l}{\(18.00{\scriptscriptstyle \pm 2.83}\)} & \multicolumn{1}{l}{\(6.843{\scriptscriptstyle \pm 0.682}\)} & \multicolumn{1}{l}{\(100.00{\scriptscriptstyle \pm 0.00}\)} & \multicolumn{1}{l}{\(0.516{\scriptscriptstyle \pm 0.012}\)} & \multicolumn{1}{l}{\(5.00{\scriptscriptstyle \pm 1.41}\)} & \multicolumn{1}{l}{\(8.091{\scriptscriptstyle \pm 0.108}\)} & \multicolumn{1}{l}{\(99.00{\scriptscriptstyle \pm 1.41}\)} & \multicolumn{1}{l}{\(0.140{\scriptscriptstyle \pm 0.027}\)} \\
\multicolumn{1}{l}{} & \multicolumn{1}{l}{} & Re-edit & \multicolumn{1}{l}{\(\mathbf{79.00}{\scriptscriptstyle \pm \mathbf{12.73}}\)} & \multicolumn{1}{l}{\(\mathbf{0.486}{\scriptscriptstyle \pm \mathbf{0.362}}\)} & \multicolumn{1}{l}{\(\mathbf{100.00}{\scriptscriptstyle \pm \mathbf{0.00}}\)} & \multicolumn{1}{l}{\(\mathbf{0.370}{\scriptscriptstyle \pm \mathbf{0.037}}\)} & \multicolumn{1}{l}{\(53.00{\scriptscriptstyle \pm 4.24}\)} & \multicolumn{1}{l}{\(1.445{\scriptscriptstyle \pm 0.667}\)} & \multicolumn{1}{l}{\(\mathbf{99.00}{\scriptscriptstyle \pm \mathbf{1.41}}\)} & \multicolumn{1}{l}{\(\mathbf{0.162}{\scriptscriptstyle \pm \mathbf{0.004}}\)} \\
\multicolumn{1}{l}{} & \multicolumn{1}{l}{} & Reverse & \multicolumn{1}{l}{\(69.00{\scriptscriptstyle \pm 4.24}\)} & \multicolumn{1}{l}{\(1.056{\scriptscriptstyle \pm 0.115}\)} & \multicolumn{1}{l}{\(98.00{\scriptscriptstyle \pm 2.83}\)} & \multicolumn{1}{l}{\(0.568{\scriptscriptstyle \pm 0.131}\)} & \multicolumn{1}{l}{\(\mathbf{63.00}{\scriptscriptstyle \pm \mathbf{4.24}}\)} & \multicolumn{1}{l}{\(\mathbf{1.123}{\scriptscriptstyle \pm \mathbf{0.435}}\)} & \multicolumn{1}{l}{\(\mathbf{99.00}{\scriptscriptstyle \pm \mathbf{1.41}}\)} & \multicolumn{1}{l}{\(0.211{\scriptscriptstyle \pm 0.015}\)} \\ \cmidrule(lr){2-11}
\multicolumn{1}{l}{} & \multicolumn{1}{l}{\multirow{4}{*}{AlphaEdit}} & Edited & \multicolumn{1}{l}{\(100.00{\scriptscriptstyle \pm 0.00}\)} & \multicolumn{1}{l}{\(0.000{\scriptscriptstyle \pm 0.000}\)} & \multicolumn{1}{l}{\(12.00{\scriptscriptstyle \pm 2.83}\)} & \multicolumn{1}{l}{\(4.636{\scriptscriptstyle \pm 0.139}\)} & \multicolumn{1}{l}{\(100.00{\scriptscriptstyle \pm 0.00}\)} & \multicolumn{1}{l}{\(0.000{\scriptscriptstyle \pm 0.000}\)} & \multicolumn{1}{l}{\(6.00{\scriptscriptstyle \pm 2.83}\)} & \multicolumn{1}{l}{\(5.679{\scriptscriptstyle \pm 0.751}\)} \\
\multicolumn{1}{l}{} & \multicolumn{1}{l}{} & Reference & \multicolumn{1}{l}{\(12.00{\scriptscriptstyle \pm 2.83}\)} & \multicolumn{1}{l}{\(7.601{\scriptscriptstyle \pm 0.561}\)} & \multicolumn{1}{l}{\(100.00{\scriptscriptstyle \pm 0.00}\)} & \multicolumn{1}{l}{\(0.523{\scriptscriptstyle \pm 0.011}\)} & \multicolumn{1}{l}{\(6.00{\scriptscriptstyle \pm 2.83}\)} & \multicolumn{1}{l}{\(7.484{\scriptscriptstyle \pm 0.667}\)} & \multicolumn{1}{l}{\(99.00{\scriptscriptstyle \pm 1.41}\)} & \multicolumn{1}{l}{\(0.198{\scriptscriptstyle \pm 0.010}\)} \\
\multicolumn{1}{l}{} & \multicolumn{1}{l}{} & Re-edit & \multicolumn{1}{l}{\(\mathbf{80.00}{\scriptscriptstyle \pm \mathbf{14.14}}\)} & \multicolumn{1}{l}{\(\mathbf{0.609}{\scriptscriptstyle \pm \mathbf{0.400}}\)} & \multicolumn{1}{l}{\(\mathbf{100.00}{\scriptscriptstyle \pm \mathbf{0.00}}\)} & \multicolumn{1}{l}{\(\mathbf{0.375}{\scriptscriptstyle \pm \mathbf{0.051}}\)} & \multicolumn{1}{l}{\(30.00{\scriptscriptstyle \pm 5.66}\)} & \multicolumn{1}{l}{\(2.530{\scriptscriptstyle \pm 0.441}\)} & \multicolumn{1}{l}{\(\mathbf{99.00}{\scriptscriptstyle \pm \mathbf{1.41}}\)} & \multicolumn{1}{l}{\(\mathbf{0.162}{\scriptscriptstyle \pm \mathbf{0.022}}\)} \\
\multicolumn{1}{l}{} & \multicolumn{1}{l}{} & Reverse & \multicolumn{1}{l}{\(56.00{\scriptscriptstyle \pm 2.83}\)} & \multicolumn{1}{l}{\(1.760{\scriptscriptstyle \pm 0.005}\)} & \multicolumn{1}{l}{\(98.00{\scriptscriptstyle \pm 2.83}\)} & \multicolumn{1}{l}{\(0.584{\scriptscriptstyle \pm 0.134}\)} & \multicolumn{1}{l}{\(\mathbf{51.00}{\scriptscriptstyle \pm \mathbf{12.73}}\)} & \multicolumn{1}{l}{\(\mathbf{2.080}{\scriptscriptstyle \pm \mathbf{0.551}}\)} & \multicolumn{1}{l}{\(98.00{\scriptscriptstyle \pm 2.83}\)} & \multicolumn{1}{l}{\(0.267{\scriptscriptstyle \pm 0.092}\)} \\ \bottomrule
\end{tabular}%
}
  }
  \caption{Performance comparison with the global reversal and re-editing baselines when reversing 50 facts simultaneously. Agreement and KL divergence are reported on both the remained set and the reversed set across different base models and editing methods. Higher agreement and lower KL divergence indicate better performance.}
  \label{50_gptj_llama2}
\end{table*}

\begin{table*}[t!]
  \centering
  \footnotesize
  \renewcommand\arraystretch{1.2}
  \setlength{\tabcolsep}{1.5mm}{
  \resizebox{\textwidth}{!}{%
\begin{tabular}{lllllllllll}
\toprule
\multicolumn{3}{l}{\multirow{3}{*}{}} & \multicolumn{4}{c}{\textbf{GPT-J 6B}} & \multicolumn{4}{c}{\textbf{LLaMA2-7B}} \\ \cmidrule(lr){4-7}\cmidrule(lr){8-11}
\multicolumn{3}{l}{} & \multicolumn{2}{c}{Remained Set} & \multicolumn{2}{c}{Reversed Set} & \multicolumn{2}{c}{Remained Set} & \multicolumn{2}{c}{Reversed Set} \\ \cmidrule(lr){4-5}\cmidrule(lr){6-7}\cmidrule(lr){8-9}\cmidrule(lr){10-11}
\multicolumn{3}{l}{} & \multicolumn{1}{c}{Agree.} & \multicolumn{1}{c}{KL Div.} & \multicolumn{1}{c}{Agree.} & \multicolumn{1}{c}{KL Div.} & \multicolumn{1}{c}{Agree.} & \multicolumn{1}{c}{KL Div.} & \multicolumn{1}{c}{Agree.} & \multicolumn{1}{c}{KL Div.} \\ \midrule
\multicolumn{1}{c}{\multirow{16}{*}{\rotatebox[origin=c]{90}{Counterfact}}} & \multicolumn{1}{l}{\multirow{4}{*}{ROME}} & Edited & \multicolumn{1}{l}{\(100.00{\scriptscriptstyle \pm 0.00}\)} & \multicolumn{1}{l}{\(0.000{\scriptscriptstyle \pm 0.000}\)} & \multicolumn{1}{l}{\(31.00{\scriptscriptstyle \pm 46.48}\)} & \multicolumn{1}{l}{\(3.585{\scriptscriptstyle \pm 3.138}\)} & \multicolumn{1}{l}{\(100.00{\scriptscriptstyle \pm 0.00}\)} & \multicolumn{1}{l}{\(0.000{\scriptscriptstyle \pm 0.000}\)} & \multicolumn{1}{l}{\(21.00{\scriptscriptstyle \pm 40.94}\)} & \multicolumn{1}{l}{\(2.810{\scriptscriptstyle \pm 2.324}\)} \\
\multicolumn{1}{l}{} & \multicolumn{1}{l}{} & Reference & \multicolumn{1}{l}{\(31.00{\scriptscriptstyle \pm 0.47}\)} & \multicolumn{1}{l}{\(4.480{\scriptscriptstyle \pm 0.038}\)} & \multicolumn{1}{l}{\(81.00{\scriptscriptstyle \pm 39.43}\)} & \multicolumn{1}{l}{\(0.163{\scriptscriptstyle \pm 0.202}\)} & \multicolumn{1}{l}{\(24.00{\scriptscriptstyle \pm 0.43}\)} & \multicolumn{1}{l}{\(3.642{\scriptscriptstyle \pm 0.031}\)} & \multicolumn{1}{l}{\(89.00{\scriptscriptstyle \pm 31.45}\)} & \multicolumn{1}{l}{\(0.114{\scriptscriptstyle \pm 0.197}\)} \\
\multicolumn{1}{l}{} & \multicolumn{1}{l}{} & Re-edit & \multicolumn{1}{l}{\(\mathbf{99.28}{\scriptscriptstyle \pm \mathbf{0.92}}\)} & \multicolumn{1}{l}{\(\mathbf{0.000}{\scriptscriptstyle \pm \mathbf{0.001}}\)} & \multicolumn{1}{l}{\(77.00{\scriptscriptstyle \pm 42.30}\)} & \multicolumn{1}{l}{\(0.216{\scriptscriptstyle \pm 0.204}\)} & \multicolumn{1}{l}{\(\mathbf{98.07}{\scriptscriptstyle \pm \mathbf{1.16}}\)} & \multicolumn{1}{l}{\(\mathbf{0.001}{\scriptscriptstyle \pm \mathbf{0.001}}\)} & \multicolumn{1}{l}{\(\mathbf{89.00}{\scriptscriptstyle \pm \mathbf{31.45}}\)} & \multicolumn{1}{l}{\(0.176{\scriptscriptstyle \pm 0.222}\)} \\
\multicolumn{1}{l}{} & \multicolumn{1}{l}{} & Reverse & \multicolumn{1}{l}{\(93.17{\scriptscriptstyle \pm 3.33}\)} & \multicolumn{1}{l}{\(0.032{\scriptscriptstyle \pm 0.026}\)} & \multicolumn{1}{l}{\(\mathbf{81.00}{\scriptscriptstyle \pm \mathbf{39.43}}\)} & \multicolumn{1}{l}{\(\mathbf{0.154}{\scriptscriptstyle \pm \mathbf{0.187}}\)} & \multicolumn{1}{l}{\(92.97{\scriptscriptstyle \pm 2.62}\)} & \multicolumn{1}{l}{\(0.030{\scriptscriptstyle \pm 0.023}\)} & \multicolumn{1}{l}{\(86.00{\scriptscriptstyle \pm 34.87}\)} & \multicolumn{1}{l}{\(\mathbf{0.157}{\scriptscriptstyle \pm \mathbf{0.381}}\)} \\ \cmidrule(lr){2-11}
\multicolumn{1}{l}{} & \multicolumn{1}{l}{\multirow{4}{*}{SimIE}} & Edited & \multicolumn{1}{l}{\(100.00{\scriptscriptstyle \pm 0.00}\)} & \multicolumn{1}{l}{\(0.000{\scriptscriptstyle \pm 0.000}\)} & \multicolumn{1}{l}{\(24.00{\scriptscriptstyle \pm 42.92}\)} & \multicolumn{1}{l}{\(3.546{\scriptscriptstyle \pm 2.940}\)} & \multicolumn{1}{l}{\(100.00{\scriptscriptstyle \pm 0.00}\)} & \multicolumn{1}{l}{\(0.000{\scriptscriptstyle \pm 0.000}\)} & \multicolumn{1}{l}{\(26.00{\scriptscriptstyle \pm 44.08}\)} & \multicolumn{1}{l}{\(2.135{\scriptscriptstyle \pm 1.949}\)} \\
\multicolumn{1}{l}{} & \multicolumn{1}{l}{} & Reference & \multicolumn{1}{l}{\(25.00{\scriptscriptstyle \pm 0.44}\)} & \multicolumn{1}{l}{\(4.389{\scriptscriptstyle \pm 0.035}\)} & \multicolumn{1}{l}{\(81.00{\scriptscriptstyle \pm 39.43}\)} & \multicolumn{1}{l}{\(0.169{\scriptscriptstyle \pm 0.205}\)} & \multicolumn{1}{l}{\(28.00{\scriptscriptstyle \pm 0.46}\)} & \multicolumn{1}{l}{\(2.808{\scriptscriptstyle \pm 0.025}\)} & \multicolumn{1}{l}{\(85.00{\scriptscriptstyle \pm 35.89}\)} & \multicolumn{1}{l}{\(0.136{\scriptscriptstyle \pm 0.189}\)} \\
\multicolumn{1}{l}{} & \multicolumn{1}{l}{} & Re-edit & \multicolumn{1}{l}{\(\mathbf{98.70}{\scriptscriptstyle \pm \mathbf{0.98}}\)} & \multicolumn{1}{l}{\(\mathbf{0.001}{\scriptscriptstyle \pm \mathbf{0.001}}\)} & \multicolumn{1}{l}{\(\mathbf{80.00}{\scriptscriptstyle \pm \mathbf{40.20}}\)} & \multicolumn{1}{l}{\(\mathbf{0.237}{\scriptscriptstyle \pm \mathbf{0.263}}\)} & \multicolumn{1}{l}{\(\mathbf{98.89}{\scriptscriptstyle \pm \mathbf{0.90}}\)} & \multicolumn{1}{l}{\(\mathbf{0.001}{\scriptscriptstyle \pm \mathbf{0.001}}\)} & \multicolumn{1}{l}{\(\mathbf{87.00}{\scriptscriptstyle \pm \mathbf{33.80}}\)} & \multicolumn{1}{l}{\(\mathbf{0.168}{\scriptscriptstyle \pm \mathbf{0.213}}\)} \\
\multicolumn{1}{l}{} & \multicolumn{1}{l}{} & Reverse & \multicolumn{1}{l}{\(74.95{\scriptscriptstyle \pm 7.84}\)} & \multicolumn{1}{l}{\(0.455{\scriptscriptstyle \pm 0.300}\)} & \multicolumn{1}{l}{\(61.00{\scriptscriptstyle \pm 49.02}\)} & \multicolumn{1}{l}{\(0.501{\scriptscriptstyle \pm 0.469}\)} & \multicolumn{1}{l}{\(92.93{\scriptscriptstyle \pm 2.37}\)} & \multicolumn{1}{l}{\(0.039{\scriptscriptstyle \pm 0.024}\)} & \multicolumn{1}{l}{\(82.00{\scriptscriptstyle \pm 38.61}\)} & \multicolumn{1}{l}{\(0.227{\scriptscriptstyle \pm 0.324}\)} \\ \cmidrule(lr){2-11}
\multicolumn{1}{l}{} & \multicolumn{1}{l}{\multirow{4}{*}{MEMIT}} & Edited & \multicolumn{1}{l}{\(100.00{\scriptscriptstyle \pm 0.00}\)} & \multicolumn{1}{l}{\(0.000{\scriptscriptstyle \pm 0.000}\)} & \multicolumn{1}{l}{\(35.00{\scriptscriptstyle \pm 47.94}\)} & \multicolumn{1}{l}{\(1.936{\scriptscriptstyle \pm 2.045}\)} & \multicolumn{1}{l}{\(100.00{\scriptscriptstyle \pm 0.00}\)} & \multicolumn{1}{l}{\(0.000{\scriptscriptstyle \pm 0.000}\)} & \multicolumn{1}{l}{\(13.00{\scriptscriptstyle \pm 33.80}\)} & \multicolumn{1}{l}{\(3.130{\scriptscriptstyle \pm 2.413}\)} \\
\multicolumn{1}{l}{} & \multicolumn{1}{l}{} & Reference & \multicolumn{1}{l}{\(18.00{\scriptscriptstyle \pm 0.39}\)} & \multicolumn{1}{l}{\(3.624{\scriptscriptstyle \pm 0.028}\)} & \multicolumn{1}{l}{\(40.00{\scriptscriptstyle \pm 49.24}\)} & \multicolumn{1}{l}{\(1.476{\scriptscriptstyle \pm 0.964}\)} & \multicolumn{1}{l}{\(10.00{\scriptscriptstyle \pm 0.30}\)} & \multicolumn{1}{l}{\(3.955{\scriptscriptstyle \pm 0.027}\)} & \multicolumn{1}{l}{\(60.00{\scriptscriptstyle \pm 49.24}\)} & \multicolumn{1}{l}{\(0.575{\scriptscriptstyle \pm 0.737}\)} \\
\multicolumn{1}{l}{} & \multicolumn{1}{l}{} & Re-edit & \multicolumn{1}{l}{\(\mathbf{99.45}{\scriptscriptstyle \pm \mathbf{0.63}}\)} & \multicolumn{1}{l}{\(\mathbf{0.000}{\scriptscriptstyle \pm \mathbf{0.000}}\)} & \multicolumn{1}{l}{\(\mathbf{40.00}{\scriptscriptstyle \pm \mathbf{49.24}}\)} & \multicolumn{1}{l}{\(\mathbf{1.376}{\scriptscriptstyle \pm \mathbf{0.990}}\)} & \multicolumn{1}{l}{\(\mathbf{98.27}{\scriptscriptstyle \pm \mathbf{1.27}}\)} & \multicolumn{1}{l}{\(\mathbf{0.004}{\scriptscriptstyle \pm \mathbf{0.002}}\)} & \multicolumn{1}{l}{\(52.00{\scriptscriptstyle \pm 50.21}\)} & \multicolumn{1}{l}{\(0.728{\scriptscriptstyle \pm 0.837}\)} \\
\multicolumn{1}{l}{} & \multicolumn{1}{l}{} & Reverse & \multicolumn{1}{l}{\(60.20{\scriptscriptstyle \pm 4.02}\)} & \multicolumn{1}{l}{\(0.878{\scriptscriptstyle \pm 0.300}\)} & \multicolumn{1}{l}{\(\mathbf{40.00}{\scriptscriptstyle \pm \mathbf{49.24}}\)} & \multicolumn{1}{l}{\(1.465{\scriptscriptstyle \pm 0.950}\)} & \multicolumn{1}{l}{\(82.02{\scriptscriptstyle \pm 11.14}\)} & \multicolumn{1}{l}{\(0.212{\scriptscriptstyle \pm 0.223}\)} & \multicolumn{1}{l}{\(\mathbf{61.00}{\scriptscriptstyle \pm \mathbf{49.02}}\)} & \multicolumn{1}{l}{\(\mathbf{0.566}{\scriptscriptstyle \pm \mathbf{0.738}}\)} \\ \cmidrule(lr){2-11}
\multicolumn{1}{l}{} & \multicolumn{1}{l}{\multirow{4}{*}{AlphaEdit}} & Edited & \multicolumn{1}{l}{\(100.00{\scriptscriptstyle \pm 0.00}\)} & \multicolumn{1}{l}{\(0.000{\scriptscriptstyle \pm 0.000}\)} & \multicolumn{1}{l}{\(36.00{\scriptscriptstyle \pm 48.24}\)} & \multicolumn{1}{l}{\(2.130{\scriptscriptstyle \pm 2.269}\)} & \multicolumn{1}{l}{\(100.00{\scriptscriptstyle \pm 0.00}\)} & \multicolumn{1}{l}{\(0.000{\scriptscriptstyle \pm 0.000}\)} & \multicolumn{1}{l}{\(25.00{\scriptscriptstyle \pm 43.52}\)} & \multicolumn{1}{l}{\(2.229{\scriptscriptstyle \pm 2.105}\)} \\
\multicolumn{1}{l}{} & \multicolumn{1}{l}{} & Reference & \multicolumn{1}{l}{\(19.00{\scriptscriptstyle \pm 0.40}\)} & \multicolumn{1}{l}{\(3.713{\scriptscriptstyle \pm 0.028}\)} & \multicolumn{1}{l}{\(40.00{\scriptscriptstyle \pm 49.24}\)} & \multicolumn{1}{l}{\(1.488{\scriptscriptstyle \pm 0.992}\)} & \multicolumn{1}{l}{\(26.00{\scriptscriptstyle \pm 0.45}\)} & \multicolumn{1}{l}{\(2.753{\scriptscriptstyle \pm 0.025}\)} & \multicolumn{1}{l}{\(60.00{\scriptscriptstyle \pm 49.24}\)} & \multicolumn{1}{l}{\(0.633{\scriptscriptstyle \pm 0.713}\)} \\
\multicolumn{1}{l}{} & \multicolumn{1}{l}{} & Re-edit & \multicolumn{1}{l}{\(\mathbf{99.46}{\scriptscriptstyle \pm \mathbf{0.72}}\)} & \multicolumn{1}{l}{\(\mathbf{0.000}{\scriptscriptstyle \pm \mathbf{0.000}}\)} & \multicolumn{1}{l}{\(\mathbf{42.00}{\scriptscriptstyle \pm \mathbf{49.60}}\)} & \multicolumn{1}{l}{\(\mathbf{1.381}{\scriptscriptstyle \pm \mathbf{1.018}}\)} & \multicolumn{1}{l}{\(\mathbf{95.82}{\scriptscriptstyle \pm \mathbf{2.37}}\)} & \multicolumn{1}{l}{\(\mathbf{0.008}{\scriptscriptstyle \pm \mathbf{0.006}}\)} & \multicolumn{1}{l}{\(56.00{\scriptscriptstyle \pm 49.89}\)} & \multicolumn{1}{l}{\(0.661{\scriptscriptstyle \pm 0.720}\)} \\
\multicolumn{1}{l}{} & \multicolumn{1}{l}{} & Reverse & \multicolumn{1}{l}{\(55.96{\scriptscriptstyle \pm 6.04}\)} & \multicolumn{1}{l}{\(0.876{\scriptscriptstyle \pm 0.297}\)} & \multicolumn{1}{l}{\(40.00{\scriptscriptstyle \pm 49.24}\)} & \multicolumn{1}{l}{\(1.467{\scriptscriptstyle \pm 0.965}\)} & \multicolumn{1}{l}{\(65.66{\scriptscriptstyle \pm 11.13}\)} & \multicolumn{1}{l}{\(0.459{\scriptscriptstyle \pm 0.283}\)} & \multicolumn{1}{l}{\(\mathbf{62.00}{\scriptscriptstyle \pm \mathbf{48.78}}\)} & \multicolumn{1}{l}{\(\mathbf{0.614}{\scriptscriptstyle \pm \mathbf{0.700}}\)} \\ \midrule

\multicolumn{1}{c}{\multirow{16}{*}{\rotatebox[origin=c]{90}{ZsRE}}} & \multicolumn{1}{l}{\multirow{4}{*}{ROME}} & Edited & \multicolumn{1}{l}{\(100.00{\scriptscriptstyle \pm 0.00}\)} & \multicolumn{1}{l}{\(0.000{\scriptscriptstyle \pm 0.000}\)} & \multicolumn{1}{l}{\(0.00{\scriptscriptstyle \pm 0.00}\)} & \multicolumn{1}{l}{\(8.830{\scriptscriptstyle \pm 3.213}\)} & \multicolumn{1}{l}{\(100.00{\scriptscriptstyle \pm 0.00}\)} & \multicolumn{1}{l}{\(0.000{\scriptscriptstyle \pm 0.000}\)} & \multicolumn{1}{l}{\(5.00{\scriptscriptstyle \pm 21.90}\)} & \multicolumn{1}{l}{\(5.608{\scriptscriptstyle \pm 3.603}\)} \\
\multicolumn{1}{l}{} & \multicolumn{1}{l}{} & Reference & \multicolumn{1}{l}{\(0.00{\scriptscriptstyle \pm 0.00}\)} & \multicolumn{1}{l}{\(9.001{\scriptscriptstyle \pm 0.029}\)} & \multicolumn{1}{l}{\(100.00{\scriptscriptstyle \pm 0.00}\)} & \multicolumn{1}{l}{\(0.070{\scriptscriptstyle \pm 0.081}\)} & \multicolumn{1}{l}{\(5.00{\scriptscriptstyle \pm 0.22}\)} & \multicolumn{1}{l}{\(8.269{\scriptscriptstyle \pm 0.034}\)} & \multicolumn{1}{l}{\(100.00{\scriptscriptstyle \pm 0.00}\)} & \multicolumn{1}{l}{\(0.038{\scriptscriptstyle \pm 0.232}\)} \\
\multicolumn{1}{l}{} & \multicolumn{1}{l}{} & Re-edit & \multicolumn{1}{l}{\(\mathbf{99.62}{\scriptscriptstyle \pm \mathbf{1.70}}\)} & \multicolumn{1}{l}{\(\mathbf{0.012}{\scriptscriptstyle \pm \mathbf{0.059}}\)} & \multicolumn{1}{l}{\(\mathbf{100.00}{\scriptscriptstyle \pm \mathbf{0.00}}\)} & \multicolumn{1}{l}{\(\mathbf{0.218}{\scriptscriptstyle \pm \mathbf{0.151}}\)} & \multicolumn{1}{l}{\(\mathbf{98.23}{\scriptscriptstyle \pm \mathbf{2.95}}\)} & \multicolumn{1}{l}{\(\mathbf{0.020}{\scriptscriptstyle \pm \mathbf{0.087}}\)} & \multicolumn{1}{l}{\(\mathbf{100.00}{\scriptscriptstyle \pm \mathbf{0.00}}\)} & \multicolumn{1}{l}{\(\mathbf{0.088}{\scriptscriptstyle \pm \mathbf{0.292}}\)} \\
\multicolumn{1}{l}{} & \multicolumn{1}{l}{} & Reverse & \multicolumn{1}{l}{\(92.32{\scriptscriptstyle \pm 6.04}\)} & \multicolumn{1}{l}{\(0.179{\scriptscriptstyle \pm 0.152}\)} & \multicolumn{1}{l}{\(98.00{\scriptscriptstyle \pm 14.07}\)} & \multicolumn{1}{l}{\(0.339{\scriptscriptstyle \pm 1.435}\)} & \multicolumn{1}{l}{\(92.32{\scriptscriptstyle \pm 2.09}\)} & \multicolumn{1}{l}{\(0.051{\scriptscriptstyle \pm 0.020}\)} & \multicolumn{1}{l}{\(99.00{\scriptscriptstyle \pm 10.00}\)} & \multicolumn{1}{l}{\(0.169{\scriptscriptstyle \pm 0.614}\)} \\ \cmidrule(lr){2-11}
\multicolumn{1}{l}{} & \multicolumn{1}{l}{\multirow{4}{*}{SimIE}} & Edited & \multicolumn{1}{l}{\(100.00{\scriptscriptstyle \pm 0.00}\)} & \multicolumn{1}{l}{\(0.000{\scriptscriptstyle \pm 0.000}\)} & \multicolumn{1}{l}{\(4.00{\scriptscriptstyle \pm 19.69}\)} & \multicolumn{1}{l}{\(7.360{\scriptscriptstyle \pm 3.423}\)} & \multicolumn{1}{l}{\(100.00{\scriptscriptstyle \pm 0.00}\)} & \multicolumn{1}{l}{\(0.000{\scriptscriptstyle \pm 0.000}\)} & \multicolumn{1}{l}{\(23.00{\scriptscriptstyle \pm 42.30}\)} & \multicolumn{1}{l}{\(3.324{\scriptscriptstyle \pm 2.430}\)} \\
\multicolumn{1}{l}{} & \multicolumn{1}{l}{} & Reference & \multicolumn{1}{l}{\(4.00{\scriptscriptstyle \pm 0.20}\)} & \multicolumn{1}{l}{\(8.882{\scriptscriptstyle \pm 0.034}\)} & \multicolumn{1}{l}{\(100.00{\scriptscriptstyle \pm 0.00}\)} & \multicolumn{1}{l}{\(0.068{\scriptscriptstyle \pm 0.096}\)} & \multicolumn{1}{l}{\(23.00{\scriptscriptstyle \pm 0.43}\)} & \multicolumn{1}{l}{\(5.954{\scriptscriptstyle \pm 0.039}\)} & \multicolumn{1}{l}{\(100.00{\scriptscriptstyle \pm 0.00}\)} & \multicolumn{1}{l}{\(0.031{\scriptscriptstyle \pm 0.149}\)} \\
\multicolumn{1}{l}{} & \multicolumn{1}{l}{} & Re-edit & \multicolumn{1}{l}{\(\mathbf{99.33}{\scriptscriptstyle \pm \mathbf{1.48}}\)} & \multicolumn{1}{l}{\(\mathbf{0.010}{\scriptscriptstyle \pm \mathbf{0.037}}\)} & \multicolumn{1}{l}{\(\mathbf{100.00}{\scriptscriptstyle \pm \mathbf{0.00}}\)} & \multicolumn{1}{l}{\(\mathbf{0.134}{\scriptscriptstyle \pm \mathbf{0.126}}\)} & \multicolumn{1}{l}{\(\mathbf{97.88}{\scriptscriptstyle \pm \mathbf{2.07}}\)} & \multicolumn{1}{l}{\(\mathbf{0.011}{\scriptscriptstyle \pm \mathbf{0.031}}\)} & \multicolumn{1}{l}{\(\mathbf{100.00}{\scriptscriptstyle \pm \mathbf{0.00}}\)} & \multicolumn{1}{l}{\(\mathbf{0.061}{\scriptscriptstyle \pm \mathbf{0.258}}\)} \\
\multicolumn{1}{l}{} & \multicolumn{1}{l}{} & Reverse & \multicolumn{1}{l}{\(96.57{\scriptscriptstyle \pm 1.36}\)} & \multicolumn{1}{l}{\(0.059{\scriptscriptstyle \pm 0.042}\)} & \multicolumn{1}{l}{\(98.00{\scriptscriptstyle \pm 14.07}\)} & \multicolumn{1}{l}{\(0.260{\scriptscriptstyle \pm 1.250}\)} & \multicolumn{1}{l}{\(89.49{\scriptscriptstyle \pm 6.25}\)} & \multicolumn{1}{l}{\(0.074{\scriptscriptstyle \pm 0.053}\)} & \multicolumn{1}{l}{\(99.00{\scriptscriptstyle \pm 10.00}\)} & \multicolumn{1}{l}{\(0.088{\scriptscriptstyle \pm 0.394}\)} \\ \cmidrule(lr){2-11}
\multicolumn{1}{l}{} & \multicolumn{1}{l}{\multirow{4}{*}{MEMIT}} & Edited & \multicolumn{1}{l}{\(100.00{\scriptscriptstyle \pm 0.00}\)} & \multicolumn{1}{l}{\(0.000{\scriptscriptstyle \pm 0.000}\)} & \multicolumn{1}{l}{\(18.00{\scriptscriptstyle \pm 38.61}\)} & \multicolumn{1}{l}{\(3.769{\scriptscriptstyle \pm 2.645}\)} & \multicolumn{1}{l}{\(100.00{\scriptscriptstyle \pm 0.00}\)} & \multicolumn{1}{l}{\(0.000{\scriptscriptstyle \pm 0.000}\)} & \multicolumn{1}{l}{\(5.00{\scriptscriptstyle \pm 21.90}\)} & \multicolumn{1}{l}{\(5.679{\scriptscriptstyle \pm 2.761}\)} \\
\multicolumn{1}{l}{} & \multicolumn{1}{l}{} & Reference & \multicolumn{1}{l}{\(18.00{\scriptscriptstyle \pm 0.39}\)} & \multicolumn{1}{l}{\(6.843{\scriptscriptstyle \pm 0.037}\)} & \multicolumn{1}{l}{\(100.00{\scriptscriptstyle \pm 0.00}\)} & \multicolumn{1}{l}{\(0.516{\scriptscriptstyle \pm 0.221}\)} & \multicolumn{1}{l}{\(5.00{\scriptscriptstyle \pm 0.22}\)} & \multicolumn{1}{l}{\(8.091{\scriptscriptstyle \pm 0.034}\)} & \multicolumn{1}{l}{\(99.00{\scriptscriptstyle \pm 10.00}\)} & \multicolumn{1}{l}{\(0.140{\scriptscriptstyle \pm 0.198}\)} \\
\multicolumn{1}{l}{} & \multicolumn{1}{l}{} & Re-edit & \multicolumn{1}{l}{\(\mathbf{99.39}{\scriptscriptstyle \pm \mathbf{1.39}}\)} & \multicolumn{1}{l}{\(\mathbf{0.009}{\scriptscriptstyle \pm \mathbf{0.037}}\)} & \multicolumn{1}{l}{\(\mathbf{100.00}{\scriptscriptstyle \pm \mathbf{0.00}}\)} & \multicolumn{1}{l}{\(\mathbf{0.375}{\scriptscriptstyle \pm \mathbf{0.214}}\)} & \multicolumn{1}{l}{\(\mathbf{96.23}{\scriptscriptstyle \pm \mathbf{3.04}}\)} & \multicolumn{1}{l}{\(\mathbf{0.024}{\scriptscriptstyle \pm \mathbf{0.072}}\)} & \multicolumn{1}{l}{\(\mathbf{99.00}{\scriptscriptstyle \pm \mathbf{10.00}}\)} & \multicolumn{1}{l}{\(0.151{\scriptscriptstyle \pm 0.159}\)} \\
\multicolumn{1}{l}{} & \multicolumn{1}{l}{} & Reverse & \multicolumn{1}{l}{\(83.43{\scriptscriptstyle \pm 11.17}\)} & \multicolumn{1}{l}{\(0.363{\scriptscriptstyle \pm 0.467}\)} & \multicolumn{1}{l}{\(98.00{\scriptscriptstyle \pm 14.07}\)} & \multicolumn{1}{l}{\(0.597{\scriptscriptstyle \pm 0.650}\)} & \multicolumn{1}{l}{\(93.33{\scriptscriptstyle \pm 2.09}\)} & \multicolumn{1}{l}{\(0.069{\scriptscriptstyle \pm 0.029}\)} & \multicolumn{1}{l}{\(\mathbf{99.00}{\scriptscriptstyle \pm \mathbf{10.00}}\)} & \multicolumn{1}{l}{\(\mathbf{0.145}{\scriptscriptstyle \pm \mathbf{0.193}}\)} \\ \cmidrule(lr){2-11}
\multicolumn{1}{l}{} & \multicolumn{1}{l}{\multirow{4}{*}{AlphaEdit}} & Edited & \multicolumn{1}{l}{\(100.00{\scriptscriptstyle \pm 0.00}\)} & \multicolumn{1}{l}{\(0.000{\scriptscriptstyle \pm 0.000}\)} & \multicolumn{1}{l}{\(12.00{\scriptscriptstyle \pm 32.66}\)} & \multicolumn{1}{l}{\(4.636{\scriptscriptstyle \pm 2.752}\)} & \multicolumn{1}{l}{\(100.00{\scriptscriptstyle \pm 0.00}\)} & \multicolumn{1}{l}{\(0.000{\scriptscriptstyle \pm 0.000}\)} & \multicolumn{1}{l}{\(6.00{\scriptscriptstyle \pm 23.87}\)} & \multicolumn{1}{l}{\(5.679{\scriptscriptstyle \pm 3.112}\)} \\
\multicolumn{1}{l}{} & \multicolumn{1}{l}{} & Reference & \multicolumn{1}{l}{\(12.00{\scriptscriptstyle \pm 0.33}\)} & \multicolumn{1}{l}{\(7.601{\scriptscriptstyle \pm 0.036}\)} & \multicolumn{1}{l}{\(100.00{\scriptscriptstyle \pm 0.00}\)} & \multicolumn{1}{l}{\(0.523{\scriptscriptstyle \pm 0.226}\)} & \multicolumn{1}{l}{\(6.00{\scriptscriptstyle \pm 0.24}\)} & \multicolumn{1}{l}{\(7.484{\scriptscriptstyle \pm 0.033}\)} & \multicolumn{1}{l}{\(99.00{\scriptscriptstyle \pm 10.00}\)} & \multicolumn{1}{l}{\(0.198{\scriptscriptstyle \pm 0.249}\)} \\
\multicolumn{1}{l}{} & \multicolumn{1}{l}{} & Re-edit & \multicolumn{1}{l}{\(\mathbf{99.35}{\scriptscriptstyle \pm \mathbf{1.00}}\)} & \multicolumn{1}{l}{\(\mathbf{0.008}{\scriptscriptstyle \pm \mathbf{0.030}}\)} & \multicolumn{1}{l}{\(\mathbf{100.00}{\scriptscriptstyle \pm \mathbf{0.00}}\)} & \multicolumn{1}{l}{\(\mathbf{0.373}{\scriptscriptstyle \pm \mathbf{0.213}}\)} & \multicolumn{1}{l}{\(\mathbf{96.65}{\scriptscriptstyle \pm \mathbf{3.42}}\)} & \multicolumn{1}{l}{\(\mathbf{0.031}{\scriptscriptstyle \pm \mathbf{0.076}}\)} & \multicolumn{1}{l}{\(\mathbf{99.00}{\scriptscriptstyle \pm \mathbf{10.00}}\)} & \multicolumn{1}{l}{\(\mathbf{0.162}{\scriptscriptstyle \pm \mathbf{0.212}}\)} \\
\multicolumn{1}{l}{} & \multicolumn{1}{l}{} & Reverse & \multicolumn{1}{l}{\(86.06{\scriptscriptstyle \pm 7.87}\)} & \multicolumn{1}{l}{\(0.363{\scriptscriptstyle \pm 0.256}\)} & \multicolumn{1}{l}{\(98.00{\scriptscriptstyle \pm 14.07}\)} & \multicolumn{1}{l}{\(0.612{\scriptscriptstyle \pm 0.672}\)} & \multicolumn{1}{l}{\(81.01{\scriptscriptstyle \pm 14.17}\)} & \multicolumn{1}{l}{\(0.533{\scriptscriptstyle \pm 0.397}\)} & \multicolumn{1}{l}{\(98.00{\scriptscriptstyle \pm 14.07}\)} & \multicolumn{1}{l}{\(0.266{\scriptscriptstyle \pm 0.736}\)} \\
\bottomrule
\end{tabular}%
}
  }
  \caption{Performance comparison with the global reversal and re-editing baselines when reversing one fact at a time. Agreement and KL divergence are reported on both the remained set and the reversed set across different base models and editing methods. Higher agreement and lower KL divergence indicate better performance.}
  \label{reverse_1_gptj_llama2}
\end{table*}

\section{More Experimental Results}
\label{moreresult}
\subsection{Results on GPT-J 6B and LLaMA2-7B}\label{gptj_llama2}
As shown in Tables~\ref{50_gptj_llama2} and~\ref{reverse_1_gptj_llama2}, we also observe several method-specific patterns from the additional results.

\paragraph{The proposed reversal framework works consistently well for single-layer editing methods.}
For single-layer editing methods such as ROME and SimIE, the proposed framework achieves strong performance on both GPT-J 6B and LLaMA2-7B. This is consistent with our main observation that edit-sensitive components are easier to isolate when the edit is concentrated in a single rewrite layer. Compared with the Re-edit baseline, our method generally achieves comparable reversed-set performance. In the Reverse-50 setting on ZsRE, it also preserves substantially more remaining edited knowledge, suggesting that spectral reversal can better avoid interference with previous edits when multiple facts are reversed simultaneously.

\paragraph{GPT-J 6B shows weaker performance under multi-layer editing methods.}
GPT-J 6B performs relatively poorly with multi-layer editing methods, especially MEMIT and AlphaEdit on CounterFact. This trend appears in both the Reverse-50 and Reverse-1 settings. One possible reason is that multi-layer editing distributes the edit signal across several MLP layers, making the corresponding edit-sensitive components more entangled across layers. Since our method optimizes the gated shrinkage layer by layer, the reversal at one layer may not fully compensate for edit effects jointly encoded across multiple layers. This issue is more visible on GPT-J 6B, suggesting that its edited representations may be less separable under multi-layer updates. In these challenging settings, Re-edit generally preserves the remaining edits better than our method, further indicating that the main limitation of our framework arises from isolating edit-sensitive components under strongly entangled multi-layer updates.

\paragraph{The degradation of GPT-J 6B under multi-layer editing is not equally severe across datasets.}
On ZsRE, GPT-J 6B still achieves strong reversed-set performance for MEMIT and AlphaEdit, although the remained-set performance is lower than that of single-layer methods. This suggests that the difficulty is not only determined by the model or editing method, but also by the dataset, the quality of the given rephrase prompts, and the strength of the edited association. CounterFact edits may induce more concentrated or conflicting changes in GPT-J 6B under multi-layer editing, and its rephrase prompts may be semantically less aligned with the original editing prompts, making selective reversal harder. The comparison with Re-edit further reflects this dataset-dependent behavior: while Re-edit is generally stronger on CounterFact, our method becomes more competitive on ZsRE and achieves better remained-set preservation in several Reverse-50 settings.

\begin{table*}[t!]
  \centering
  \footnotesize
  \renewcommand\arraystretch{1.5}
  \setlength{\tabcolsep}{1.1mm}{
  \resizebox{\textwidth}{!}{%
\begin{tabular}{llllllllllllll}
\toprule
\multicolumn{2}{l}{\multirow{3}{*}{}}                          & \multicolumn{4}{c}{\textbf{GPT2-XL}}                                                                                                                      & \multicolumn{4}{c}{\textbf{Mistral-7B}}                                                                                                                    & \multicolumn{4}{c}{\textbf{LLaMA3-8B}}                                                                                                                      \\ \cmidrule(lr){3-6}\cmidrule(lr){7-10}\cmidrule(lr){11-14} 
\multicolumn{2}{l}{}                                           & \multicolumn{2}{c}{Remained Set}                                                      & \multicolumn{2}{c}{Reversed Set}                                 & \multicolumn{2}{c}{Remained Set}                                                       & \multicolumn{2}{c}{Reversed Set}                                 & \multicolumn{2}{c}{Remained Set}                                                      & \multicolumn{2}{c}{Reversed Set}                                   \\ \cmidrule(lr){3-4}\cmidrule(lr){5-6}\cmidrule(lr){7-8}\cmidrule(lr){9-10}\cmidrule(lr){11-12}\cmidrule(lr){13-14} 
\multicolumn{2}{l}{}                                           & \multicolumn{1}{c}{Agree.}               & \multicolumn{1}{c}{KL Div.}               & \multicolumn{1}{c}{Agree.}               & \multicolumn{1}{c}{KL Div.}                & \multicolumn{1}{c}{Agree.}                & \multicolumn{1}{c}{KL Div.}               & \multicolumn{1}{c}{Agree.}               & \multicolumn{1}{c}{KL Div.}                & \multicolumn{1}{c}{Agree.}               & \multicolumn{1}{c}{KL Div.}               & \multicolumn{1}{c}{Agree.}                & \multicolumn{1}{c}{KL Div.}                \\ \midrule
\multicolumn{1}{l}{\multirow{3}{*}{\rotatebox[origin=c]{90}{Counterfact}}} & Full Gate & \multicolumn{1}{l}{\(74.00{\scriptscriptstyle \pm 5.66}\)}          & \multicolumn{1}{l}{\(0.497{\scriptscriptstyle \pm 0.143}\)}          & \multicolumn{1}{l}{\(96.00{\scriptscriptstyle \pm 0.00}\)}          & \(\textbf{0.022}{\scriptscriptstyle \pm \textbf{0.001}}\) & \multicolumn{1}{l}{\(90.00{\scriptscriptstyle \pm 5.66}\)}           & \multicolumn{1}{l}{\(0.073{\scriptscriptstyle \pm 0.011}\)}          & \multicolumn{1}{l}{\(\textbf{84.00}{\scriptscriptstyle \pm \textbf{5.66}}\)} & \(\textbf{0.103}{\scriptscriptstyle \pm \textbf{0.014}}\) & \multicolumn{1}{l}{\(84.00{\scriptscriptstyle \pm 5.66}\)}          & \multicolumn{1}{l}{\(0.239{\scriptscriptstyle \pm 0.051}\)}          & \multicolumn{1}{l}{\(\textbf{76.00}{\scriptscriptstyle \pm \textbf{11.31}}\)} & \(\textbf{0.336}{\scriptscriptstyle \pm \textbf{0.241}}\) \\ \addlinespace[2pt] 
\multicolumn{1}{l}{}                             & Only U Side & \multicolumn{1}{l}{\(\textbf{90.00}{\scriptscriptstyle \pm \textbf{2.83}}\)} & \multicolumn{1}{l}{\(\textbf{0.086}{\scriptscriptstyle \pm \textbf{0.025}}\)} & \multicolumn{1}{l}{\(93.00{\scriptscriptstyle \pm 1.41}\)}          & \(0.059{\scriptscriptstyle \pm 0.005}\)          & \multicolumn{1}{l}{\(88.00{\scriptscriptstyle \pm 0.00}\)}           & \multicolumn{1}{l}{\(0.053{\scriptscriptstyle \pm 0.013}\)}          & \multicolumn{1}{l}{\(53.00{\scriptscriptstyle \pm 9.90}\)}          & \(0.658{\scriptscriptstyle \pm 0.058}\)          & \multicolumn{1}{l}{\(87.00{\scriptscriptstyle \pm 7.07}\)}          & \multicolumn{1}{l}{\(0.116{\scriptscriptstyle \pm 0.052}\)}          & \multicolumn{1}{l}{\(53.00{\scriptscriptstyle \pm 12.73}\)}          & \(1.066{\scriptscriptstyle \pm 0.412}\)          \\
\multicolumn{1}{l}{}                             & Only V Side & \multicolumn{1}{l}{\(77.00{\scriptscriptstyle \pm 4.24}\)}          & \multicolumn{1}{l}{\(0.509{\scriptscriptstyle \pm 0.167}\)}          & \multicolumn{1}{l}{\(\textbf{97.00}{\scriptscriptstyle \pm \textbf{1.41}}\)} & \(0.025{\scriptscriptstyle \pm 0.002}\)          & \multicolumn{1}{l}{\(\textbf{93.00}{\scriptscriptstyle \pm \textbf{1.41}}\)}  & \multicolumn{1}{l}{\(\textbf{0.040}{\scriptscriptstyle \pm \textbf{0.015}}\)} & \multicolumn{1}{l}{\(74.00{\scriptscriptstyle \pm 5.66}\)}          & \(0.208{\scriptscriptstyle \pm 0.002}\)          & \multicolumn{1}{l}{\(\textbf{91.00}{\scriptscriptstyle \pm \textbf{7.07}}\)} & \multicolumn{1}{l}{\(\textbf{0.059}{\scriptscriptstyle \pm \textbf{0.014}}\)} & \multicolumn{1}{l}{\(74.00{\scriptscriptstyle \pm 8.49}\)}           & \(0.515{\scriptscriptstyle \pm 0.302}\)          \\ \midrule
\multicolumn{1}{l}{\multirow{3}{*}{\rotatebox[origin=c]{90}{ZsRE}}}        & Full Gate & \multicolumn{1}{l}{\(77.00{\scriptscriptstyle \pm 4.24}\)}          & \multicolumn{1}{l}{\(0.221{\scriptscriptstyle \pm 0.056}\)}          & \multicolumn{1}{l}{\(\textbf{97.00}{\scriptscriptstyle \pm \textbf{4.24}}\)} & \(\textbf{0.389}{\scriptscriptstyle \pm \textbf{0.230}}\) & \multicolumn{1}{l}{\(93.00{\scriptscriptstyle \pm 1.41}\)}           & \multicolumn{1}{l}{\(0.095{\scriptscriptstyle \pm 0.022}\)}          & \multicolumn{1}{l}{\(95.00{\scriptscriptstyle \pm 1.41}\)}          & \(\textbf{0.248}{\scriptscriptstyle \pm \textbf{0.064}}\) & \multicolumn{1}{l}{\(87.00{\scriptscriptstyle \pm 1.41}\)}          & \multicolumn{1}{l}{\(0.441{\scriptscriptstyle \pm 0.066}\)}          & \multicolumn{1}{l}{\(\textbf{73.00}{\scriptscriptstyle \pm \textbf{4.24}}\)}  & \(\textbf{0.487}{\scriptscriptstyle \pm \textbf{0.298}}\) \\ \addlinespace[2pt] 
\multicolumn{1}{l}{}                             & Only U Side & \multicolumn{1}{l}{\(\textbf{93.00}{\scriptscriptstyle \pm \textbf{4.24}}\)} & \multicolumn{1}{l}{\(\textbf{0.023}{\scriptscriptstyle \pm \textbf{0.006}}\)} & \multicolumn{1}{l}{\(87.00{\scriptscriptstyle \pm 1.41}\)}          & \(0.886{\scriptscriptstyle \pm 0.188}\)          & \multicolumn{1}{l}{\(\textbf{100.00}{\scriptscriptstyle \pm \textbf{0.00}}\)} & \multicolumn{1}{l}{\(\textbf{0.005}{\scriptscriptstyle \pm \textbf{0.001}}\)} & \multicolumn{1}{l}{\(43.00{\scriptscriptstyle \pm 1.41}\)}          & \(2.352{\scriptscriptstyle \pm 0.227}\)          & \multicolumn{1}{l}{\(\textbf{99.00}{\scriptscriptstyle \pm \textbf{1.41}}\)} & \multicolumn{1}{l}{\(\textbf{0.013}{\scriptscriptstyle \pm \textbf{0.003}}\)} & \multicolumn{1}{l}{\(4.00{\scriptscriptstyle \pm 2.83}\)}            & \(5.315{\scriptscriptstyle \pm 0.216}\)          \\
\multicolumn{1}{l}{}                             & Only V Side & \multicolumn{1}{l}{\(78.00{\scriptscriptstyle \pm 2.83}\)}          & \multicolumn{1}{l}{\(0.215{\scriptscriptstyle \pm 0.013}\)}          & \multicolumn{1}{l}{\(95.00{\scriptscriptstyle \pm 4.24}\)}          & \(0.505{\scriptscriptstyle \pm 0.277}\)          & \multicolumn{1}{l}{\(98.00{\scriptscriptstyle \pm 2.83}\)}           & \multicolumn{1}{l}{\(0.039{\scriptscriptstyle \pm 0.012}\)}          & \multicolumn{1}{l}{\(\textbf{96.00}{\scriptscriptstyle \pm \textbf{0.00}}\)} & \(0.395{\scriptscriptstyle \pm 0.098}\)          & \multicolumn{1}{l}{\(96.00{\scriptscriptstyle \pm 2.83}\)}          & \multicolumn{1}{l}{\(0.095{\scriptscriptstyle \pm 0.015}\)}          & \multicolumn{1}{l}{\(56.00{\scriptscriptstyle \pm 2.83}\)}           & \(1.011{\scriptscriptstyle \pm 0.339}\)          \\ \bottomrule
\end{tabular}%
}
  }
  \caption{
Effect of single-side gating on rephrase prompts. We compare the full method with Only-\(U\) and Only-\(V\) variants, where gates are applied only to the left or right singular vectors. Agreement and KL divergence are reported on both the remained set and the reversed set. Higher agreement and lower KL divergence indicate better performance.
}
  \label{one_side_rephrase}
\end{table*}

\begin{table*}[t!]
  \centering
  \footnotesize
  \renewcommand\arraystretch{1.5}
  \setlength{\tabcolsep}{1.1mm}{
  \resizebox{\textwidth}{!}{%
\begin{tabular}{llllllllllllll}
\toprule
\multicolumn{2}{l}{\multirow{3}{*}{}}                          & \multicolumn{4}{c}{\textbf{GPT2-XL}}                                                                                                                       & \multicolumn{4}{c}{\textbf{Mistral-7B}}                                                                                                                   & \multicolumn{4}{c}{\textbf{LLaMA3-8B}}                                                                                                                     \\ \cmidrule(lr){3-6}\cmidrule(lr){7-10}\cmidrule(lr){11-14} 
\multicolumn{2}{l}{}                                           & \multicolumn{2}{c}{Remained Set}                                                       & \multicolumn{2}{c}{Reversed Set}                                 & \multicolumn{2}{c}{Remained Set}                                                      & \multicolumn{2}{c}{Reversed Set}                                 & \multicolumn{2}{c}{Remained Set}                                                      & \multicolumn{2}{c}{Reversed Set}                                  \\ \cmidrule(lr){3-4}\cmidrule(lr){5-6}\cmidrule(lr){7-8}\cmidrule(lr){9-10}\cmidrule(lr){11-12}\cmidrule(lr){13-14} 
\multicolumn{2}{l}{}                                           & \multicolumn{1}{c}{Agree.}                & \multicolumn{1}{c}{KL Div.}               & \multicolumn{1}{c}{Agree.}               & \multicolumn{1}{c}{KL Div.}                & \multicolumn{1}{c}{Agree.}               & \multicolumn{1}{c}{KL Div.}               & \multicolumn{1}{c}{Agree.}               & \multicolumn{1}{c}{KL Div.}                & \multicolumn{1}{c}{Agree.}               & \multicolumn{1}{c}{KL Div.}               & \multicolumn{1}{c}{Agree.}               & \multicolumn{1}{c}{KL Div.}                \\ \midrule
\multicolumn{1}{l}{\multirow{3}{*}{\rotatebox[origin=c]{90}{Counterfact}}} & Full Gate & \multicolumn{1}{l}{\(89.00{\scriptscriptstyle \pm 1.41}\)}           & \multicolumn{1}{l}{\(0.585{\scriptscriptstyle \pm 0.036}\)}          & \multicolumn{1}{l}{\(59.00{\scriptscriptstyle \pm 4.24}\)}          & \(\textbf{0.659}{\scriptscriptstyle \pm \textbf{0.211}}\) & \multicolumn{1}{l}{\(96.00{\scriptscriptstyle \pm 0.00}\)}          & \multicolumn{1}{l}{\(0.088{\scriptscriptstyle \pm 0.023}\)}          & \multicolumn{1}{l}{\(\textbf{49.00}{\scriptscriptstyle \pm \textbf{9.90}}\)} & \(\textbf{1.420}{\scriptscriptstyle \pm \textbf{0.592}}\) & \multicolumn{1}{l}{\(100.00{\scriptscriptstyle \pm 0.00}\)}         & \multicolumn{1}{l}{\(0.051{\scriptscriptstyle \pm 0.031}\)}          & \multicolumn{1}{l}{\(\textbf{46.00}{\scriptscriptstyle \pm \textbf{8.49}}\)} & \(\textbf{2.220}{\scriptscriptstyle \pm \textbf{0.981}}\) \\ \addlinespace[2pt] 
\multicolumn{1}{l}{}                             & Only U Side & \multicolumn{1}{l}{\(\textbf{100.00}{\scriptscriptstyle \pm \textbf{0.00}}\)} & \multicolumn{1}{l}{\(\textbf{0.022}{\scriptscriptstyle \pm \textbf{0.009}}\)} & \multicolumn{1}{l}{\(31.00{\scriptscriptstyle \pm 4.24}\)}          & \(2.237{\scriptscriptstyle \pm 0.254}\)          & \multicolumn{1}{l}{\(96.00{\scriptscriptstyle \pm 2.83}\)}          & \multicolumn{1}{l}{\(0.041{\scriptscriptstyle \pm 0.013}\)}          & \multicolumn{1}{l}{\(14.00{\scriptscriptstyle \pm 5.66}\)}          & \(5.145{\scriptscriptstyle \pm 0.145}\)          & \multicolumn{1}{l}{\(100.00{\scriptscriptstyle \pm 0.00}\)}         & \multicolumn{1}{l}{\(0.004{\scriptscriptstyle \pm 0.002}\)}          & \multicolumn{1}{l}{\(1.00{\scriptscriptstyle \pm 1.41}\)}           & \(7.367{\scriptscriptstyle \pm 0.007}\)          \\
\multicolumn{1}{l}{}                             & Only V Side & \multicolumn{1}{l}{\(91.00{\scriptscriptstyle \pm 1.41}\)}           & \multicolumn{1}{l}{\(0.613{\scriptscriptstyle \pm 0.160}\)}          & \multicolumn{1}{l}{\(\textbf{59.00}{\scriptscriptstyle \pm \textbf{1.41}}\)} & \(0.703{\scriptscriptstyle \pm 0.309}\)          & \multicolumn{1}{l}{\(\textbf{97.00}{\scriptscriptstyle \pm \textbf{4.24}}\)} & \multicolumn{1}{l}{\(\textbf{0.011}{\scriptscriptstyle \pm \textbf{0.007}}\)} & \multicolumn{1}{l}{\(18.00{\scriptscriptstyle \pm 8.49}\)}          & \(3.420{\scriptscriptstyle \pm 0.302}\)          & \multicolumn{1}{l}{\(100.00{\scriptscriptstyle \pm 0.00}\)}         & \multicolumn{1}{l}{\(\textbf{0.002}{\scriptscriptstyle \pm \textbf{0.000}}\)} & \multicolumn{1}{l}{\(12.00{\scriptscriptstyle \pm 0.00}\)}          & \(5.046{\scriptscriptstyle \pm 0.293}\)          \\ \midrule
\multicolumn{1}{l}{\multirow{3}{*}{\rotatebox[origin=c]{90}{ZsRE}}}        & Full Gate & \multicolumn{1}{l}{\(92.00{\scriptscriptstyle \pm 5.66}\)}           & \multicolumn{1}{l}{\(0.135{\scriptscriptstyle \pm 0.003}\)}          & \multicolumn{1}{l}{\(\textbf{71.00}{\scriptscriptstyle \pm \textbf{7.07}}\)} & \(\textbf{1.678}{\scriptscriptstyle \pm \textbf{0.550}}\) & \multicolumn{1}{l}{\(94.00{\scriptscriptstyle \pm 2.83}\)}          & \multicolumn{1}{l}{\(0.110{\scriptscriptstyle \pm 0.077}\)}          & \multicolumn{1}{l}{\(\textbf{80.00}{\scriptscriptstyle \pm \textbf{8.49}}\)} & \(\textbf{0.948}{\scriptscriptstyle \pm \textbf{0.216}}\) & \multicolumn{1}{l}{\(93.00{\scriptscriptstyle \pm 1.41}\)}          & \multicolumn{1}{l}{\(0.263{\scriptscriptstyle \pm 0.062}\)}          & \multicolumn{1}{l}{\(\textbf{56.00}{\scriptscriptstyle \pm \textbf{2.83}}\)} & \(\textbf{1.148}{\scriptscriptstyle \pm \textbf{0.386}}\) \\ \addlinespace[2pt] 
\multicolumn{1}{l}{}                             & Only U Side & \multicolumn{1}{l}{\(\textbf{93.00}{\scriptscriptstyle \pm \textbf{1.41}}\)}  & \multicolumn{1}{l}{\(\textbf{0.016}{\scriptscriptstyle \pm \textbf{0.000}}\)} & \multicolumn{1}{l}{\(22.00{\scriptscriptstyle \pm 5.66}\)}          & \(3.369{\scriptscriptstyle \pm 0.467}\)          & \multicolumn{1}{l}{\(\textbf{97.00}{\scriptscriptstyle \pm \textbf{1.41}}\)} & \multicolumn{1}{l}{\(\textbf{0.003}{\scriptscriptstyle \pm \textbf{0.001}}\)} & \multicolumn{1}{l}{\(22.00{\scriptscriptstyle \pm 2.83}\)}          & \(3.954{\scriptscriptstyle \pm 0.396}\)          & \multicolumn{1}{l}{\(\textbf{99.00}{\scriptscriptstyle \pm \textbf{1.41}}\)} & \multicolumn{1}{l}{\(\textbf{0.005}{\scriptscriptstyle \pm \textbf{0.002}}\)} & \multicolumn{1}{l}{\(3.00{\scriptscriptstyle \pm 4.24}\)}           & \(8.477{\scriptscriptstyle \pm 0.505}\)          \\
\multicolumn{1}{l}{}                             & Only V Side & \multicolumn{1}{l}{\(91.00{\scriptscriptstyle \pm 7.07}\)}           & \multicolumn{1}{l}{\(0.159{\scriptscriptstyle \pm 0.015}\)}          & \multicolumn{1}{l}{\(57.00{\scriptscriptstyle \pm 9.90}\)}          & \(1.882{\scriptscriptstyle \pm 0.727}\)          & \multicolumn{1}{l}{\(94.00{\scriptscriptstyle \pm 2.83}\)}          & \multicolumn{1}{l}{\(0.074{\scriptscriptstyle \pm 0.051}\)}          & \multicolumn{1}{l}{\(69.00{\scriptscriptstyle \pm 7.07}\)}          & \(1.476{\scriptscriptstyle \pm 0.429}\)          & \multicolumn{1}{l}{\(96.00{\scriptscriptstyle \pm 2.83}\)}          & \multicolumn{1}{l}{\(0.054{\scriptscriptstyle \pm 0.028}\)}          & \multicolumn{1}{l}{\(30.00{\scriptscriptstyle \pm 5.66}\)}          & \(3.082{\scriptscriptstyle \pm 0.741}\)          \\ \bottomrule
\end{tabular}%
}
  }
  \caption{
Effect of single-side gating on original editing prompts. Reversal is optimized on rephrase prompts and evaluated on the original editing prompts. Compared with single-side gating, the full method shows stronger prompt generalization. Higher agreement and lower KL divergence indicate better performance.
}
  \label{one_side_original}
\end{table*}

\paragraph{Reversing one fact at a time is generally easier than reversing 50 facts simultaneously.}
Compared with Reverse-50, both our method and Re-edit generally perform better when reversing one fact at a time. Re-edit is particularly strong in the Reverse-1 setting, where it usually preserves nearly all remaining edits while maintaining high reversed-set performance. This suggests that direct re-editing introduces relatively limited interference when only one fact is reversed. In contrast, when 50 facts are reversed simultaneously, especially on ZsRE, our method can provide better preservation of the remaining edits, indicating an advantage of selective spectral reversal when interference from repeated re-editing becomes stronger. However, for GPT-J 6B with MEMIT and AlphaEdit on CounterFact, Reverse-1 does not substantially improve over Reverse-50. This indicates that the main bottleneck in this case may not be the number of reversed facts, but the entanglement introduced by multi-layer editing itself.

\subsection{Effect of Single-Side Gating}

To analyze the contribution of the two singular-vector sides, we compare the full gate with two variants: Only-\(U\), which applies gates only to the left singular vectors, and Only-\(V\), which applies gates only to the right singular vectors. The results on rephrase prompts and original editing prompts are shown in Tables~\ref{one_side_rephrase} and~\ref{one_side_original}, respectively.

On rephrase prompts, single-side gating can often achieve performance close to the full gate. This suggests that either side of the singular vectors can provide sufficient flexibility to fit the optimization prompts in some settings. However, the results on original editing prompts show a clear difference. Both Only-\(U\) and Only-\(V\) variants suffer from much weaker reversal performance, especially on the reversed set, while the full gate generalizes substantially better. This indicates that single-side gating may overfit the rephrase prompts and fail to capture edit-sensitive components that are robust across prompt forms.

These results suggest that jointly gating both \(U\) and \(V\) is important for prompt-level generalization. While one-side gating may be enough to fit the optimization signal, full two-side gating provides a more expressive and stable way to shrink entries within each dominant singular component, allowing the reversal to better transfer from rephrase prompts to the original editing prompts.

\begin{table*}[t!]
  \centering
  \footnotesize
  \renewcommand\arraystretch{1.2}
  \setlength{\tabcolsep}{1.5mm}{
  \resizebox{\textwidth}{!}{%
\begin{tabular}{lllllllllll}
\toprule
\multicolumn{3}{l}{\multirow{3}{*}{}} & \multicolumn{4}{c}{\textbf{GPT2-XL}} & \multicolumn{4}{c}{\textbf{LLaMA3-8B}} \\ \cmidrule(lr){4-7}\cmidrule(lr){8-11}
\multicolumn{3}{l}{} & \multicolumn{2}{c}{Remained Set} & \multicolumn{2}{c}{Reversed Set} & \multicolumn{2}{c}{Remained Set} & \multicolumn{2}{c}{Reversed Set} \\ \cmidrule(lr){4-5}\cmidrule(lr){6-7}\cmidrule(lr){8-9}\cmidrule(lr){10-11}
\multicolumn{3}{l}{} & \multicolumn{1}{c}{Agree.} & \multicolumn{1}{c}{KL Div.} & \multicolumn{1}{c}{Agree.} & \multicolumn{1}{c}{KL Div.} & \multicolumn{1}{c}{Agree.} & \multicolumn{1}{c}{KL Div.} & \multicolumn{1}{c}{Agree.} & \multicolumn{1}{c}{KL Div.} \\ \midrule
\multicolumn{1}{l}{\multirow{12}{*}{\rotatebox[origin=c]{90}{Counterfact}}} & \multicolumn{1}{l}{\multirow{4}{*}{SimIE}} & Edited & \multicolumn{1}{l}{\(100.00{\scriptscriptstyle \pm 0.00}\)} & \multicolumn{1}{l}{\(0.000{\scriptscriptstyle \pm 0.000}\)} & \multicolumn{1}{l}{\(5.00{\scriptscriptstyle \pm 1.41}\)} & \multicolumn{1}{l}{\(4.255{\scriptscriptstyle \pm 0.033}\)} & \multicolumn{1}{l}{\(100.00{\scriptscriptstyle \pm 0.00}\)} & \multicolumn{1}{l}{\(0.000{\scriptscriptstyle \pm 0.000}\)} & \multicolumn{1}{l}{\(14.00{\scriptscriptstyle \pm 5.66}\)} & \(4.841{\scriptscriptstyle \pm 0.554}\) \\
\multicolumn{1}{l}{} & \multicolumn{1}{l}{} & Reference & \multicolumn{1}{l}{\(9.53{\scriptscriptstyle \pm 0.07}\)} & \multicolumn{1}{l}{\(4.990{\scriptscriptstyle \pm 0.000}\)} & \multicolumn{1}{l}{\(94.00{\scriptscriptstyle \pm 0.00}\)} & \multicolumn{1}{l}{\(0.026{\scriptscriptstyle \pm 0.001}\)} & \multicolumn{1}{l}{\(21.32{\scriptscriptstyle \pm 0.37}\)} & \multicolumn{1}{l}{\(5.487{\scriptscriptstyle \pm 0.027}\)} & \multicolumn{1}{l}{\(83.00{\scriptscriptstyle \pm 1.41}\)} & \(0.215{\scriptscriptstyle \pm 0.096}\) \\
\multicolumn{1}{l}{} & \multicolumn{1}{l}{} & Re-edit & \multicolumn{1}{l}{\(\mathbf{75.42}{\scriptscriptstyle \pm \mathbf{0.22}}\)} & \multicolumn{1}{l}{\(\mathbf{0.194}{\scriptscriptstyle \pm \mathbf{0.009}}\)} & \multicolumn{1}{l}{\(78.00{\scriptscriptstyle \pm 5.66}\)} & \multicolumn{1}{l}{\(0.512{\scriptscriptstyle \pm 0.092}\)} & \multicolumn{1}{l}{\(\mathbf{80.63}{\scriptscriptstyle \pm \mathbf{0.00}}\)} & \multicolumn{1}{l}{\(\mathbf{0.205}{\scriptscriptstyle \pm \mathbf{0.004}}\)} & \multicolumn{1}{l}{\(80.00{\scriptscriptstyle \pm 5.66}\)} & \(0.303{\scriptscriptstyle \pm 0.085}\) \\
\multicolumn{1}{l}{} & \multicolumn{1}{l}{} & Reverse & \multicolumn{1}{l}{\(35.84{\scriptscriptstyle \pm 2.31}\)} & \multicolumn{1}{l}{\(1.576{\scriptscriptstyle \pm 0.212}\)} & \multicolumn{1}{l}{\(\mathbf{92.00}{\scriptscriptstyle \pm \mathbf{2.83}}\)} & \multicolumn{1}{l}{\(\mathbf{0.061}{\scriptscriptstyle \pm \mathbf{0.001}}\)} & \multicolumn{1}{l}{\(62.89{\scriptscriptstyle \pm 2.31}\)} & \multicolumn{1}{l}{\(0.917{\scriptscriptstyle \pm 0.194}\)} & \multicolumn{1}{l}{\(\mathbf{81.00}{\scriptscriptstyle \pm \mathbf{1.41}}\)} & \(\mathbf{0.256}{\scriptscriptstyle \pm \mathbf{0.087}}\) \\ \cmidrule(lr){2-11}
\multicolumn{1}{l}{} & \multicolumn{1}{l}{\multirow{4}{*}{MEMIT}} & Edited & \multicolumn{1}{l}{\(100.00{\scriptscriptstyle \pm 0.00}\)} & \multicolumn{1}{l}{\(0.000{\scriptscriptstyle \pm 0.000}\)} & \multicolumn{1}{l}{\(22.00{\scriptscriptstyle \pm 2.83}\)} & \multicolumn{1}{l}{\(2.713{\scriptscriptstyle \pm 0.155}\)} & \multicolumn{1}{l}{\(100.00{\scriptscriptstyle \pm 0.00}\)} & \multicolumn{1}{l}{\(0.000{\scriptscriptstyle \pm 0.000}\)} & \multicolumn{1}{l}{\(0.00{\scriptscriptstyle \pm 0.00}\)} & \(16.972{\scriptscriptstyle \pm 0.500}\) \\
\multicolumn{1}{l}{} & \multicolumn{1}{l}{} & Reference & \multicolumn{1}{l}{\(21.16{\scriptscriptstyle \pm 0.30}\)} & \multicolumn{1}{l}{\(3.846{\scriptscriptstyle \pm 0.006}\)} & \multicolumn{1}{l}{\(82.00{\scriptscriptstyle \pm 8.49}\)} & \multicolumn{1}{l}{\(0.123{\scriptscriptstyle \pm 0.020}\)} & \multicolumn{1}{l}{\(0.00{\scriptscriptstyle \pm 0.00}\)} & \multicolumn{1}{l}{\(8.468{\scriptscriptstyle \pm 0.011}\)} & \multicolumn{1}{l}{\(63.00{\scriptscriptstyle \pm 4.24}\)} & \(0.731{\scriptscriptstyle \pm 0.167}\) \\
\multicolumn{1}{l}{} & \multicolumn{1}{l}{} & Re-edit & \multicolumn{1}{l}{\(\mathbf{73.68}{\scriptscriptstyle \pm \mathbf{1.49}}\)} & \multicolumn{1}{l}{\(\mathbf{0.328}{\scriptscriptstyle \pm \mathbf{0.020}}\)} & \multicolumn{1}{l}{\(73.00{\scriptscriptstyle \pm 7.07}\)} & \multicolumn{1}{l}{\(0.216{\scriptscriptstyle \pm 0.035}\)} & \multicolumn{1}{l}{\(\mathbf{100.00}{\scriptscriptstyle \pm \mathbf{0.00}}\)} & \multicolumn{1}{l}{\(\mathbf{0.001}{\scriptscriptstyle \pm \mathbf{0.000}}\)} & \multicolumn{1}{l}{\(0.00{\scriptscriptstyle \pm 0.00}\)} & \(16.573{\scriptscriptstyle \pm 0.616}\) \\
\multicolumn{1}{l}{} & \multicolumn{1}{l}{} & Reverse & \multicolumn{1}{l}{\(61.42{\scriptscriptstyle \pm 4.54}\)} & \multicolumn{1}{l}{\(0.791{\scriptscriptstyle \pm 0.177}\)} & \multicolumn{1}{l}{\(\mathbf{79.00}{\scriptscriptstyle \pm \mathbf{7.07}}\)} & \multicolumn{1}{l}{\(\mathbf{0.124}{\scriptscriptstyle \pm \mathbf{0.023}}\)} & \multicolumn{1}{l}{\(27.53{\scriptscriptstyle \pm 23.89}\)} & \multicolumn{1}{l}{\(10.985{\scriptscriptstyle \pm 4.332}\)} & \multicolumn{1}{l}{\(\mathbf{30.00}{\scriptscriptstyle \pm \mathbf{14.14}}\)} & \(\mathbf{6.118}{\scriptscriptstyle \pm \mathbf{2.739}}\) \\ \cmidrule(lr){2-11}
\multicolumn{1}{l}{} & \multicolumn{1}{l}{\multirow{4}{*}{AlphaEdit}} & Edited & \multicolumn{1}{l}{\(100.00{\scriptscriptstyle \pm 0.00}\)} & \multicolumn{1}{l}{\(0.000{\scriptscriptstyle \pm 0.000}\)} & \multicolumn{1}{l}{\(27.00{\scriptscriptstyle \pm 1.41}\)} & \multicolumn{1}{l}{\(2.582{\scriptscriptstyle \pm 0.302}\)} & \multicolumn{1}{l}{\(100.00{\scriptscriptstyle \pm 0.00}\)} & \multicolumn{1}{l}{\(0.000{\scriptscriptstyle \pm 0.000}\)} & \multicolumn{1}{l}{\(21.00{\scriptscriptstyle \pm 9.90}\)} & \(4.398{\scriptscriptstyle \pm 0.224}\) \\
\multicolumn{1}{l}{} & \multicolumn{1}{l}{} & Reference & \multicolumn{1}{l}{\(28.84{\scriptscriptstyle \pm 0.15}\)} & \multicolumn{1}{l}{\(3.468{\scriptscriptstyle \pm 0.012}\)} & \multicolumn{1}{l}{\(82.00{\scriptscriptstyle \pm 2.83}\)} & \multicolumn{1}{l}{\(0.125{\scriptscriptstyle \pm 0.018}\)} & \multicolumn{1}{l}{\(24.47{\scriptscriptstyle \pm 0.67}\)} & \multicolumn{1}{l}{\(4.619{\scriptscriptstyle \pm 0.013}\)} & \multicolumn{1}{l}{\(56.00{\scriptscriptstyle \pm 0.00}\)} & \(0.729{\scriptscriptstyle \pm 0.082}\) \\
\multicolumn{1}{l}{} & \multicolumn{1}{l}{} & Re-edit & \multicolumn{1}{l}{\(\mathbf{88.68}{\scriptscriptstyle \pm \mathbf{0.97}}\)} & \multicolumn{1}{l}{\(\mathbf{0.077}{\scriptscriptstyle \pm \mathbf{0.015}}\)} & \multicolumn{1}{l}{\(79.00{\scriptscriptstyle \pm 4.24}\)} & \multicolumn{1}{l}{\(0.220{\scriptscriptstyle \pm 0.044}\)} & \multicolumn{1}{l}{\(\mathbf{55.95}{\scriptscriptstyle \pm \mathbf{5.28}}\)} & \multicolumn{1}{l}{\(\mathbf{1.066}{\scriptscriptstyle \pm \mathbf{0.249}}\)} & \multicolumn{1}{l}{\(\mathbf{62.00}{\scriptscriptstyle \pm \mathbf{5.66}}\)} & \(0.767{\scriptscriptstyle \pm 0.090}\) \\
\multicolumn{1}{l}{} & \multicolumn{1}{l}{} & Reverse & \multicolumn{1}{l}{\(76.05{\scriptscriptstyle \pm 1.56}\)} & \multicolumn{1}{l}{\(0.331{\scriptscriptstyle \pm 0.041}\)} & \multicolumn{1}{l}{\(\mathbf{80.00}{\scriptscriptstyle \pm \mathbf{2.83}}\)} & \multicolumn{1}{l}{\(\mathbf{0.127}{\scriptscriptstyle \pm \mathbf{0.020}}\)} & \multicolumn{1}{l}{\(43.63{\scriptscriptstyle \pm 0.82}\)} & \multicolumn{1}{l}{\(2.134{\scriptscriptstyle \pm 0.193}\)} & \multicolumn{1}{l}{\(55.00{\scriptscriptstyle \pm 1.41}\)} & \(\mathbf{0.722}{\scriptscriptstyle \pm \mathbf{0.072}}\) \\ \midrule
\multicolumn{1}{l}{\multirow{12}{*}{\rotatebox[origin=c]{90}{ZsRE}}} & \multicolumn{1}{l}{\multirow{4}{*}{SimIE}} & Edited & \multicolumn{1}{l}{\(100.00{\scriptscriptstyle \pm 0.00}\)} & \multicolumn{1}{l}{\(0.000{\scriptscriptstyle \pm 0.000}\)} & \multicolumn{1}{l}{\(3.00{\scriptscriptstyle \pm 4.24}\)} & \multicolumn{1}{l}{\(6.685{\scriptscriptstyle \pm 0.347}\)} & \multicolumn{1}{l}{\(100.00{\scriptscriptstyle \pm 0.00}\)} & \multicolumn{1}{l}{\(0.000{\scriptscriptstyle \pm 0.000}\)} & \multicolumn{1}{l}{\(6.00{\scriptscriptstyle \pm 0.00}\)} & \(6.324{\scriptscriptstyle \pm 0.043}\) \\
\multicolumn{1}{l}{} & \multicolumn{1}{l}{} & Reference & \multicolumn{1}{l}{\(3.37{\scriptscriptstyle \pm 0.00}\)} & \multicolumn{1}{l}{\(7.307{\scriptscriptstyle \pm 0.022}\)} & \multicolumn{1}{l}{\(100.00{\scriptscriptstyle \pm 0.00}\)} & \multicolumn{1}{l}{\(0.009{\scriptscriptstyle \pm 0.001}\)} & \multicolumn{1}{l}{\(6.42{\scriptscriptstyle \pm 0.15}\)} & \multicolumn{1}{l}{\(5.838{\scriptscriptstyle \pm 0.002}\)} & \multicolumn{1}{l}{\(79.00{\scriptscriptstyle \pm 1.41}\)} & \(0.091{\scriptscriptstyle \pm 0.010}\) \\
\multicolumn{1}{l}{} & \multicolumn{1}{l}{} & Re-edit & \multicolumn{1}{l}{\(56.21{\scriptscriptstyle \pm 17.42}\)} & \multicolumn{1}{l}{\(1.229{\scriptscriptstyle \pm 0.577}\)} & \multicolumn{1}{l}{\(\mathbf{100.00}{\scriptscriptstyle \pm \mathbf{0.00}}\)} & \multicolumn{1}{l}{\(0.191{\scriptscriptstyle \pm 0.003}\)} & \multicolumn{1}{l}{\(62.84{\scriptscriptstyle \pm 11.46}\)} & \multicolumn{1}{l}{\(1.163{\scriptscriptstyle \pm 0.493}\)} & \multicolumn{1}{l}{\(\mathbf{75.00}{\scriptscriptstyle \pm \mathbf{1.41}}\)} & \(\mathbf{0.219}{\scriptscriptstyle \pm \mathbf{0.045}}\) \\
\multicolumn{1}{l}{} & \multicolumn{1}{l}{} & Reverse & \multicolumn{1}{l}{\(\mathbf{72.63}{\scriptscriptstyle \pm \mathbf{9.23}}\)} & \multicolumn{1}{l}{\(\mathbf{0.699}{\scriptscriptstyle \pm \mathbf{0.306}}\)} & \multicolumn{1}{l}{\(\mathbf{100.00}{\scriptscriptstyle \pm \mathbf{0.00}}\)} & \multicolumn{1}{l}{\(\mathbf{0.113}{\scriptscriptstyle \pm \mathbf{0.028}}\)} & \multicolumn{1}{l}{\(\mathbf{78.05}{\scriptscriptstyle \pm \mathbf{3.94}}\)} & \multicolumn{1}{l}{\(\mathbf{0.557}{\scriptscriptstyle \pm \mathbf{0.072}}\)} & \multicolumn{1}{l}{\(70.00{\scriptscriptstyle \pm 2.83}\)} & \(0.365{\scriptscriptstyle \pm 0.055}\) \\ \cmidrule(lr){2-11}
\multicolumn{1}{l}{} & \multicolumn{1}{l}{\multirow{4}{*}{MEMIT}} & Edited & \multicolumn{1}{l}{\(100.00{\scriptscriptstyle \pm 0.00}\)} & \multicolumn{1}{l}{\(0.000{\scriptscriptstyle \pm 0.000}\)} & \multicolumn{1}{l}{\(6.00{\scriptscriptstyle \pm 0.00}\)} & \multicolumn{1}{l}{\(4.080{\scriptscriptstyle \pm 0.049}\)} & \multicolumn{1}{l}{\(100.00{\scriptscriptstyle \pm 0.00}\)} & \multicolumn{1}{l}{\(0.000{\scriptscriptstyle \pm 0.000}\)} & \multicolumn{1}{l}{\(0.00{\scriptscriptstyle \pm 0.00}\)} & \(14.911{\scriptscriptstyle \pm 0.051}\) \\
\multicolumn{1}{l}{} & \multicolumn{1}{l}{} & Reference & \multicolumn{1}{l}{\(7.63{\scriptscriptstyle \pm 0.07}\)} & \multicolumn{1}{l}{\(5.227{\scriptscriptstyle \pm 0.023}\)} & \multicolumn{1}{l}{\(100.00{\scriptscriptstyle \pm 0.00}\)} & \multicolumn{1}{l}{\(0.094{\scriptscriptstyle \pm 0.011}\)} & \multicolumn{1}{l}{\(0.11{\scriptscriptstyle \pm 0.00}\)} & \multicolumn{1}{l}{\(12.499{\scriptscriptstyle \pm 0.001}\)} & \multicolumn{1}{l}{\(51.00{\scriptscriptstyle \pm 1.41}\)} & \(0.488{\scriptscriptstyle \pm 0.009}\) \\
\multicolumn{1}{l}{} & \multicolumn{1}{l}{} & Re-edit & \multicolumn{1}{l}{\(40.95{\scriptscriptstyle \pm 33.20}\)} & \multicolumn{1}{l}{\(1.485{\scriptscriptstyle \pm 1.280}\)} & \multicolumn{1}{l}{\(\mathbf{100.00}{\scriptscriptstyle \pm \mathbf{0.00}}\)} & \multicolumn{1}{l}{\(0.207{\scriptscriptstyle \pm 0.008}\)} & \multicolumn{1}{l}{\(\mathbf{90.26}{\scriptscriptstyle \pm \mathbf{7.67}}\)} & \multicolumn{1}{l}{\(\mathbf{0.385}{\scriptscriptstyle \pm \mathbf{0.474}}\)} & \multicolumn{1}{l}{\(0.00{\scriptscriptstyle \pm 0.00}\)} & \(11.401{\scriptscriptstyle \pm 3.919}\) \\
\multicolumn{1}{l}{} & \multicolumn{1}{l}{} & Reverse & \multicolumn{1}{l}{\(\mathbf{72.47}{\scriptscriptstyle \pm \mathbf{1.71}}\)} & \multicolumn{1}{l}{\(\mathbf{0.383}{\scriptscriptstyle \pm \mathbf{0.046}}\)} & \multicolumn{1}{l}{\(98.00{\scriptscriptstyle \pm 0.00}\)} & \multicolumn{1}{l}{\(\mathbf{0.201}{\scriptscriptstyle \pm \mathbf{0.056}}\)} & \multicolumn{1}{l}{\(37.95{\scriptscriptstyle \pm 6.03}\)} & \multicolumn{1}{l}{\(4.904{\scriptscriptstyle \pm 1.174}\)} & \multicolumn{1}{l}{\(\mathbf{2.00}{\scriptscriptstyle \pm \mathbf{0.00}}\)} & \(\mathbf{8.005}{\scriptscriptstyle \pm \mathbf{0.100}}\) \\ \cmidrule(lr){2-11}
\multicolumn{1}{l}{} & \multicolumn{1}{l}{\multirow{4}{*}{AlphaEdit}} & Edited & \multicolumn{1}{l}{\(100.00{\scriptscriptstyle \pm 0.00}\)} & \multicolumn{1}{l}{\(0.000{\scriptscriptstyle \pm 0.000}\)} & \multicolumn{1}{l}{\(13.00{\scriptscriptstyle \pm 7.07}\)} & \multicolumn{1}{l}{\(4.417{\scriptscriptstyle \pm 0.354}\)} & \multicolumn{1}{l}{\(100.00{\scriptscriptstyle \pm 0.00}\)} & \multicolumn{1}{l}{\(0.000{\scriptscriptstyle \pm 0.000}\)} & \multicolumn{1}{l}{\(9.00{\scriptscriptstyle \pm 1.41}\)} & \(6.001{\scriptscriptstyle \pm 0.161}\) \\
\multicolumn{1}{l}{} & \multicolumn{1}{l}{} & Reference & \multicolumn{1}{l}{\(7.37{\scriptscriptstyle \pm 0.30}\)} & \multicolumn{1}{l}{\(5.807{\scriptscriptstyle \pm 0.018}\)} & \multicolumn{1}{l}{\(100.00{\scriptscriptstyle \pm 0.00}\)} & \multicolumn{1}{l}{\(0.087{\scriptscriptstyle \pm 0.004}\)} & \multicolumn{1}{l}{\(9.53{\scriptscriptstyle \pm 0.07}\)} & \multicolumn{1}{l}{\(5.179{\scriptscriptstyle \pm 0.018}\)} & \multicolumn{1}{l}{\(47.00{\scriptscriptstyle \pm 1.41}\)} & \(0.570{\scriptscriptstyle \pm 0.018}\) \\
\multicolumn{1}{l}{} & \multicolumn{1}{l}{} & Re-edit & \multicolumn{1}{l}{\(57.79{\scriptscriptstyle \pm 26.35}\)} & \multicolumn{1}{l}{\(0.955{\scriptscriptstyle \pm 0.752}\)} & \multicolumn{1}{l}{\(\mathbf{100.00}{\scriptscriptstyle \pm \mathbf{0.00}}\)} & \multicolumn{1}{l}{\(0.235{\scriptscriptstyle \pm 0.014}\)} & \multicolumn{1}{l}{\(13.89{\scriptscriptstyle \pm 0.60}\)} & \multicolumn{1}{l}{\(4.114{\scriptscriptstyle \pm 0.132}\)} & \multicolumn{1}{l}{\(37.00{\scriptscriptstyle \pm 1.41}\)} & \(0.815{\scriptscriptstyle \pm 0.087}\) \\
\multicolumn{1}{l}{} & \multicolumn{1}{l}{} & Reverse & \multicolumn{1}{l}{\(\mathbf{81.16}{\scriptscriptstyle \pm \mathbf{1.34}}\)} & \multicolumn{1}{l}{\(\mathbf{0.285}{\scriptscriptstyle \pm \mathbf{0.008}}\)} & \multicolumn{1}{l}{\(98.00{\scriptscriptstyle \pm 0.00}\)} & \multicolumn{1}{l}{\(\mathbf{0.214}{\scriptscriptstyle \pm \mathbf{0.081}}\)} & \multicolumn{1}{l}{\(\mathbf{42.79}{\scriptscriptstyle \pm \mathbf{10.35}}\)} & \multicolumn{1}{l}{\(\mathbf{2.281}{\scriptscriptstyle \pm \mathbf{0.459}}\)} & \multicolumn{1}{l}{\(\mathbf{46.00}{\scriptscriptstyle \pm \mathbf{0.00}}\)} & \(\mathbf{0.744}{\scriptscriptstyle \pm \mathbf{0.054}}\) \\ \bottomrule
\end{tabular}%
}
  }
 \caption{
Selective reversal results after 1000 edits, evaluated on rephrase prompts. The model is edited with 1000 facts, and 50 facts are reversed simultaneously. Agreement and KL divergence are reported on both the remained set and the reversed set across different base models and editing methods. Higher agreement and lower KL divergence indicate better performance.
}
  \label{1000_reverse_rephrase}
\end{table*}

\begin{table*}[t!]
  \centering
  \footnotesize
  \renewcommand\arraystretch{1.2}
  \setlength{\tabcolsep}{1.5mm}{
  \resizebox{\textwidth}{!}{%
\begin{tabular}{lllllllllll}
\toprule
\multicolumn{3}{l}{\multirow{3}{*}{}} & \multicolumn{4}{c}{\textbf{GPT2-XL}} & \multicolumn{4}{c}{\textbf{LLaMA3-8B}} \\ \cmidrule(lr){4-7}\cmidrule(lr){8-11} 
\multicolumn{3}{l}{} & \multicolumn{2}{c}{Remained Set} & \multicolumn{2}{c}{Reversed Set} & \multicolumn{2}{c}{Remained Set} & \multicolumn{2}{c}{Reversed Set} \\ \cmidrule(lr){4-5}\cmidrule(lr){6-7}\cmidrule(lr){8-9}\cmidrule(lr){10-11} 
\multicolumn{3}{l}{} & \multicolumn{1}{c}{Agree.} & \multicolumn{1}{c}{KL Div.} & \multicolumn{1}{c}{Agree.} & \multicolumn{1}{c}{KL Div.} & \multicolumn{1}{c}{Agree.} & \multicolumn{1}{c}{KL Div.} & \multicolumn{1}{c}{Agree.} & \multicolumn{1}{c}{KL Div.} \\ \midrule
\multicolumn{1}{l}{\multirow{12}{*}{\rotatebox[origin=c]{90}{Counterfact}}} & \multicolumn{1}{l}{\multirow{4}{*}{SimIE}} & Edited & \multicolumn{1}{l}{\(100.00{\scriptscriptstyle \pm 0.00}\)} & \multicolumn{1}{l}{\(0.000{\scriptscriptstyle \pm 0.000}\)} & \multicolumn{1}{l}{\(0.00{\scriptscriptstyle \pm 0.00}\)} & \multicolumn{1}{l}{\(7.190{\scriptscriptstyle \pm 0.185}\)} & \multicolumn{1}{l}{\(100.00{\scriptscriptstyle \pm 0.00}\)} & \multicolumn{1}{l}{\(0.000{\scriptscriptstyle \pm 0.000}\)} & \multicolumn{1}{l}{\(0.00{\scriptscriptstyle \pm 0.00}\)} & \multicolumn{1}{l}{\(8.675{\scriptscriptstyle \pm 0.130}\)} \\
\multicolumn{1}{l}{} & \multicolumn{1}{l}{} & Reference & \multicolumn{1}{l}{\(1.21{\scriptscriptstyle \pm 0.07}\)} & \multicolumn{1}{l}{\(7.904{\scriptscriptstyle \pm 0.009}\)} & \multicolumn{1}{l}{\(92.00{\scriptscriptstyle \pm 0.00}\)} & \multicolumn{1}{l}{\(0.039{\scriptscriptstyle \pm 0.016}\)} & \multicolumn{1}{l}{\(1.95{\scriptscriptstyle \pm 0.07}\)} & \multicolumn{1}{l}{\(9.351{\scriptscriptstyle \pm 0.035}\)} & \multicolumn{1}{l}{\(76.00{\scriptscriptstyle \pm 8.49}\)} & \multicolumn{1}{l}{\(0.294{\scriptscriptstyle \pm 0.067}\)} \\
\multicolumn{1}{l}{} & \multicolumn{1}{l}{} & Re-edit & \multicolumn{1}{l}{\(\mathbf{98.05}{\scriptscriptstyle \pm \mathbf{0.37}}\)} & \multicolumn{1}{l}{\(\mathbf{0.054}{\scriptscriptstyle \pm \mathbf{0.001}}\)} & \multicolumn{1}{l}{\(37.00{\scriptscriptstyle \pm 9.90}\)} & \multicolumn{1}{l}{\(2.054{\scriptscriptstyle \pm 0.057}\)} & \multicolumn{1}{l}{\(\mathbf{96.16}{\scriptscriptstyle \pm \mathbf{0.37}}\)} & \multicolumn{1}{l}{\(\mathbf{0.076}{\scriptscriptstyle \pm \mathbf{0.007}}\)} & \multicolumn{1}{l}{\(47.00{\scriptscriptstyle \pm 1.41}\)} & \multicolumn{1}{l}{\(2.093{\scriptscriptstyle \pm 0.356}\)} \\
\multicolumn{1}{l}{} & \multicolumn{1}{l}{} & Reverse & \multicolumn{1}{l}{\(51.32{\scriptscriptstyle \pm 9.90}\)} & \multicolumn{1}{l}{\(2.267{\scriptscriptstyle \pm 0.489}\)} & \multicolumn{1}{l}{\(\mathbf{62.00}{\scriptscriptstyle \pm \mathbf{8.49}}\)} & \multicolumn{1}{l}{\(\mathbf{1.030}{\scriptscriptstyle \pm \mathbf{0.053}}\)} & \multicolumn{1}{l}{\(78.53{\scriptscriptstyle \pm 4.76}\)} & \multicolumn{1}{l}{\(0.744{\scriptscriptstyle \pm 0.191}\)} & \multicolumn{1}{l}{\(\mathbf{52.00}{\scriptscriptstyle \pm \mathbf{0.00}}\)} & \multicolumn{1}{l}{\(\mathbf{1.546}{\scriptscriptstyle \pm \mathbf{0.173}}\)} \\ \cmidrule(lr){2-11}
\multicolumn{1}{l}{} & \multicolumn{1}{l}{\multirow{4}{*}{MEMIT}} & Edited & \multicolumn{1}{l}{\(100.00{\scriptscriptstyle \pm 0.00}\)} & \multicolumn{1}{l}{\(0.000{\scriptscriptstyle \pm 0.000}\)} & \multicolumn{1}{l}{\(4.00{\scriptscriptstyle \pm 2.83}\)} & \multicolumn{1}{l}{\(5.014{\scriptscriptstyle \pm 0.632}\)} & \multicolumn{1}{l}{\(100.00{\scriptscriptstyle \pm 0.00}\)} & \multicolumn{1}{l}{\(0.000{\scriptscriptstyle \pm 0.000}\)} & \multicolumn{1}{l}{\(0.00{\scriptscriptstyle \pm 0.00}\)} & \multicolumn{1}{l}{\(17.151{\scriptscriptstyle \pm 0.058}\)} \\
\multicolumn{1}{l}{} & \multicolumn{1}{l}{} & Reference & \multicolumn{1}{l}{\(5.16{\scriptscriptstyle \pm 0.15}\)} & \multicolumn{1}{l}{\(6.321{\scriptscriptstyle \pm 0.030}\)} & \multicolumn{1}{l}{\(86.00{\scriptscriptstyle \pm 2.83}\)} & \multicolumn{1}{l}{\(0.173{\scriptscriptstyle \pm 0.033}\)} & \multicolumn{1}{l}{\(0.00{\scriptscriptstyle \pm 0.00}\)} & \multicolumn{1}{l}{\(8.622{\scriptscriptstyle \pm 0.004}\)} & \multicolumn{1}{l}{\(65.00{\scriptscriptstyle \pm 7.07}\)} & \multicolumn{1}{l}{\(0.914{\scriptscriptstyle \pm 0.148}\)} \\
\multicolumn{1}{l}{} & \multicolumn{1}{l}{} & Re-edit & \multicolumn{1}{l}{\(\mathbf{85.89}{\scriptscriptstyle \pm \mathbf{0.74}}\)} & \multicolumn{1}{l}{\(\mathbf{0.209}{\scriptscriptstyle \pm \mathbf{0.032}}\)} & \multicolumn{1}{l}{\(66.00{\scriptscriptstyle \pm 5.66}\)} & \multicolumn{1}{l}{\(0.749{\scriptscriptstyle \pm 0.034}\)} & \multicolumn{1}{l}{\(\mathbf{100.00}{\scriptscriptstyle \pm \mathbf{0.00}}\)} & \multicolumn{1}{l}{\(\mathbf{0.001}{\scriptscriptstyle \pm \mathbf{0.000}}\)} & \multicolumn{1}{l}{\(0.00{\scriptscriptstyle \pm 0.00}\)} & \multicolumn{1}{l}{\(16.301{\scriptscriptstyle \pm 0.278}\)} \\
\multicolumn{1}{l}{} & \multicolumn{1}{l}{} & Reverse & \multicolumn{1}{l}{\(76.63{\scriptscriptstyle \pm 5.51}\)} & \multicolumn{1}{l}{\(0.617{\scriptscriptstyle \pm 0.161}\)} & \multicolumn{1}{l}{\(\mathbf{67.00}{\scriptscriptstyle \pm \mathbf{1.41}}\)} & \multicolumn{1}{l}{\(\mathbf{0.510}{\scriptscriptstyle \pm \mathbf{0.009}}\)} & \multicolumn{1}{l}{\(22.32{\scriptscriptstyle \pm 16.82}\)} & \multicolumn{1}{l}{\(12.241{\scriptscriptstyle \pm 4.350}\)} & \multicolumn{1}{l}{\(\mathbf{5.00}{\scriptscriptstyle \pm \mathbf{7.07}}\)} & \multicolumn{1}{l}{\(\mathbf{10.076}{\scriptscriptstyle \pm \mathbf{0.807}}\)} \\ \cmidrule(lr){2-11}
\multicolumn{1}{l}{} & \multicolumn{1}{l}{\multirow{4}{*}{AlphaEdit}} & Edited & \multicolumn{1}{l}{\(100.00{\scriptscriptstyle \pm 0.00}\)} & \multicolumn{1}{l}{\(0.000{\scriptscriptstyle \pm 0.000}\)} & \multicolumn{1}{l}{\(0.00{\scriptscriptstyle \pm 0.00}\)} & \multicolumn{1}{l}{\(6.639{\scriptscriptstyle \pm 0.091}\)} & \multicolumn{1}{l}{\(100.00{\scriptscriptstyle \pm 0.00}\)} & \multicolumn{1}{l}{\(0.000{\scriptscriptstyle \pm 0.000}\)} & \multicolumn{1}{l}{\(1.00{\scriptscriptstyle \pm 1.41}\)} & \multicolumn{1}{l}{\(9.891{\scriptscriptstyle \pm 0.500}\)} \\
\multicolumn{1}{l}{} & \multicolumn{1}{l}{} & Reference & \multicolumn{1}{l}{\(1.37{\scriptscriptstyle \pm 0.00}\)} & \multicolumn{1}{l}{\(7.962{\scriptscriptstyle \pm 0.009}\)} & \multicolumn{1}{l}{\(84.00{\scriptscriptstyle \pm 2.83}\)} & \multicolumn{1}{l}{\(0.178{\scriptscriptstyle \pm 0.031}\)} & \multicolumn{1}{l}{\(1.89{\scriptscriptstyle \pm 0.00}\)} & \multicolumn{1}{l}{\(8.263{\scriptscriptstyle \pm 0.023}\)} & \multicolumn{1}{l}{\(55.00{\scriptscriptstyle \pm 1.41}\)} & \multicolumn{1}{l}{\(1.118{\scriptscriptstyle \pm 0.183}\)} \\
\multicolumn{1}{l}{} & \multicolumn{1}{l}{} & Re-edit & \multicolumn{1}{l}{\(\mathbf{99.42}{\scriptscriptstyle \pm \mathbf{0.22}}\)} & \multicolumn{1}{l}{\(\mathbf{0.026}{\scriptscriptstyle \pm \mathbf{0.005}}\)} & \multicolumn{1}{l}{\(66.00{\scriptscriptstyle \pm 8.49}\)} & \multicolumn{1}{l}{\(1.033{\scriptscriptstyle \pm 0.152}\)} & \multicolumn{1}{l}{\(\mathbf{73.16}{\scriptscriptstyle \pm \mathbf{10.12}}\)} & \multicolumn{1}{l}{\(\mathbf{1.173}{\scriptscriptstyle \pm \mathbf{0.524}}\)} & \multicolumn{1}{l}{\(\mathbf{45.00}{\scriptscriptstyle \pm \mathbf{1.41}}\)} & \multicolumn{1}{l}{\(\mathbf{1.708}{\scriptscriptstyle \pm \mathbf{0.064}}\)} \\
\multicolumn{1}{l}{} & \multicolumn{1}{l}{} & Reverse & \multicolumn{1}{l}{\(95.26{\scriptscriptstyle \pm 0.74}\)} & \multicolumn{1}{l}{\(0.240{\scriptscriptstyle \pm 0.030}\)} & \multicolumn{1}{l}{\(\mathbf{68.00}{\scriptscriptstyle \pm \mathbf{2.83}}\)} & \multicolumn{1}{l}{\(\mathbf{0.669}{\scriptscriptstyle \pm \mathbf{0.082}}\)} & \multicolumn{1}{l}{\(51.21{\scriptscriptstyle \pm 2.90}\)} & \multicolumn{1}{l}{\(2.735{\scriptscriptstyle \pm 0.337}\)} & \multicolumn{1}{l}{\(42.00{\scriptscriptstyle \pm 5.66}\)} & \multicolumn{1}{l}{\(1.902{\scriptscriptstyle \pm 0.023}\)} \\ \midrule
\multicolumn{1}{l}{\multirow{12}{*}{\rotatebox[origin=c]{90}{ZsRE}}} & \multicolumn{1}{l}{\multirow{4}{*}{SimIE}} & Edited & \multicolumn{1}{l}{\(100.00{\scriptscriptstyle \pm 0.00}\)} & \multicolumn{1}{l}{\(0.000{\scriptscriptstyle \pm 0.000}\)} & \multicolumn{1}{l}{\(0.00{\scriptscriptstyle \pm 0.00}\)} & \multicolumn{1}{l}{\(8.319{\scriptscriptstyle \pm 0.132}\)} & \multicolumn{1}{l}{\(100.00{\scriptscriptstyle \pm 0.00}\)} & \multicolumn{1}{l}{\(0.000{\scriptscriptstyle \pm 0.000}\)} & \multicolumn{1}{l}{\(6.00{\scriptscriptstyle \pm 0.00}\)} & \multicolumn{1}{l}{\(7.013{\scriptscriptstyle \pm 0.321}\)} \\
\multicolumn{1}{l}{} & \multicolumn{1}{l}{} & Reference & \multicolumn{1}{l}{\(0.21{\scriptscriptstyle \pm 0.00}\)} & \multicolumn{1}{l}{\(8.337{\scriptscriptstyle \pm 0.011}\)} & \multicolumn{1}{l}{\(100.00{\scriptscriptstyle \pm 0.00}\)} & \multicolumn{1}{l}{\(0.011{\scriptscriptstyle \pm 0.002}\)} & \multicolumn{1}{l}{\(6.53{\scriptscriptstyle \pm 0.00}\)} & \multicolumn{1}{l}{\(6.392{\scriptscriptstyle \pm 0.001}\)} & \multicolumn{1}{l}{\(85.00{\scriptscriptstyle \pm 4.24}\)} & \multicolumn{1}{l}{\(0.082{\scriptscriptstyle \pm 0.000}\)} \\
\multicolumn{1}{l}{} & \multicolumn{1}{l}{} & Re-edit & \multicolumn{1}{l}{\(68.95{\scriptscriptstyle \pm 18.01}\)} & \multicolumn{1}{l}{\(1.092{\scriptscriptstyle \pm 0.544}\)} & \multicolumn{1}{l}{\(\mathbf{100.00}{\scriptscriptstyle \pm \mathbf{0.00}}\)} & \multicolumn{1}{l}{\(\mathbf{0.291}{\scriptscriptstyle \pm \mathbf{0.023}}\)} & \multicolumn{1}{l}{\(71.00{\scriptscriptstyle \pm 11.09}\)} & \multicolumn{1}{l}{\(1.051{\scriptscriptstyle \pm 0.458}\)} & \multicolumn{1}{l}{\(\mathbf{64.00}{\scriptscriptstyle \pm \mathbf{5.66}}\)} & \multicolumn{1}{l}{\(\mathbf{0.452}{\scriptscriptstyle \pm \mathbf{0.098}}\)} \\
\multicolumn{1}{l}{} & \multicolumn{1}{l}{} & Reverse & \multicolumn{1}{l}{\(\mathbf{83.21}{\scriptscriptstyle \pm \mathbf{7.67}}\)} & \multicolumn{1}{l}{\(\mathbf{0.560}{\scriptscriptstyle \pm \mathbf{0.277}}\)} & \multicolumn{1}{l}{\(91.00{\scriptscriptstyle \pm 1.41}\)} & \multicolumn{1}{l}{\(0.495{\scriptscriptstyle \pm 0.060}\)} & \multicolumn{1}{l}{\(\mathbf{84.26}{\scriptscriptstyle \pm \mathbf{2.75}}\)} & \multicolumn{1}{l}{\(\mathbf{0.502}{\scriptscriptstyle \pm \mathbf{0.049}}\)} & \multicolumn{1}{l}{\(57.00{\scriptscriptstyle \pm 4.24}\)} & \multicolumn{1}{l}{\(0.762{\scriptscriptstyle \pm 0.306}\)} \\ \cmidrule(lr){2-11}
\multicolumn{1}{l}{} & \multicolumn{1}{l}{\multirow{4}{*}{MEMIT}} & Edited & \multicolumn{1}{l}{\(100.00{\scriptscriptstyle \pm 0.00}\)} & \multicolumn{1}{l}{\(0.000{\scriptscriptstyle \pm 0.000}\)} & \multicolumn{1}{l}{\(1.00{\scriptscriptstyle \pm 1.41}\)} & \multicolumn{1}{l}{\(4.861{\scriptscriptstyle \pm 0.020}\)} & \multicolumn{1}{l}{\(100.00{\scriptscriptstyle \pm 0.00}\)} & \multicolumn{1}{l}{\(0.000{\scriptscriptstyle \pm 0.000}\)} & \multicolumn{1}{l}{\(0.00{\scriptscriptstyle \pm 0.00}\)} & \multicolumn{1}{l}{\(15.054{\scriptscriptstyle \pm 0.232}\)} \\
\multicolumn{1}{l}{} & \multicolumn{1}{l}{} & Reference & \multicolumn{1}{l}{\(4.63{\scriptscriptstyle \pm 0.15}\)} & \multicolumn{1}{l}{\(5.958{\scriptscriptstyle \pm 0.013}\)} & \multicolumn{1}{l}{\(100.00{\scriptscriptstyle \pm 0.00}\)} & \multicolumn{1}{l}{\(0.102{\scriptscriptstyle \pm 0.004}\)} & \multicolumn{1}{l}{\(0.21{\scriptscriptstyle \pm 0.00}\)} & \multicolumn{1}{l}{\(12.457{\scriptscriptstyle \pm 0.011}\)} & \multicolumn{1}{l}{\(64.00{\scriptscriptstyle \pm 2.83}\)} & \multicolumn{1}{l}{\(0.477{\scriptscriptstyle \pm 0.026}\)} \\
\multicolumn{1}{l}{} & \multicolumn{1}{l}{} & Re-edit & \multicolumn{1}{l}{\(44.84{\scriptscriptstyle \pm 36.77}\)} & \multicolumn{1}{l}{\(1.559{\scriptscriptstyle \pm 1.412}\)} & \multicolumn{1}{l}{\(\mathbf{100.00}{\scriptscriptstyle \pm \mathbf{0.00}}\)} & \multicolumn{1}{l}{\(\mathbf{0.306}{\scriptscriptstyle \pm \mathbf{0.059}}\)} & \multicolumn{1}{l}{\(\mathbf{89.05}{\scriptscriptstyle \pm \mathbf{10.12}}\)} & \multicolumn{1}{l}{\(\mathbf{0.446}{\scriptscriptstyle \pm \mathbf{0.558}}\)} & \multicolumn{1}{l}{\(0.00{\scriptscriptstyle \pm 0.00}\)} & \multicolumn{1}{l}{\(11.507{\scriptscriptstyle \pm 3.873}\)} \\
\multicolumn{1}{l}{} & \multicolumn{1}{l}{} & Reverse & \multicolumn{1}{l}{\(\mathbf{78.95}{\scriptscriptstyle \pm \mathbf{1.34}}\)} & \multicolumn{1}{l}{\(\mathbf{0.358}{\scriptscriptstyle \pm \mathbf{0.032}}\)} & \multicolumn{1}{l}{\(91.00{\scriptscriptstyle \pm 4.24}\)} & \multicolumn{1}{l}{\(0.559{\scriptscriptstyle \pm 0.046}\)} & \multicolumn{1}{l}{\(40.11{\scriptscriptstyle \pm 5.36}\)} & \multicolumn{1}{l}{\(4.819{\scriptscriptstyle \pm 1.190}\)} & \multicolumn{1}{l}{\(\mathbf{1.00}{\scriptscriptstyle \pm \mathbf{1.41}}\)} & \multicolumn{1}{l}{\(\mathbf{9.438}{\scriptscriptstyle \pm \mathbf{0.230}}\)} \\ \cmidrule(lr){2-11}
\multicolumn{1}{l}{} & \multicolumn{1}{l}{\multirow{4}{*}{AlphaEdit}} & Edited & \multicolumn{1}{l}{\(100.00{\scriptscriptstyle \pm 0.00}\)} & \multicolumn{1}{l}{\(0.000{\scriptscriptstyle \pm 0.000}\)} & \multicolumn{1}{l}{\(3.00{\scriptscriptstyle \pm 1.41}\)} & \multicolumn{1}{l}{\(6.221{\scriptscriptstyle \pm 0.580}\)} & \multicolumn{1}{l}{\(100.00{\scriptscriptstyle \pm 0.00}\)} & \multicolumn{1}{l}{\(0.000{\scriptscriptstyle \pm 0.000}\)} & \multicolumn{1}{l}{\(9.00{\scriptscriptstyle \pm 1.41}\)} & \multicolumn{1}{l}{\(6.807{\scriptscriptstyle \pm 0.008}\)} \\
\multicolumn{1}{l}{} & \multicolumn{1}{l}{} & Reference & \multicolumn{1}{l}{\(2.47{\scriptscriptstyle \pm 0.07}\)} & \multicolumn{1}{l}{\(7.060{\scriptscriptstyle \pm 0.009}\)} & \multicolumn{1}{l}{\(100.00{\scriptscriptstyle \pm 0.00}\)} & \multicolumn{1}{l}{\(0.094{\scriptscriptstyle \pm 0.008}\)} & \multicolumn{1}{l}{\(9.84{\scriptscriptstyle \pm 0.22}\)} & \multicolumn{1}{l}{\(5.690{\scriptscriptstyle \pm 0.003}\)} & \multicolumn{1}{l}{\(58.00{\scriptscriptstyle \pm 2.83}\)} & \multicolumn{1}{l}{\(0.648{\scriptscriptstyle \pm 0.029}\)} \\
\multicolumn{1}{l}{} & \multicolumn{1}{l}{} & Re-edit & \multicolumn{1}{l}{\(66.47{\scriptscriptstyle \pm 26.87}\)} & \multicolumn{1}{l}{\(0.984{\scriptscriptstyle \pm 0.818}\)} & \multicolumn{1}{l}{\(\mathbf{99.00}{\scriptscriptstyle \pm \mathbf{1.41}}\)} & \multicolumn{1}{l}{\(\mathbf{0.334}{\scriptscriptstyle \pm \mathbf{0.046}}\)} & \multicolumn{1}{l}{\(13.68{\scriptscriptstyle \pm 0.74}\)} & \multicolumn{1}{l}{\(4.470{\scriptscriptstyle \pm 0.154}\)} & \multicolumn{1}{l}{\(27.00{\scriptscriptstyle \pm 4.24}\)} & \multicolumn{1}{l}{\(1.015{\scriptscriptstyle \pm 0.164}\)} \\
\multicolumn{1}{l}{} & \multicolumn{1}{l}{} & Reverse & \multicolumn{1}{l}{\(\mathbf{88.58}{\scriptscriptstyle \pm \mathbf{1.41}}\)} & \multicolumn{1}{l}{\(\mathbf{0.256}{\scriptscriptstyle \pm \mathbf{0.036}}\)} & \multicolumn{1}{l}{\(83.00{\scriptscriptstyle \pm 1.41}\)} & \multicolumn{1}{l}{\(0.724{\scriptscriptstyle \pm 0.144}\)} & \multicolumn{1}{l}{\(\mathbf{48.95}{\scriptscriptstyle \pm \mathbf{9.53}}\)} & \multicolumn{1}{l}{\(\mathbf{2.300}{\scriptscriptstyle \pm \mathbf{0.513}}\)} & \multicolumn{1}{l}{\(\mathbf{43.00}{\scriptscriptstyle \pm \mathbf{1.41}}\)} & \multicolumn{1}{l}{\(\mathbf{0.922}{\scriptscriptstyle \pm \mathbf{0.029}}\)} \\
\bottomrule
\end{tabular}%
}
  }
 \caption{
Selective reversal results after 1000 edits, evaluated on original editing prompts. The model is edited with 1000 facts, and 50 facts are reversed simultaneously using rephrase prompts for optimization. Agreement and KL divergence are reported on both the remained set and the reversed set across different base models and editing methods. Higher agreement and lower KL divergence indicate better performance.
}
  \label{1000_reverse_original}
\end{table*}

\subsection{Selective Reversal after 1000 Edits}
\label{1000edits}

To further evaluate whether the proposed spectral reversal framework can scale to a larger number of edits, we conduct additional experiments on models edited with 1000 facts. Since ROME performs poorly under this large-scale sequential editing setting, we only report results for SimIE, MEMIT, and AlphaEdit. We evaluate the reversal of 50 facts at a time and report both rephrase-prompt performance and original-prompt performance in Tables~\ref{1000_reverse_rephrase} and~\ref{1000_reverse_original}.

Overall, the results show that the proposed method can still achieve meaningful selective reversal in the 1000-edit setting, although the performance is less stable than in the 100-edit setting. On GPT2-XL, our reversed model maintains strong reversed-set performance across most editing methods and datasets while preserving a non-trivial portion of the remained edits. This suggests that even after many edits, the proposed spectral gating method can still identify edit-sensitive components and selectively suppress them.

The comparison with Re-edit further reveals a trade-off between reversal effectiveness and preservation of the remaining edits. On CounterFact, Re-edit generally preserves more of the remained set, whereas our method often achieves stronger reversed-set performance. In contrast, on ZsRE, our method often preserves more remaining edits than Re-edit while maintaining comparable reversed-set performance, particularly on GPT2-XL. These results indicate that the relative advantage of spectral reversal and direct re-editing depends on the dataset and the structure of the accumulated edits.

The results also show that large-scale editing makes selective reversal more difficult. Compared with the 100-edit setting, the remained-set performance drops more noticeably, especially for multi-layer editing methods. This is consistent with our hypothesis that individual edits may be sparsely represented within the dominant singular components, but as the number of edits increases, their edit-sensitive components become more likely to overlap or entangle. Therefore, reversing 50 edits after 1000 edits is substantially more difficult than reversing edits from a model containing only 100 edits.

The performance also depends strongly on the editing method and base model. MEMIT shows weak overall selective reversal performance for both the Re-edit baseline and our spectral reversal method in several settings, especially on LLaMA3-8B. This suggests that the difficulty is not unique to our spectral reversal method, but is partly caused by the strongly entangled edited representations produced by large-scale multi-layer editing. One possible reason is that MEMIT distributes the edited information across multiple layers, making the corresponding edit-sensitive components difficult to isolate through layer-wise spectral shrinkage. In addition, the large KL divergence between the edited and pretrained models, together with the weaker performance of the Reference model in these settings, suggests that the global-reversal reference becomes less reliable when the edited model moves substantially away from the pretrained state.

Finally, comparing Tables~\ref{1000_reverse_rephrase} and~\ref{1000_reverse_original}, reversed-set performance generally decreases on the original editing prompts, as expected because the model is optimized using rephrase prompts. Nevertheless, our method still substantially improves over the edited model in many settings, indicating that the learned spectral modification is not restricted to the specific rephrase prompts used during optimization. The comparison with Re-edit shows a similar dataset-dependent trade-off on the original prompts. Overall, these results demonstrate that selective reversal remains feasible after 1000 edits, while also highlighting the increased difficulty of balancing reversal and preservation as the accumulated edits become more entangled.

\begin{table*}[t!]
  \centering
  \footnotesize
  \renewcommand\arraystretch{1.2}
  \setlength{\tabcolsep}{1.1mm}{
  \resizebox{\textwidth}{!}{%
  \begin{tabular}{lllllllllllllll}
  \toprule
\multicolumn{3}{l}{\multirow{3}{*}{}} & \multicolumn{4}{c}{\textbf{GPT2-XL}} & \multicolumn{4}{c}{\textbf{Mistral-7B}} & \multicolumn{4}{c}{\textbf{LLaMA3-8B}} \\ \cmidrule(lr){4-7}\cmidrule(lr){8-11}\cmidrule(lr){12-15}
\multicolumn{3}{l}{} & \multicolumn{2}{c}{Remained Set} & \multicolumn{2}{c}{Reversed Set} & \multicolumn{2}{c}{Remained Set} & \multicolumn{2}{c}{Reversed Set} & \multicolumn{2}{c}{Remained Set} & \multicolumn{2}{c}{Reversed Set} \\ \cmidrule(lr){4-5}\cmidrule(lr){6-7}\cmidrule(lr){8-9}\cmidrule(lr){10-11}\cmidrule(lr){12-13}\cmidrule(lr){14-15}
\multicolumn{3}{l}{} & \multicolumn{1}{c}{Agree.} & \multicolumn{1}{c}{KL Div.} & \multicolumn{1}{c}{Agree.} & \multicolumn{1}{c}{KL Div.} & \multicolumn{1}{c}{Agree.} & \multicolumn{1}{c}{KL Div.} & \multicolumn{1}{c}{Agree.} & \multicolumn{1}{c}{KL Div.} & \multicolumn{1}{c}{Agree.} & \multicolumn{1}{c}{KL Div.} & \multicolumn{1}{c}{Agree.} & \multicolumn{1}{c}{KL Div.} \\ \midrule

\multicolumn{1}{c}{\multirow{8}{*}{\rotatebox[origin=c]{90}{CounterFact}}} & \multicolumn{1}{l}{\multirow{2}{*}{ROME}} & Method-Reverse & \multicolumn{1}{l}{\(77.00{\scriptscriptstyle \pm 9.90}\)} & \multicolumn{1}{l}{\(0.466{\scriptscriptstyle \pm 0.232}\)} & \multicolumn{1}{l}{\(\mathbf{87.00}{\scriptscriptstyle \pm \mathbf{7.07}}\)} & \multicolumn{1}{l}{\(\mathbf{0.118}{\scriptscriptstyle \pm \mathbf{0.011}}\)} & \multicolumn{1}{l}{\(83.00{\scriptscriptstyle \pm 1.41}\)} & \multicolumn{1}{l}{\(0.130{\scriptscriptstyle \pm 0.013}\)} & \multicolumn{1}{l}{\(73.00{\scriptscriptstyle \pm 4.24}\)} & \multicolumn{1}{l}{\(0.197{\scriptscriptstyle \pm 0.008}\)} & \multicolumn{1}{l}{\(72.00{\scriptscriptstyle \pm 0.00}\)} & \multicolumn{1}{l}{\(0.425{\scriptscriptstyle \pm 0.063}\)} & \multicolumn{1}{l}{\(70.00{\scriptscriptstyle \pm 5.66}\)} & \multicolumn{1}{l}{\(0.409{\scriptscriptstyle \pm 0.168}\)} \\
\multicolumn{1}{l}{} & \multicolumn{1}{l}{} & Alpha-Reverse & \multicolumn{1}{l}{\(\mathbf{85.00}{\scriptscriptstyle \pm \mathbf{4.24}}\)} & \multicolumn{1}{l}{\(\mathbf{0.112}{\scriptscriptstyle \pm \mathbf{0.038}}\)} & \multicolumn{1}{l}{\(85.00{\scriptscriptstyle \pm 7.07}\)} & \multicolumn{1}{l}{\(0.125{\scriptscriptstyle \pm 0.018}\)} & \multicolumn{1}{l}{\(\mathbf{94.00}{\scriptscriptstyle \pm \mathbf{2.83}}\)} & \multicolumn{1}{l}{\(\mathbf{0.020}{\scriptscriptstyle \pm \mathbf{0.005}}\)} & \multicolumn{1}{l}{\(\mathbf{81.00}{\scriptscriptstyle \pm \mathbf{9.90}}\)} & \multicolumn{1}{l}{\(\mathbf{0.130}{\scriptscriptstyle \pm \mathbf{0.009}}\)} & \multicolumn{1}{l}{\(\mathbf{91.00}{\scriptscriptstyle \pm \mathbf{1.41}}\)} & \multicolumn{1}{l}{\(\mathbf{0.053}{\scriptscriptstyle \pm \mathbf{0.014}}\)} & \multicolumn{1}{l}{\(\mathbf{76.00}{\scriptscriptstyle \pm \mathbf{2.83}}\)} & \multicolumn{1}{l}{\(\mathbf{0.284}{\scriptscriptstyle \pm \mathbf{0.144}}\)} \\ \cmidrule(lr){2-15}
\multicolumn{1}{l}{} & \multicolumn{1}{l}{\multirow{2}{*}{SimIE}} & Method-Reverse & \multicolumn{1}{l}{\(51.00{\scriptscriptstyle \pm 12.73}\)} & \multicolumn{1}{l}{\(1.322{\scriptscriptstyle \pm 0.485}\)} & \multicolumn{1}{l}{\(\mathbf{87.00}{\scriptscriptstyle \pm \mathbf{1.41}}\)} & \multicolumn{1}{l}{\(\mathbf{0.072}{\scriptscriptstyle \pm \mathbf{0.003}}\)} & \multicolumn{1}{l}{\(60.00{\scriptscriptstyle \pm 14.14}\)} & \multicolumn{1}{l}{\(0.633{\scriptscriptstyle \pm 0.060}\)} & \multicolumn{1}{l}{\(70.00{\scriptscriptstyle \pm 11.31}\)} & \multicolumn{1}{l}{\(0.261{\scriptscriptstyle \pm 0.051}\)} & \multicolumn{1}{l}{\(46.00{\scriptscriptstyle \pm 0.00}\)} & \multicolumn{1}{l}{\(1.777{\scriptscriptstyle \pm 0.532}\)} & \multicolumn{1}{l}{\(72.00{\scriptscriptstyle \pm 5.66}\)} & \multicolumn{1}{l}{\(0.458{\scriptscriptstyle \pm 0.260}\)} \\
\multicolumn{1}{l}{} & \multicolumn{1}{l}{} & Alpha-Reverse & \multicolumn{1}{l}{\(\mathbf{90.00}{\scriptscriptstyle \pm \mathbf{11.31}}\)} & \multicolumn{1}{l}{\(\mathbf{0.095}{\scriptscriptstyle \pm \mathbf{0.000}}\)} & \multicolumn{1}{l}{\(84.00{\scriptscriptstyle \pm 5.66}\)} & \multicolumn{1}{l}{\(0.111{\scriptscriptstyle \pm 0.003}\)} & \multicolumn{1}{l}{\(\mathbf{91.00}{\scriptscriptstyle \pm \mathbf{4.24}}\)} & \multicolumn{1}{l}{\(\mathbf{0.016}{\scriptscriptstyle \pm \mathbf{0.008}}\)} & \multicolumn{1}{l}{\(\mathbf{72.00}{\scriptscriptstyle \pm \mathbf{8.49}}\)} & \multicolumn{1}{l}{\(\mathbf{0.194}{\scriptscriptstyle \pm \mathbf{0.047}}\)} & \multicolumn{1}{l}{\(\mathbf{91.00}{\scriptscriptstyle \pm \mathbf{4.24}}\)} & \multicolumn{1}{l}{\(\mathbf{0.046}{\scriptscriptstyle \pm \mathbf{0.003}}\)} & \multicolumn{1}{l}{\(\mathbf{74.00}{\scriptscriptstyle \pm \mathbf{14.14}}\)} & \multicolumn{1}{l}{\(\mathbf{0.311}{\scriptscriptstyle \pm \mathbf{0.171}}\)} \\ \cmidrule(lr){2-15}
\multicolumn{1}{l}{} & \multicolumn{1}{l}{\multirow{2}{*}{MEMIT}} & Method-Reverse & \multicolumn{1}{l}{\(88.00{\scriptscriptstyle \pm 8.49}\)} & \multicolumn{1}{l}{\(0.030{\scriptscriptstyle \pm 0.006}\)} & \multicolumn{1}{l}{\(\mathbf{80.00}{\scriptscriptstyle \pm \mathbf{5.66}}\)} & \multicolumn{1}{l}{\(\mathbf{0.164}{\scriptscriptstyle \pm \mathbf{0.006}}\)} & \multicolumn{1}{l}{\(36.00{\scriptscriptstyle \pm 5.66}\)} & \multicolumn{1}{l}{\(2.302{\scriptscriptstyle \pm 0.014}\)} & \multicolumn{1}{l}{\(29.00{\scriptscriptstyle \pm 4.24}\)} & \multicolumn{1}{l}{\(2.699{\scriptscriptstyle \pm 0.246}\)} & \multicolumn{1}{l}{\(7.00{\scriptscriptstyle \pm 1.41}\)} & \multicolumn{1}{l}{\(6.459{\scriptscriptstyle \pm 0.663}\)} & \multicolumn{1}{l}{\(10.00{\scriptscriptstyle \pm 0.00}\)} & \multicolumn{1}{l}{\(4.604{\scriptscriptstyle \pm 0.290}\)} \\
\multicolumn{1}{l}{} & \multicolumn{1}{l}{} & Alpha-Reverse & \multicolumn{1}{l}{\(\mathbf{90.00}{\scriptscriptstyle \pm \mathbf{2.83}}\)} & \multicolumn{1}{l}{\(\mathbf{0.022}{\scriptscriptstyle \pm \mathbf{0.009}}\)} & \multicolumn{1}{l}{\(79.00{\scriptscriptstyle \pm 7.07}\)} & \multicolumn{1}{l}{\(0.174{\scriptscriptstyle \pm 0.015}\)} & \multicolumn{1}{l}{\(\mathbf{84.00}{\scriptscriptstyle \pm \mathbf{0.00}}\)} & \multicolumn{1}{l}{\(\mathbf{0.125}{\scriptscriptstyle \pm \mathbf{0.047}}\)} & \multicolumn{1}{l}{\(\mathbf{65.00}{\scriptscriptstyle \pm \mathbf{9.90}}\)} & \multicolumn{1}{l}{\(\mathbf{0.542}{\scriptscriptstyle \pm \mathbf{0.044}}\)} & \multicolumn{1}{l}{\(\mathbf{77.00}{\scriptscriptstyle \pm \mathbf{9.90}}\)} & \multicolumn{1}{l}{\(\mathbf{0.380}{\scriptscriptstyle \pm \mathbf{0.072}}\)} & \multicolumn{1}{l}{\(\mathbf{59.00}{\scriptscriptstyle \pm \mathbf{4.24}}\)} & \multicolumn{1}{l}{\(\mathbf{0.716}{\scriptscriptstyle \pm \mathbf{0.003}}\)} \\ \cmidrule(lr){2-15}
\multicolumn{1}{l}{} & \multicolumn{1}{l}{\multirow{2}{*}{AlphaEdit}} & Method-Reverse & \multicolumn{1}{l}{\(\mathbf{96.00}{\scriptscriptstyle \pm \mathbf{2.83}}\)} & \multicolumn{1}{l}{\(\mathbf{0.028}{\scriptscriptstyle \pm \mathbf{0.010}}\)} & \multicolumn{1}{l}{\(\mathbf{81.00}{\scriptscriptstyle \pm \mathbf{7.07}}\)} & \multicolumn{1}{l}{\(\mathbf{0.200}{\scriptscriptstyle \pm \mathbf{0.014}}\)} & \multicolumn{1}{l}{\(\mathbf{82.00}{\scriptscriptstyle \pm \mathbf{14.14}}\)} & \multicolumn{1}{l}{\(\mathbf{0.194}{\scriptscriptstyle \pm \mathbf{0.111}}\)} & \multicolumn{1}{l}{\(\mathbf{56.00}{\scriptscriptstyle \pm \mathbf{11.31}}\)} & \multicolumn{1}{l}{\(\mathbf{0.551}{\scriptscriptstyle \pm \mathbf{0.022}}\)} & \multicolumn{1}{l}{\(\mathbf{79.00}{\scriptscriptstyle \pm \mathbf{7.07}}\)} & \multicolumn{1}{l}{\(\mathbf{0.302}{\scriptscriptstyle \pm \mathbf{0.028}}\)} & \multicolumn{1}{l}{\(\mathbf{51.00}{\scriptscriptstyle \pm \mathbf{1.41}}\)} & \multicolumn{1}{l}{\(\mathbf{1.003}{\scriptscriptstyle \pm \mathbf{0.119}}\)} \\
\multicolumn{1}{l}{} & \multicolumn{1}{l}{} & Alpha-Reverse & \multicolumn{1}{l}{\(\mathbf{96.00}{\scriptscriptstyle \pm \mathbf{2.83}}\)} & \multicolumn{1}{l}{\(\mathbf{0.028}{\scriptscriptstyle \pm \mathbf{0.010}}\)} & \multicolumn{1}{l}{\(\mathbf{81.00}{\scriptscriptstyle \pm \mathbf{7.07}}\)} & \multicolumn{1}{l}{\(\mathbf{0.200}{\scriptscriptstyle \pm \mathbf{0.014}}\)} & \multicolumn{1}{l}{\(\mathbf{82.00}{\scriptscriptstyle \pm \mathbf{14.14}}\)} & \multicolumn{1}{l}{\(\mathbf{0.194}{\scriptscriptstyle \pm \mathbf{0.111}}\)} & \multicolumn{1}{l}{\(\mathbf{56.00}{\scriptscriptstyle \pm \mathbf{11.31}}\)} & \multicolumn{1}{l}{\(\mathbf{0.551}{\scriptscriptstyle \pm \mathbf{0.022}}\)} & \multicolumn{1}{l}{\(\mathbf{79.00}{\scriptscriptstyle \pm \mathbf{7.07}}\)} & \multicolumn{1}{l}{\(\mathbf{0.302}{\scriptscriptstyle \pm \mathbf{0.028}}\)} & \multicolumn{1}{l}{\(\mathbf{51.00}{\scriptscriptstyle \pm \mathbf{1.41}}\)} & \multicolumn{1}{l}{\(\mathbf{1.003}{\scriptscriptstyle \pm \mathbf{0.119}}\)} \\ \midrule

\multicolumn{1}{c}{\multirow{8}{*}{\rotatebox[origin=c]{90}{ZsRE}}} & \multicolumn{1}{l}{\multirow{2}{*}{ROME}} & Method-Reverse & \multicolumn{1}{l}{\(16.00{\scriptscriptstyle \pm 11.31}\)} & \multicolumn{1}{l}{\(3.210{\scriptscriptstyle \pm 1.346}\)} & \multicolumn{1}{l}{\(98.00{\scriptscriptstyle \pm 0.00}\)} & \multicolumn{1}{l}{\(\mathbf{0.137}{\scriptscriptstyle \pm \mathbf{0.002}}\)} & \multicolumn{1}{l}{\(39.00{\scriptscriptstyle \pm 1.41}\)} & \multicolumn{1}{l}{\(3.122{\scriptscriptstyle \pm 0.381}\)} & \multicolumn{1}{l}{\(\mathbf{97.00}{\scriptscriptstyle \pm \mathbf{1.41}}\)} & \multicolumn{1}{l}{\(0.240{\scriptscriptstyle \pm 0.019}\)} & \multicolumn{1}{l}{\(25.00{\scriptscriptstyle \pm 1.41}\)} & \multicolumn{1}{l}{\(3.959{\scriptscriptstyle \pm 0.103}\)} & \multicolumn{1}{l}{\(\mathbf{63.00}{\scriptscriptstyle \pm \mathbf{7.07}}\)} & \multicolumn{1}{l}{\(0.346{\scriptscriptstyle \pm 0.084}\)} \\
\multicolumn{1}{l}{} & \multicolumn{1}{l}{} & Alpha-Reverse & \multicolumn{1}{l}{\(\mathbf{47.00}{\scriptscriptstyle \pm \mathbf{18.38}}\)} & \multicolumn{1}{l}{\(\mathbf{1.298}{\scriptscriptstyle \pm \mathbf{1.166}}\)} & \multicolumn{1}{l}{\(\mathbf{99.00}{\scriptscriptstyle \pm \mathbf{1.41}}\)} & \multicolumn{1}{l}{\(0.194{\scriptscriptstyle \pm 0.003}\)} & \multicolumn{1}{l}{\(\mathbf{77.00}{\scriptscriptstyle \pm \mathbf{7.07}}\)} & \multicolumn{1}{l}{\(\mathbf{0.540}{\scriptscriptstyle \pm \mathbf{0.316}}\)} & \multicolumn{1}{l}{\(\mathbf{97.00}{\scriptscriptstyle \pm \mathbf{1.41}}\)} & \multicolumn{1}{l}{\(\mathbf{0.100}{\scriptscriptstyle \pm \mathbf{0.013}}\)} & \multicolumn{1}{l}{\(\mathbf{86.00}{\scriptscriptstyle \pm \mathbf{11.31}}\)} & \multicolumn{1}{l}{\(\mathbf{0.748}{\scriptscriptstyle \pm \mathbf{0.606}}\)} & \multicolumn{1}{l}{\(61.00{\scriptscriptstyle \pm 7.07}\)} & \multicolumn{1}{l}{\(\mathbf{0.303}{\scriptscriptstyle \pm \mathbf{0.083}}\)} \\ \cmidrule(lr){2-15}
\multicolumn{1}{l}{} & \multicolumn{1}{l}{\multirow{2}{*}{SimIE}} & Method-Reverse & \multicolumn{1}{l}{\(21.00{\scriptscriptstyle \pm 1.41}\)} & \multicolumn{1}{l}{\(4.169{\scriptscriptstyle \pm 0.587}\)} & \multicolumn{1}{l}{\(\mathbf{100.00}{\scriptscriptstyle \pm \mathbf{0.00}}\)} & \multicolumn{1}{l}{\(\mathbf{0.071}{\scriptscriptstyle \pm \mathbf{0.013}}\)} & \multicolumn{1}{l}{\(44.00{\scriptscriptstyle \pm 2.83}\)} & \multicolumn{1}{l}{\(2.433{\scriptscriptstyle \pm 0.467}\)} & \multicolumn{1}{l}{\(95.00{\scriptscriptstyle \pm 1.41}\)} & \multicolumn{1}{l}{\(0.129{\scriptscriptstyle \pm 0.009}\)} & \multicolumn{1}{l}{\(12.00{\scriptscriptstyle \pm 2.83}\)} & \multicolumn{1}{l}{\(5.123{\scriptscriptstyle \pm 0.475}\)} & \multicolumn{1}{l}{\(62.00{\scriptscriptstyle \pm 16.97}\)} & \multicolumn{1}{l}{\(0.289{\scriptscriptstyle \pm 0.066}\)} \\
\multicolumn{1}{l}{} & \multicolumn{1}{l}{} & Alpha-Reverse & \multicolumn{1}{l}{\(\mathbf{64.00}{\scriptscriptstyle \pm \mathbf{11.31}}\)} & \multicolumn{1}{l}{\(\mathbf{1.000}{\scriptscriptstyle \pm \mathbf{0.475}}\)} & \multicolumn{1}{l}{\(99.00{\scriptscriptstyle \pm 1.41}\)} & \multicolumn{1}{l}{\(0.147{\scriptscriptstyle \pm 0.015}\)} & \multicolumn{1}{l}{\(\mathbf{76.00}{\scriptscriptstyle \pm \mathbf{14.14}}\)} & \multicolumn{1}{l}{\(\mathbf{0.480}{\scriptscriptstyle \pm \mathbf{0.382}}\)} & \multicolumn{1}{l}{\(\mathbf{96.00}{\scriptscriptstyle \pm \mathbf{0.00}}\)} & \multicolumn{1}{l}{\(\mathbf{0.093}{\scriptscriptstyle \pm \mathbf{0.033}}\)} & \multicolumn{1}{l}{\(\mathbf{61.00}{\scriptscriptstyle \pm \mathbf{7.07}}\)} & \multicolumn{1}{l}{\(\mathbf{0.868}{\scriptscriptstyle \pm \mathbf{0.446}}\)} & \multicolumn{1}{l}{\(\mathbf{76.00}{\scriptscriptstyle \pm \mathbf{11.31}}\)} & \multicolumn{1}{l}{\(\mathbf{0.197}{\scriptscriptstyle \pm \mathbf{0.062}}\)} \\ \cmidrule(lr){2-15}
\multicolumn{1}{l}{} & \multicolumn{1}{l}{\multirow{2}{*}{MEMIT}} & Method-Reverse & \multicolumn{1}{l}{\(66.00{\scriptscriptstyle \pm 0.00}\)} & \multicolumn{1}{l}{\(\mathbf{0.628}{\scriptscriptstyle \pm \mathbf{0.081}}\)} & \multicolumn{1}{l}{\(\mathbf{99.00}{\scriptscriptstyle \pm \mathbf{1.41}}\)} & \multicolumn{1}{l}{\(\mathbf{0.141}{\scriptscriptstyle \pm \mathbf{0.024}}\)} & \multicolumn{1}{l}{\(23.00{\scriptscriptstyle \pm 7.07}\)} & \multicolumn{1}{l}{\(6.221{\scriptscriptstyle \pm 0.484}\)} & \multicolumn{1}{l}{\(97.00{\scriptscriptstyle \pm 1.41}\)} & \multicolumn{1}{l}{\(0.903{\scriptscriptstyle \pm 0.276}\)} & \multicolumn{1}{l}{\(2.00{\scriptscriptstyle \pm 2.83}\)} & \multicolumn{1}{l}{\(9.832{\scriptscriptstyle \pm 2.528}\)} & \multicolumn{1}{l}{\(19.00{\scriptscriptstyle \pm 24.04}\)} & \multicolumn{1}{l}{\(4.025{\scriptscriptstyle \pm 2.280}\)} \\
\multicolumn{1}{l}{} & \multicolumn{1}{l}{} & Alpha-Reverse & \multicolumn{1}{l}{\(\mathbf{67.00}{\scriptscriptstyle \pm \mathbf{1.41}}\)} & \multicolumn{1}{l}{\(0.639{\scriptscriptstyle \pm 0.038}\)} & \multicolumn{1}{l}{\(\mathbf{99.00}{\scriptscriptstyle \pm \mathbf{1.41}}\)} & \multicolumn{1}{l}{\(0.142{\scriptscriptstyle \pm 0.024}\)} & \multicolumn{1}{l}{\(\mathbf{56.00}{\scriptscriptstyle \pm \mathbf{25.46}}\)} & \multicolumn{1}{l}{\(\mathbf{1.291}{\scriptscriptstyle \pm \mathbf{0.953}}\)} & \multicolumn{1}{l}{\(\mathbf{98.00}{\scriptscriptstyle \pm \mathbf{0.00}}\)} & \multicolumn{1}{l}{\(\mathbf{0.210}{\scriptscriptstyle \pm \mathbf{0.049}}\)} & \multicolumn{1}{l}{\(\mathbf{47.00}{\scriptscriptstyle \pm \mathbf{7.07}}\)} & \multicolumn{1}{l}{\(\mathbf{2.382}{\scriptscriptstyle \pm \mathbf{0.144}}\)} & \multicolumn{1}{l}{\(\mathbf{38.00}{\scriptscriptstyle \pm \mathbf{2.83}}\)} & \multicolumn{1}{l}{\(\mathbf{0.652}{\scriptscriptstyle \pm \mathbf{0.013}}\)} \\ \cmidrule(lr){2-15}
\multicolumn{1}{l}{} & \multicolumn{1}{l}{\multirow{2}{*}{AlphaEdit}} & Method-Reverse & \multicolumn{1}{l}{\(\mathbf{53.00}{\scriptscriptstyle \pm \mathbf{29.70}}\)} & \multicolumn{1}{l}{\(\mathbf{1.629}{\scriptscriptstyle \pm \mathbf{1.765}}\)} & \multicolumn{1}{l}{\(\mathbf{99.00}{\scriptscriptstyle \pm \mathbf{1.41}}\)} & \multicolumn{1}{l}{\(\mathbf{0.207}{\scriptscriptstyle \pm \mathbf{0.010}}\)} & \multicolumn{1}{l}{\(\mathbf{71.00}{\scriptscriptstyle \pm \mathbf{12.73}}\)} & \multicolumn{1}{l}{\(\mathbf{0.756}{\scriptscriptstyle \pm \mathbf{0.111}}\)} & \multicolumn{1}{l}{\(\mathbf{98.00}{\scriptscriptstyle \pm \mathbf{0.00}}\)} & \multicolumn{1}{l}{\(\mathbf{0.199}{\scriptscriptstyle \pm \mathbf{0.027}}\)} & \multicolumn{1}{l}{\(\mathbf{46.00}{\scriptscriptstyle \pm \mathbf{2.83}}\)} & \multicolumn{1}{l}{\(\mathbf{2.830}{\scriptscriptstyle \pm \mathbf{0.313}}\)} & \multicolumn{1}{l}{\(\mathbf{40.00}{\scriptscriptstyle \pm \mathbf{16.97}}\)} & \multicolumn{1}{l}{\(\mathbf{0.614}{\scriptscriptstyle \pm \mathbf{0.158}}\)} \\
\multicolumn{1}{l}{} & \multicolumn{1}{l}{} & Alpha-Reverse & \multicolumn{1}{l}{\(\mathbf{53.00}{\scriptscriptstyle \pm \mathbf{29.70}}\)} & \multicolumn{1}{l}{\(\mathbf{1.629}{\scriptscriptstyle \pm \mathbf{1.765}}\)} & \multicolumn{1}{l}{\(\mathbf{99.00}{\scriptscriptstyle \pm \mathbf{1.41}}\)} & \multicolumn{1}{l}{\(\mathbf{0.207}{\scriptscriptstyle \pm \mathbf{0.010}}\)} & \multicolumn{1}{l}{\(\mathbf{71.00}{\scriptscriptstyle \pm \mathbf{12.73}}\)} & \multicolumn{1}{l}{\(\mathbf{0.756}{\scriptscriptstyle \pm \mathbf{0.111}}\)} & \multicolumn{1}{l}{\(\mathbf{98.00}{\scriptscriptstyle \pm \mathbf{0.00}}\)} & \multicolumn{1}{l}{\(\mathbf{0.199}{\scriptscriptstyle \pm \mathbf{0.027}}\)} & \multicolumn{1}{l}{\(\mathbf{46.00}{\scriptscriptstyle \pm \mathbf{2.83}}\)} & \multicolumn{1}{l}{\(\mathbf{2.830}{\scriptscriptstyle \pm \mathbf{0.313}}\)} & \multicolumn{1}{l}{\(\mathbf{40.00}{\scriptscriptstyle \pm \mathbf{16.97}}\)} & \multicolumn{1}{l}{\(\mathbf{0.614}{\scriptscriptstyle \pm \mathbf{0.158}}\)} \\

\bottomrule
\end{tabular}%
}
}
\caption{Comparison between Method-Reverse and Alpha-Reverse baselines under the Reverse-50 setting. Agreement and KL divergence are reported on both the remained set and the reversed set across different base models and editing methods. Higher agreement and lower KL divergence indicate better performance. For AlphaEdit-edited models, the two implementations are identical by definition and therefore yield the same performance.}
\label{tab:method_alpha_reverse_comparison}
\end{table*}

\section{Comparison of Re-editing Strategies}
\label{app:reedit_same_method}

In the main experiments, we implement the Re-edit baseline using AlphaEdit~\cite{alphaedit} across all edited models. Another natural implementation is to use the same editing method as the original forward edit for re-editing, which may intuitively be more compatible with the corresponding edited model. We denote this variant as \textit{Method-Reverse} and the AlphaEdit-based implementation as \textit{Alpha-Reverse}. Their performance under the Reverse-50 setting is shown in Table~\ref{tab:method_alpha_reverse_comparison}.

Overall, Alpha-Reverse achieves better performance than Method-Reverse across most settings, particularly in preserving the remaining edited knowledge. The improvement is especially clear for multi-layer editing methods such as MEMIT, where Method-Reverse can substantially damage the remained set, while Alpha-Reverse preserves considerably more of the remaining edits. On the reversed set, Alpha-Reverse also achieves generally comparable or better performance, indicating that its stronger preservation of the remained set does not come at the cost of reversal effectiveness.

One possible reason is that, compared with other editing methods, AlphaEdit constrains its updates to a null-space subspace, which may reduce interference with previous edits even when information about these edits is unavailable.

These results suggest that using AlphaEdit as a unified Re-edit implementation is more effective than reusing the original forward editing method. Moreover, it does not require identifying the specific editing algorithm originally applied to the model, making it a more practical baseline for the selective reversal setting.

\section{Effect of Renormalization}
\label{app:renormalization}

The entry-wise gates selectively suppress edit-sensitive entries in the dominant singular vectors. However, directly applying the gates will decrease the magnitude of the corresponding rank-one singular component, which may weaken the representation of the remaining edited knowledge encoded in the same component. To mitigate this effect, we renormalize the gated singular vectors such that each modified rank-one component preserves the Frobenius norm of its original component.

Specifically, let the $i$-th rank-one singular component be
\begin{equation}
W_i = \sigma_i u_i v_i^\top .
\end{equation}

Its Frobenius norm is
\begin{equation}
\begin{aligned}
\|W_i\|_F
&=
|\sigma_i| \|u_i v_i^\top\|_F\\
&=
|\sigma_i| \|u_i\|_2 \|v_i\|_2.
\end{aligned}
\end{equation}

After applying the entry-wise gates, we renormalize the gated singular vectors to their original norms:
\begin{equation}
\begin{aligned}
\tilde{u}_i
&=
\frac{\|u_i\|_2}
{\|g_i^u \odot u_i\|_2}
\left(g_i^u \odot u_i\right),
\\
\tilde{v}_i
&=
\frac{\|v_i\|_2}
{\|g_i^v \odot v_i\|_2}
\left(g_i^v \odot v_i\right).
\end{aligned}
\end{equation}

Therefore, the modified rank-one component
\(
\tilde{W}_i
\)
satisfies
\begin{equation}
\begin{aligned}
\|\tilde{W}_i\|_F
&=
|\sigma_i| \|\tilde{u}_i \tilde{v}_i^\top\|_F\\
&=
|\sigma_i|
\|\tilde{u}_i\|_2
\|\tilde{v}_i\|_2 \\
&=
|\sigma_i|
\|u_i\|_2
\|v_i\|_2\\
&=
\|W_i\|_F.
\end{aligned}
\end{equation}

To further illustrate the effect of renormalization, let
\begin{equation}
\bar{W}_i
=
\sigma_i
(g_i^u \odot u_i)
(g_i^v \odot v_i)^\top
\end{equation}
denote the gated component before renormalization. Then, for any input $k$,
\begin{equation}
\tilde{W}_i k
=
c_i \bar{W}_i k,\quad
c_i
=
\frac{\|u_i\|_2\|v_i\|_2}
{\|g_i^u\odot u_i\|_2
 \|g_i^v\odot v_i\|_2}.
\end{equation}

Thus, renormalization scales the gated rank-one component without changing the output direction induced by the gating operation. It preserves the relative suppression pattern learned by the gates while compensating for the overall magnitude reduction introduced by shrinkage, thereby strengthening the representation of the remaining edited knowledge relative to the unnormalized gated component.

\end{document}